%% file: ArticleIJCV2.tex
\pdfoutput=1
\RequirePackage{fix-cm}

\documentclass[twocolumn]{svjour3}          
\smartqed  
\usepackage{graphicx}
\usepackage{verbatim} 
\usepackage{amssymb}
\usepackage{amsmath}
\usepackage{hyphenat}
\usepackage{algorithmic}
\usepackage{subfigure}
\usepackage{wasysym}
\usepackage{universal}
\usepackage{color}
\usepackage{xspace}
\usepackage{array}
\definecolor{pink}{RGB}{255,0,255}

\makeatletter
\DeclareRobustCommand\onedot{\futurelet\@let@token\@onedot}
\def\@onedot{\ifx\@let@token.\else.\null\fi\xspace}

\def\eg{\emph{e.g}\onedot} 
\def\ie{\emph{i.e}\onedot}

\def\etal{\emph{et al}\onedot}
\makeatother

\journalname{IJCV}
\begin{document}

\title{Hypothesize and Bound: A Computational Focus of Attention Mechanism for Simultaneous 3D Shape Reconstruction, Pose Estimation and Classification from a Single 2D Image}


\titlerunning{H\&B: A Computational FoAM for Simultaneous 3D Reconstruction, Pose Estimation, and Classification}
\author{Diego Rother \and Siddharth Mahendran \and Ren\'e Vidal}   


\institute{D. Rother \and S. Mahendran \and R. Vidal \at
              Johns Hopkins University \\
              Tel.: +1-410-516-6736\\
              \email{diroth@gmail.com}           
}

\date{Received: date / Accepted: date}

\maketitle

\input Abstract
\input Introduction
\input PriorWork
\input ProblemDefinition
\input LowerBound
\input UpperBound

\input BoundingMechanism

\input Summary
\input Results
\input Conclusions

\begin{acknowledgements}
We would like to thank R. Sandhu for providing us with a program to run the framework in \cite{San09} and for his invaluable help using this program.
\end{acknowledgements}

\bibliographystyle{spmpsci}      
\bibliography{Rother}   

\clearpage
\input ComputingSummaries
\input ResultsExtra

\end{document}

%% file: Abstract.tex
\begin{abstract}
This article presents a mathematical framework to \emph{simultaneously} tackle the problems of 3D reconstruction, pose estimation and object classification, from \emph{a single 2D image}.
In sharp contrast with state of the art methods that rely primarily on 2D information and solve each of these three problems separately or iteratively, we propose a mathematical framework that incorporates prior ``knowledge" about the 3D shapes of different object classes and solves these problems jointly and simultaneously, using a hypothesize-and-bound (H\&B) algorithm \cite{Rot10}.

In the proposed H\&B algorithm one hypothesis is defined for each possible pair [object class, object pose], and the algorithm selects the hypothesis $H$ that maximizes a function $L(H)$ encoding how well each hypothesis ``explains" the input image. To find this maximum efficiently, the function $L(H)$ is not evaluated \emph{exactly} for each hypothesis $H$, but rather upper and lower bounds for it are computed at a much lower cost.
In order to obtain bounds for $L(H)$ that are tight yet inexpensive to compute, we extend the theory of shapes described in \cite{Rot10} to handle projections of shapes. This extension allows us to define a probabilistic relationship between the prior knowledge given in 3D and the 2D input image. This relationship is derived from first principles and is proven to be the only relationship having the properties that we intuitively expect from a ``projection."

In addition to the efficiency and optimality characteristics of H\&B algorithms, the proposed framework has the desirable property of integrating information in the 2D image \emph{with} information in the 3D prior to estimate the optimal reconstruction.
While this article focuses primarily on the problem mentioned above, we believe that the theory presented herein has multiple other potential applications.

\keywords{3D reconstruction \and pose estimation \and object classification \and shapes \and shape priors \and hypothesize-and-verify \and coarse-to-fine \and probabilistic inference \and graphical models \and image understanding}

\end{abstract}

%% file: Introduction.tex
\section{Introduction}
\label{sec:Introduction}

It is in general easy for humans to ``perceive" three dimensional (3D) objects, even when presented with a two dimensional (2D) image alone. This situation in which one does not rely on binocular (stereo) vision to ``perceive" 3D commonly arises when one is closing an eye, looking at a picture or screen, or simply processing objects in the monocular parts of the visual field. The ability to ``perceive" the world in 3D is essential to interact with the environment and to ``understand" the observed images, which arguably is only achieved when the underlying 3D structure of a scene is understood.
By trying to design machines that replicate this ability, we come to appreciate the tremendous complexity of this problem and the marvelous proficiency of our own visual system.

\subsection{In defense of 3D ``knowledge"}
This problem, namely 3D reconstruction from a single 2D image, is inherently ambiguous when no other constraints are imposed. Humans presumably solve it by relying on knowledge about the principles underlying the creation of images (exploited in \S \ref{sec:ProjectionTerm}), knowledge about the specific object classes involved (see below), and knowledge about the laws of nature that govern the physical interactions between objects. 

While the knowledge about the object classes can take many forms, in this work \emph{we focus on 3D shape priors} (described in detail in \S \ref{sec:ShapePriorTerm}). This choice of a 3D representation of knowledge, as opposed to a 2D representation, has several important consequences.
First, it results in a framework that can handle images of an object from any viewpoint, rather than an arbitrary viewpoint included in the training dataset. In other words, it makes the framework \emph{viewpoint independent}. Second, it allows us to store the knowledge ``efficiently," because there is no need to store specific information for each viewpoint. Instead, the same (common) ``knowledge" is used for any viewpoint. Third, because this knowledge is \emph{common} to all the viewpoints, the number of training examples required to acquire it is reduced. And fourth, the fact that the prior knowledge is represented in 3D allows us to easily impose constrains on the physical interactions between objects that would be very hard to impose otherwise (\eg, that an object is resting on a supporting plane, not ``flying" above it).

\subsection{In defense of synergy}
\label{sec:Synergy}

In order to incorporate prior ``knowledge" about an object in its 3D reconstruction, in particular prior knowledge about its 3D shape, it is necessary to classify the object and estimate its location/pose. On the one hand, classifying the object is essential to be able to consider only the \emph{specific} prior knowledge about the class of the \emph{particular} object in the image, among the possibly vast amount of available \emph{general} knowledge. This specific knowledge is particularly important to reconstruct the parts of the object that are not visible in the image (\eg, those that are occluded), since in this case the prior knowledge might be the only source of information. On the other hand, estimating the pose of the object is necessary to incorporate the prior geometric knowledge in the right spatial locations. For example, knowing that an object is a `mug,' tells us that it is expected to have a handle somewhere; knowing also the \emph{pose} of the object tells us \emph{where} the handle is supposed to be. Similarly to classify the object and estimate its pose, it is helpful to rely on the information in its 3D reconstruction.

This suggests that all these problems (3D reconstruction, object classification and pose estimation) are intimately related, and hence it might be advantageous to solve them simultaneously rather than in any particular order. For this reason, in this paper we \emph{simultaneously} address the problems of 3D reconstruction, object classification and pose estimation, from a single 2D image. For simplicity, we will restrict our attention to cases in which the input image is known to contain a \emph{single} object from one of several \emph{known} classes.

\subsection{In defense of shape}
Despite the fact that \emph{appearance} is often  a good indicator of class identity (see the large body of work relying on local features for recognition \cite{Rot083}), there are cases in which shape might be a more informative cue. For example, there are cases in which the appearance of the objects in a class is too variable and/or unrelated to the class identity to be of any use, while their (3D) shape is well preserved (\eg, consider the class `mugs').

Moreover, shapes can often be extracted very reliably from videos (\eg, for the important class of fixed camera surveillance videos). For example, using existing \emph{background modelling} techniques (\eg, \cite{Mit04,DLT03}), it is possible to compute a \emph{foreground probability image} encoding the shape of the foreground object in the image. This image contains, at each pixel, the probability that the foreground object is seen at that pixel. Thus this image contains only the ``shape information" while all the ``appearance information" has been ``filtered out." This foreground probability image, which is one of the inputs to our system, can be computed in different ways, and our algorithm is independent of the particular algorithm used to compute it.

Therefore, though appearance and shape cues are in general complementary, for concreteness, \emph{we only consider shape cues. Appearance is only considered in a preprocessing step (\ie, not as part of our framework) to compute the foreground probability images.} Hence, since this framework does not rely on detecting local features, it can handle featureless objects, or complement other approaches that do use local features.

\subsection{The general inference framework}
In order to solve the problems mentioned above \emph{simultaneously}, while exploiting shape cues and prior 3D knowledge about the object classes, we define a probabilistic graphical model encoding the relationships \linebreak[4] among the variables: class $K$, pose $T$, input image $f$, and 3D reconstruction $v$ (described in detail in \S \ref{sec:ProblemDefinition}). Because of the large number of variables in this graphical model (the dimensions of $f$ and $v$ are very high), and due to the existence of a huge number of loops among them, standard inference methods are either very inefficient or not guaranteed to find the optimal solution.

For this reason we solve our problem using the hypo\-the\-size-and-verify pa\-ra\-digm. In this paradigm one hypothesis $H$ is defined for every possible ``state of the world," and the goal is to select the hypothesis that best ``explains" the input image. In other words, the goal is to select the hypothesis $H_*$ that solves
\begin{equation}
H_* = \operatorname*{argmax}_{H \in \mathbb{H}} L(H),
\label{eq:HaVProblem}
\end{equation}
where $\mathbb{H}$ is the set of all possible hypotheses, referred to as the \emph{hypothesis space}, and $L(H)$ is a function, referred to as the \emph{evidence}, that quantifies how well each hypothesis ``explains" the input (better hypotheses produce higher values). This evidence is derived from the system's joint probability, which is obtained from the graphical model mentioned above.

In the specific problem addressed in this article the hypothesis space $\mathbb{H}$ contains every hypothesis $H_{ij}$ defined by every possible object class $K_i$, and by every possible object pose $T_j$ (\ie, $H_{ij} \triangleq (K_i, T_j)$). By selecting the hypothesis $H_{ij}$ that solves \eqref{eq:HaVProblem}, the hypothesize-and-verify approach simultaneously es\-ti\-mates the class $K_i$ and the pose $T_j$ of the object in the image. As we shall later see, the 3D reconstruction $v$ is estimated during the computation of the evidence.
Since the number of hypotheses in the set $\mathbb{H}$ is potentially very large, it is essential to evaluate $L(H)$ very efficiently. For this purpose we introduced in \cite{Rot10} a class of algorithms to efficiently implement the hypothesize-and-verify paradigm. This class of algorithms, known as \emph{hypothesize-and-\linebreak[4]bound}, is described next.

\subsection{Hypothesize-and-bound algorithms}
\label{sec:SCAlgorithms}
\emph{Hypothesize-and-bound} (H\&B) algorithms have two \linebreak[4] parts. The first part consists of a \emph{bounding mechanism} (BM) to compute lower and upper bounds, $\underline{L}(H)$ and $\overline{L}(H)$, respectively, for the evidence $L(H)$ of a hypothesis $H$. These bounds are in general much cheaper to compute than the evidence itself, and are often enough to discard many hypotheses (note that a hypothesis $H_1$ can be safely discarded if $\overline{L}(H_1) < \underline{L}(H_2)$ for some other hypothesis $H_2$). On the other hand, these bounds are not as ``precise" as the evidence itself, in the sense that they only define an \emph{interval} $[\underline{L}(H), \overline{L}(H)]$ where the evidence for a hypothesis is guaranteed to be. Nevertheless, the width of this interval (or \emph{margin}) can be made as small as desired by investing additional computational cycles into the refinement of the bounds. In other words, given a number of computational cycles to be spent on a hypothesis, the BM returns an interval on the real line where the evidence for the hypothesis is guaranteed to lie. If additional computational cycles are later allocated to the hypothesis, the BM permits to efficiently refine the bounds defining this interval.

The second part of an H\&B algorithm is a \emph{focus of attention mechanism} (FoAM) to sensibly and dynamically allocate the available computational resources \linebreak[4] among the different hypotheses whose bounds are to be refined. Initially the FoAM calls the BM to compute rough and cheap bounds for each hypothesis. Then, during each iteration, the FoAM selects one hypothesis and calls the BM to refine its bounds. This process continues until either a hypothesis is proved optimal, or a group of hypotheses cannot be further refined or discarded (these hypotheses are said to be \emph{indistinguishable} given the current input). Such a hypothesis, or group of hypotheses, maximizes the evidence \emph{regardless of the exact values of all the evidences (which do not need to be computed)}.
Interestingly, the total number of computational cycles spent depends on the order in which the bounds are refined. Thus this order is carefully chosen by the FoAM to minimize the total computation. The FoAM is explained in greater detail in \cite[\S 3]{Rot10}.

H\&B algorithms are general optimization procedures that can be applied to many different problems. To do so, however, a different evidence and a different BM has to be developed for each particular problem (the same FoAM, on the other hand, can be used in every problem). To develop a BM for the current problem, in \S \ref{sec:ProjectionTerm} we extend the theory of shapes presented in \cite[\S 5]{Rot10}. Understanding this theory will be essential to follow the derivations in later sections.

\subsection{Organization of this article}
The remainder of this paper is organized as follows. In \S \ref{sec:PriorWork} we place the current work in the context of prior relevant work. In \S \ref{sec:ProblemDefinition} we formally define the problem to be solved, defining the evidence $L(H)$ for it. Then in \S \ref{sec:LowerBound} and \S \ref{sec:UpperBound} we derive formulas to compute lower and upper bounds for $L(H)$, respectively, and in \S \ref{sec:BoundingMechanism} we describe how to implement the BM using these formulas. After that we summarize the proposed framework in \S \ref{sec:MethodSummary}, present experimental results obtained with it in \S \ref{sec:Results}, and conclude in \S \ref{sec:Conclusions} with a discussion of key contributions and directions for future research. Additional details, such as proofs of the theoretical results stated in the paper, are included in the supplementary material.

%% file: PriorWork.tex
\section{Prior work}
\label{sec:PriorWork}

As mentioned in \S \ref{sec:Introduction}, in this work we focus on the problem of simultaneous 3D reconstruction, pose estimation and object classification, from a single foreground probability image. Since in the absence of other constraints this problem is ill posed, approaches to solve it must rely on some form of prior knowledge about the possible classes of the object to be reconstructed. These approaches, in general, differ on the representation used for the reconstruction, the encoding scheme used for the prior knowledge, and the procedure to obtain the solution from the input image and the prior knowledge.

Savarese and Fei-Fei \cite{Sav07}, for example, proposed an approach to simultaneously classify an object, estimate its pose, and obtain a crude 3D reconstruction from a single image. The 3D reconstruction consists of a few planar faces (or ``parts") linked together by homographic transformations. The object class' prior knowledge is encoded in this case by the appearance descriptor of the parts and by the homographic transformations linking them. Saxena \etal \cite{Sax08} and Hoiem \etal \cite{Hoi05}, on the other hand, focus on the related problem of scene reconstruction from a single image. In these works a planar patch in the reconstructed surface is defined for each superpixel in the input image. The 3D orientation of these patches is inferred using a learned probabilistic graphical model that relates these orientations to features of the corresponding superpixels. Prior knowledge in this case is encoded in the learned relationship between superpixel features and patch 3D orientations. In contrast with our approach, these approaches rely on the appearance of the object (or scene), which as previously mentioned, can be highly variable for some object classes.

The use of 3D shape information (or ``geometry"), on the other hand, has a long tradition in computer vision \cite{Mun06}. Since the early days many methods have been proposed for the reconstruction and pose estimation of ``well defined" object classes from a single image. However, requiring object classes to be ``well defined" often resulted in methods that dealt with somewhat artificial object classes, not frequently found in the real world (\eg, polyhedral shapes \cite{Rob65} and generalized cylinders \cite{Bin71}). In general, these methods proceed by extracting geometric features (\eg, corners and edges) from an image, grouping these features to form hypotheses, and then validating these hypotheses using geometric constraints. 
One problem with these methods is that it is difficult to extend them to handle classes of real objects which might be very complex and might not contain geometric features at all. 
A second problem with these methods is that they could be very sensitive to erroneously detected features, due to their lack of reliance on statistical formulations.

More recently a number of other methods for 3D reconstruction from a single image have been proposed for \emph{specific} object classes (\eg \cite{Bow98,Sig06}). In general these methods consist of a parametric model of the object class to be represented and a procedure to find the best fit between the projection of the model and the input image. 
Prior knowledge in this case is encoded in the design of the model (\eg, which parts an articulated model has, and how they are connected).
Object classes that have been modeled in this way include trees/grasses \cite{Han03} and people \cite{Wan03}. 
Model-based approaches are best suited to reconstruct objects of the particular class they were designed for and are difficult to extend beyond this class, since the model is typically designed \emph{manually} for that particular class.

In contrast, more general representations that can learn about a class of objects from exemplars (as our approach does), can be trained on new classes without having to redesign the representation anew each time. One example of such a general representation can be found in the work of Sandhu \etal \cite{San09}, which uses a level set formulation coupled with shape priors to segment an object in a single image and estimate its pose. The prior shape knowledge is learned for an object class from a set of training exemplars of the class. To construct the shape prior Sandhu \etal compute the signed distance function (SDF) for each 3D shape in the training set, and then learn the principal components of this set of SDFs.

While we consider this work to be the most similar to ours regarding its goals, the two approaches have two major differences. First, that work is not guaranteed to find the global optimum, as our approach does, but only a local optimum that critically depends on the initial condition (this is further discussed in \S \ref{sec:PerformanceAssesment}). Second, Sandhu \etal do not address the tasks of classification or 3D reconstruction. While it could appear that that work could be modified to handle these tasks, we argue that these modifications are not trivial. For example, a \emph{3D reconstruction} could be computed from the linear combination of SDFs estimated by that framework. This is not trivial, however, because a linear combination of SDFs is not itself a SDF. Similarly the \emph{class} could be estimated by considering a mixture model, or simply running the framework multiple times with different priors and keeping the best solution. However, since the framework has no optimality guarantees, this would make the method even more prone to get stuck in a local optimum.

Hence, to the best of our knowledge there are no other works focusing on exactly the same problem, except our own work in \cite{RotICCV09} of which the current work is a formalization and extension. There are two major differences between these two works. The first difference is that the segmentation and the 3D reconstruction considered in the current work are continuous, while in \cite{RotICCV09} they are discrete. This allows us to compute in this work tighter bounds for the evidence $L(H)$, based on the theory of shapes described in \cite{Rot10}. The second difference is that the current model corrects a bias discovered in the model in \cite{RotICCV09}. Given two similar 3D shapes with equal projection on the camera plane, this bias made the framework in \cite{RotICCV09} select the shape that was further away from the camera. The current model, on the other hand, has no preference for either shape. This is explained in detail in \S \ref{sec:ProjectionTerm}.

%% file: ProblemDefinition.tex
\section{Problem formulation}
\label{sec:ProblemDefinition}

\begin{figure}
\begin{center}
\includegraphics[width=0.90\columnwidth]{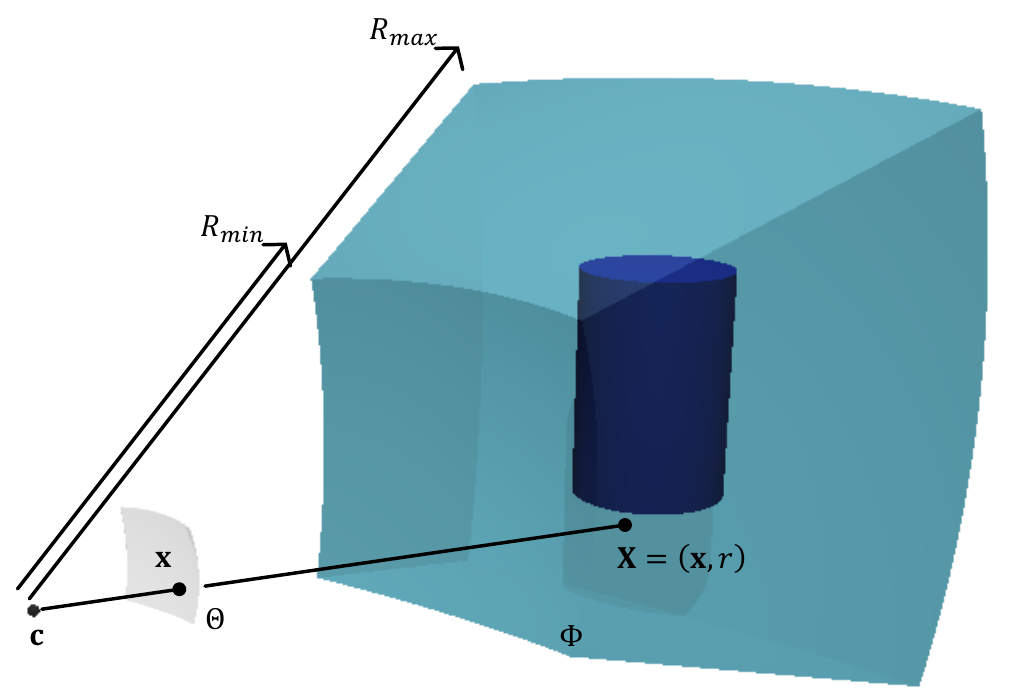}
\end{center}
\vspace{-10pt}
\caption{Setup for the problem at hand. The camera is defined by the camera center $\vec{c}$ and a patch $\Theta$ in the unit sphere ($\Theta \subset S^2$). The world set $\Phi$ is defined as those 3D points that project to $\Theta$ and whose distance to the camera center is in the interval $[R_{min},R_{max}]$. Any 3D point $\vec{X} \in \Phi$ projects to a single point $\vec{x}$ in the camera retina. A single object (represented by the blue cylinder) is assumed to be in $\Phi$.}
\vspace{-15pt}
\label{fig:Setup}
\end{figure}

\begin{figure} \sidecaption
\vspace{0pt}
\includegraphics[width=0.45\columnwidth]{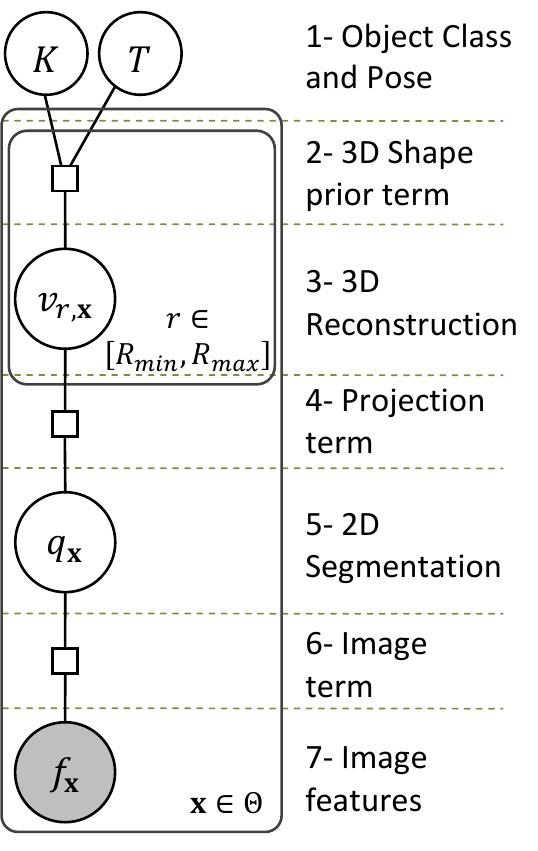}
\vspace{0pt}
\caption{Factor graph proposed to solve the problem. A \emph{factor graph}, \cite{Bis06}, has a \emph{variable node} (circle) for each variable, and a \emph{factor node} (square) for each factor in the system's joint probability. Factor nodes are connected to the variable nodes of the variables in the factor. Observed variables are shaded. A plate indicates that there is an instance of the nodes in the plate for each element in a set (indicated on the lower right). The plates in this graph hide the existing loops.}
\label{fig:FactorGraph}
\vspace{-10pt}
\end{figure}

In this section we formally define the problem of joint classification, pose estimation and 3D reconstruction from a single 2D image, which we alluded to in previous sections. This problem is defined as follows.
Let $f:\Theta \to \mathbb{R}^c$ ($c \in \mathbb{N}$) be a 2D image of $c$-dimensional ``features" produced as the noisy 2D projection of a single 3D object (Fig. \ref{fig:Setup}). This object is assumed to belong to a class from a set of known classes. Given this input image $f$ and the 3D shape priors (defined later), our problem is to estimate the class $K$ of the object, its pose $T$, recover a 2D segmentation $q$ of the object in the image, and estimate a 3D reconstruction $v$ of the object in 3D space. The relationships among these variables are depicted in the factor graph of Fig. \ref{fig:FactorGraph}.

In order to estimate $q$, $v$ and the hypothesis $H \triangleq (K,T)$ given the observations $f$, we use the maximum a posteriori estimator \cite[Chapter 11]{Kay93}. Thus we find the mode of the posterior distribution, which is given by the product of the likelihood function, $P(f|q,v,H)$, and the prior distribution, $P(q,v,H)$. From the independence assumptions depicted in Fig. \ref{fig:FactorGraph} the posterior distribution is equal, up to a constant, to
\begin{align}
P(f|q,v,H) P(q,v,H) \! = \! P(f|q) P(q|v) P(v|H) P(H).
\label{eq:JointProbability}
\end{align}
Our goal can now be stated as finding the values of $K$, $T$, $q$, and $v$, that maximize \eqref{eq:JointProbability}. In doing so we would be estimating $K$, $T$, $q$, and $v$, \emph{simultaneously}.

Before we formally define each one of the terms on the rhs of \eqref{eq:JointProbability}, in \S \ref{sec:ContinuousShapesReview} we briefly review the theory of shapes introduced in \cite{Rot10}.
In \S \ref{sec:ImageTerm}, \S \ref{sec:ShapePriorTerm} and \S \ref{sec:ProjectionTerm} we use this theory to formally define $P(f|q)$, $P(q|v)$ and $P(v|H)$, respectively.
We conclude the section by putting these terms together to obtain an expression for the evidence $L(H)$ of a hypothesis $H$, which is closely related to the posterior distribution in \eqref{eq:JointProbability}.

\subsection{Review: continuous and discrete shapes and their likelihoods}
\label{sec:ContinuousShapesReview}

In the previous section we mentioned the 2D segmentation $q$ and the 3D reconstruction $v$ without explicitly defining a specific representation for them. These entities are two instances of what we call \emph{continuous shapes}, as defined next.

\begin{definition}[Continuous shape]
\label{def:ContinuousShape}
Given a set $\Omega \subset \mathbb{R}^d$, a set $s \subset \Omega$ is a \emph{continuous shape} if: 1) it is open, and 2) its boundary has zero measure. Alternatively, a continuous shape $s$ can also be regarded as the function $s:\Omega \to \{0, 1\}$ defined by
\begin{equation}
s(\vec{x}) \triangleq \left \{
\begin{array}{ll}
1, & \quad \text{if} \ \vec{x} \in s, \\
0, & \quad \text{otherwise}.
\end{array} \right.
\end{equation}
\end{definition}

In order to define the terms $P(f|q)$ and $P(v|H)$ involving the continuous shapes $q$ and $v$ in \eqref{eq:JointProbability}, we define the likelihood of a continuous shape $s$ by extending the definition of the likelihood of a \emph{discrete shape} $\hat{s}$, defined next.

\begin{definition}[Discrete shape]
\label{def:DiscreteShape}
Given a partition \linebreak[4] $\Pi(\Omega) \!= \! \{ \Omega_1, \dots, \Omega_n \}$ of a set $\Omega \! \subset \! \mathbb{R}^d$ (\ie, a collection of sets such that $\bigcup_i{\Omega_i} \!= \!\Omega$, and $\Omega_i \cap \Omega_j \! = \! \emptyset$ for $i \!\ne \!j$) the \emph{discrete shape} $\hat{s}$ is defined as the function $\hat{s}: \Pi(\Omega) \to \{0,1\}$.
\end{definition}
Notice that a continuous shape $s$ can be \emph{produced} from a discrete shape $\hat{s}$ (denoted as $s \sim \hat{s}$) as $s(\vec{x}) = \hat{s}(\Omega_i)$, for all $\vec{x} \in \Omega_i$ and $i = 1, \dots, n$. Discrete shapes will be used in \S \ref{sec:LowerBound} to approximate continuous shapes and to compute lower bounds for their likelihoods.

\begin{definition}[Bernoulli field]
\label{def:BernoulliField}
A \emph{discrete Bernoulli Field} (BF) is a family of independent Bernoulli random variables $\hat{B} = \{\hat{B}_1, \dots, \hat{B}_n \}$ characterized by the success rates $p_{\hat{B}}(i) \triangleq P(\hat{B}_i = 1)$.
The log-likelihood of the discrete shape $\hat{s}$ according to the discrete BF $\hat{B}$ is then computed as
\vspace{-5pt}
\begin{align}
\log{P(\hat{B} = \hat{s})} & \triangleq \sum_{i=1}^n{\log{P(\hat{B}_i = \hat{s}(\Omega_i))}}.
\label{eq:DiscreteBF}
\end{align}

For a discrete BF $\hat{B}$, we define the \emph{constant term} $Z_{\hat{B}}$ and the \emph{logit function} $\delta_{\hat{B}}(i)$ to be,
\begin{align}
Z_{\hat{B}} & \triangleq \sum_{i=1}^n{\log{(1 - p_{\hat{B}}(i))}}, \ \text{and} \\
\delta_{\hat{B}}(i) & \triangleq \log{\left(\frac{p_{\hat{B}}(i)} {1 - p_{\hat{B}}(i)} \right)},
\end{align}
respectively. Then it can be shown that \eqref{eq:DiscreteBF} can be rewritten as
\vspace{-5pt}
\begin{align}
\log{P(\hat{B} = \hat{s})} = Z_{\hat{B}} + \sum_{i=1}^n{ \hat{s}(\Omega_i) \delta_{\hat{B}}(i)}.
\label{eq:DiscreteShapeLikelihood}
\end{align}

In order to compute the likelihood of a continuous shape, we first define \emph{continuous} BFs by analogy with \emph{discrete} BFs. Similarly to those, a \emph{continuous BF} is also a collection of independent Bernoulli random variables. In a continuous BF $B$, however, one variable $B(\vec{x})$ is defined for each point $\vec{x} \in \Omega$, rather than for each index $i \in \{1, \dots, n \}$. The success rates for the variables in the BF are given by the function $p_B(\vec{x}) \triangleq P(B(\vec{x}) = 1)$. A continuous BF $B$ can be constructed from a discrete BF $\hat{B}$ by defining its success rates to be $p_{B}(\vec{x}) = p_{\hat{B}}(i)$, for all $\vec{x} \in \Omega_i$ and for $i = 1, \dots, n$. This continuous BF is said to be \emph{produced} from the discrete BF and is denoted as $B \sim \hat{B}$.
By analogy with \eqref{eq:DiscreteBF} we define the ``log-likelihood" of a continuous shape $s$ according to a continuous BF $B$ as
\begin{align}
\log{P(B = s)} & \triangleq \frac{1}{u_\Omega} \int_\Omega{\log{P(B(\vec{x}) = s(\vec{x}))} \ d\vec{x}},
\label{eq:ContinuousLikelihood}
\end{align}
where $u_\Omega$ is a constant to be described later, referred to as the \emph{equivalent unit size}.

If we now define the constant term $Z_{B}$ and the logit function $\delta_{B}$ for the BF $B$, respectively as
\begin{align}
Z_{B} & \triangleq \int_{\Omega}{\log{(1 - p_B(\vec{x}) ) }} \ d\vec{x}, \quad \text{and} \\
\delta_{B}(\vec{x}) & \triangleq \log{\left( \frac{p_B(\vec{x})} {1 - p_B(\vec{x}) } \right)},
\end{align}
the likelihood in \eqref{eq:ContinuousLikelihood} can be rewritten as
\begin{align}
\log{P(B = s)} = \frac{1}{u_\Omega} \left[Z_B + \int_\Omega{ s(\vec{x}) \delta_B(\vec{x})} \right].
\label{eq:ContinuousShapeLikelihood}
\end{align}
\end{definition}

The following proposition shows that, under certain conditions, the continuous and discrete ``log-likelihoods" in \eqref{eq:ContinuousShapeLikelihood} and \eqref{eq:DiscreteShapeLikelihood}, respectively, coincide. For this reason we said that \eqref{eq:ContinuousShapeLikelihood} extends \eqref{eq:DiscreteShapeLikelihood}.

\begin{proposition}[Relationship between likelihoods of continuous and discrete shapes]
Let $\Pi(\Omega)$ be a partition of a set $\Omega$ such that $|\omega| = u_\Omega \ \forall \omega \in \Pi(\Omega)$, and let $\hat{B}$ and $\hat{s}$ be a discrete BF and a discrete shape, respectively, defined on $\Pi(\Omega)$. Finally, let $B$ be a continuous BF and let $s$ be a continuous shape such that $B \sim \hat{B}$ and $s \sim \hat{s}$. Then, the log-likelihoods of the continuous and discrete shapes are equal, \ie,
\begin{align}
\log{P(B = s)} = \log{P(\hat{B} = \hat{s})}.
\end{align}
\begin{proof}
\hspace{-4pt}: Immediate from the definitions. \qed
\end{proof}
\end{proposition}

Note that the equivalent unit size $u_\Omega$ ``scales" the value in brackets in \eqref{eq:ContinuousShapeLikelihood} according to the resolution of the partition $\Pi(\Omega)$, making it comparable to \eqref{eq:DiscreteShapeLikelihood}. Now we are ready to define the terms on the rhs of \eqref{eq:JointProbability}. In the following, when no ambiguity is possible, we will abuse notation and write $P(s)$ or $P(s(\vec{x}) \! = \! 1)$ instead of $P(B \! = \! s)$ and $P(B(\vec{x}) \! = \! 1)$, respectively.

\subsection{Image term: $\log{P(f|q)}$}
\label{sec:ImageTerm}
The 2D segmentation $q$ that we want to estimate (level 5 of Fig. \ref{fig:FactorGraph}) is represented as a 2D continuous shape defined on the image domain $\Theta$. This segmentation $q$ states whether each point $\vec{x} \in \Theta$ is deemed by the framework to be in the \texttt{Background} (if $q(\vec{x})=0$) or in the \texttt{Foreground} (if $q(\vec{x})=1$).

We assume that the state $q(\vec{x})$ of a point $\vec{x} \in \Theta$ \emph{cannot} be observed directly, but rather that it defines the pdf of a feature $f(\vec{x})$ at $\vec{x}$ that \emph{is} observed. For example, $f(\vec{x})$ could simply indicate color, depth, the output of a classifier or in general any feature directly observed at the point $\vec{x}$, or computed from other features observed in the neighborhood of $\vec{x}$.
Moreover, we suppose that if a point $\vec{x}$ belongs to the \texttt{Background}, its feature $f(\vec{x})$ is distributed according to the pdf $p_{\vec{x}}(f(\vec{x}) | q(\vec{x})=0)$, while if it belongs to the \texttt{Foreground}, $f(\vec{x})$ is distributed according to $p_{\vec{x}}(f(\vec{x}) | q(\vec{x})=1)$. This feature $f(\vec{x})$ is assumed to be independent of the feature $f(\vec{y})$ and the state $q(\vec{y})$ at every other point $\vec{y} \in \Theta$, given $q(\vec{x})$.

The subscript $\vec{x}$ in $p_{\vec{x}}$ was added to emphasize the fact that a \emph{different} pdf \emph{might} be used for every point. In other words, it could be that $p_{\vec{x}}(f_0 | q_0) \ne p_{\vec{y}}(f_0 | q_0)$ if $\vec{x} \ne \vec{y}$ and $f_0$ and $q_0$ are two arbitrary values of $f$ and $q$, respectively. For example, in the experiments of \S \ref{sec:Results} a different pdf $p_{\vec{x}}(f(\vec{x}) | q(\vec{x})=0)$ was learned for every point $\vec{x}$ in the background. In this case \emph{multiple} Gaussian pdf's were used (one per point). On the other hand a single pdf, $p(f(\vec{x}) | q(\vec{x})=1)$, a mixture of Gaussians, was used for all the points in the foreground (the same one for all the points in the foreground).

Then from \eqref{eq:ContinuousLikelihood} the conditional ``log-density" of the observed features $f$, given the 2D shape $q$, that we refer to as the \emph{image term}, is given by
\begin{equation}
\log{P(f|q)} \triangleq \frac{1}{u_{\Theta}} \int_{\Theta}{\log{p_{\vec{x}}(f(\vec{x}) | q(\vec{x}))} \ d\vec{x}},
\label{eq:RelationshipFandQ}
\end{equation}
where the equivalent unit size in $\Theta$, $u_{\Theta}$, is a constant to be fixed. Defining the continuous BF $B_f$ with success rates 
\begin{equation}
p_{B_f}(\vec{x}) \! \triangleq \! \frac{p_{\vec{x}} \left(f(\vec{x}) | q(\vec{x}) \!= \!1 \right)} {p_{\vec{x}}\left(f(\vec{x})|q(\vec{x}) \!=\!0 \right) + p_{\vec{x}}\left(f(\vec{x})|q(\vec{x})\!=\!1 \right)},
\label{eq:BernoulliFieldInput}
\end{equation}
it follows that \eqref{eq:RelationshipFandQ} is equal, up to a constant, to the ``log-likelihood" of the shape $q$ according to the BF $B_f$, \ie, $\log{P(f|q)} = \log{P(B_f = q)} + C_1$. Therefore, using \eqref{eq:ContinuousShapeLikelihood}, the image term can be written as
\begin{equation}
\log{P(f|q)} = \frac{1}{u_{\Theta}} \left[Z_{B_f} + \int_{\Theta}{q(\vec{x}) \delta_{B_f}(\vec{x})\ d\vec{x}}\right] + C_1.
\label{eq:ImageTerm}
\end{equation}

\subsection{3D shape prior term: $\log{P(v|H)}$}
\label{sec:ShapePriorTerm}

While the segmentation $q$ is a \emph{2D continuous shape} on the 2D image domain $\Theta$, the reconstruction $v$ (that we also want to estimate) is a \emph{3D continuous shape} on the set $\Phi \subset \mathbb{R}^3$. This 3D reconstruction $v$ (level 3 of Fig. \ref{fig:FactorGraph}), states whether each point $\vec{X} \in \Phi$ is deemed by the framework to be \texttt{In} the reconstruction (if $v(\vec{X})=1$), or \texttt{Out} of it (if $v(\vec{X})=0$).
In this reconstruction the coordinates of each 3D point $\vec{X}$ are expressed in the \emph{world coordinate system} (WCS) defined on the set $\Phi$.

As mentioned before our problem of interest is ill posed unless some form of prior knowledge about the shape of the objects is incorporated. We assume that the object class $K$ (level 1 of Fig. \ref{fig:FactorGraph}) is one out of $N_K$ distinct possible object classes, each one characterized by a 3D shape prior $B_K$ encoding our prior geometric knowledge about the object class. This knowledge is stated with respect to an \emph{intrinsic 3D coordinate system} (ICS) defined for each class. In other words, all the objects of the class are assumed to be in a canonical (normalized) pose in this ICS. 
Each shape prior $B_K$ is encoded as a BF (also referred to as $B_K$), such that for each point $\vec{X}'$ in the ICS of the class, the success rate $p_{B_K}(\vec{X}') \triangleq P(v'(\vec{X}')=1 | K)$ indicates the probability that the point $\vec{X}'$ would be \texttt{In} the 3D reconstruction $v'$ defined in the ICS, given the class $K$ of the object. We assume that $p_{B_K}$ is zero everywhere, except (possibly) on a region $\Phi_K \subset \mathbb{R}^3$ called the \emph{support of the class} $K$.

Note that the shape prior $B_K$ and the 3D continuous shape $v'$ alluded to in the previous paragraph are defined in the ICS of the class. To obtain the corresponding entities in the WCS, we define the transformation $T:\mathbb{R}^3 \to \mathbb{R}^3$ that maps a point $\vec{X}'$ in the ICS to the point $\vec{X} \triangleq T(\vec{X}')$ in the WCS. This transformation is referred to as the \emph{pose} and is another unknown to be estimated (level 1 of Fig. \ref{fig:FactorGraph}).
The transformation $T$ relates the desired 3D reconstruction $v$ and the \emph{hypothesis} BF $B_H \triangleq B_{K,T}$ defined in the WCS, to the reconstruction $v'$ and the \emph{class} BF $B_K$ defined in the ICS.
Specifically the 3D reconstruction $v(\vec{X})$ and the success rates of $B_H$, $p_{B_H}(\vec{X}) \triangleq P(v(\vec{X})=1 | H)$ (level 2 of Fig. \ref{fig:FactorGraph}), are given by
\begin{align}
v(\vec{X}) & = v'(T^{-1} (\vec{X})) \\
p_{B_H}(\vec{X}) & = p_{B_K}(T^{-1} (\vec{X})).
\end{align}
The BF $B_H$ thus encodes the probability that a point $\vec{X}$ is \texttt{In} the reconstruction $v$. The \emph{support of the hypothesis} $H$ is given by $\Phi_H \triangleq T(\Phi_K)$.
 
Therefore, from \eqref{eq:ContinuousShapeLikelihood}, the conditional ``log-pro\-ba\-bi\-li\-ty" of the continuous shape $v$, according to the BF $B_H$, is given by
\begin{align}
\log{P(v|H)} &\triangleq \frac{1}{u_{\Phi}(T)} \int_{\Phi}{\log{P(B_H(\vec{X}) = v(\vec{X}))} \ d\vec{X}} = \nonumber \\
& \frac{1}{u_{\Phi}(T)} \left[Z_{B_H} + \int_{\Phi}{v(\vec{X}) \delta_{B_H}(\vec{X})\ d\vec{X}}\right] \!,
\label{eq:ShapePriorTerm}
\end{align}
where the equivalent unit size in $\Phi$, $u_{\Phi}(T)$, depends on the transformation $T$ and is defined below. We refer to \eqref{eq:ShapePriorTerm} as the \emph{shape prior term}.

In order to derive an expression for the equivalent unit size $u_{\Phi}(T)$, we want to enforce that the ``log-pro\-ba\-bi\-li\-ty" of two hypotheses corresponding to objects that are similar (\ie, that are related by a change of scale), and have the same projection on the camera retina, are equal. Enforcing this will prevent the system from having a bias towards either smaller objects closer to the camera, or bigger objects farther away from the camera. In Proposition \ref{prop:RatioUnitSizes} (in the supplementary material) we show that to accomplish this, the unit size $u_{\Phi}(T)$ must be of the form $u_{\Phi}(T) = |J(T)| / \lambda'$, where $|J(T)|$ is the Jacobian of the transformation $T$, and $\lambda' > 0$ is an arbitrary constant.

\subsection{Projection term: $\log{P(q|v)}$}
\label{sec:ProjectionTerm}

The segmentation $q$ and the reconstruction $v$, defined in previous sections, are certainly not independent, as we expect $q$ to be (at least) ``close" to the projection of $v$ on the camera retina. The \emph{projection term}, $\log{P(q|v)}$, encodes the ``log-probability" of obtaining a segmentation $q$ in the camera retina $\Theta$, given that a reconstruction $v$ is present in the space in front of the camera $\Phi$.
In order to define this term more formally, we first need to understand the relationship between the sets $\Theta$ and $\Phi$, encoded by the camera transformation.

The \emph{camera transformation} maps points from the 3D space $\Phi$ into the 2D camera retina $\Theta$ (Fig. \ref{fig:Setup}). For simplicity, we consider a spherical retina rather than a planar retina. In other words, the image domain $\Theta$ is a subset of the unit sphere ($\Theta \subset S^2$).
Given a point $\vec{c} \in {\mathbb R}^3$, referred to as the \emph{camera center}, a correspondence is established between points in $\Phi$ and $\Theta$: for each \emph{point} $\vec{x} \in \Theta$, the points in the set $R(\vec{x}) \! \triangleq \! \{\vec{X} \!\in\! \Phi: \vec{X}\!=\!\vec{c} \!+\! r \vec{x}, \ r \!\in\! [0,\infty) \}$ are said to \emph{project} to $\vec{x}$. This set is referred to as the \emph{ray} of $\vec{x}$. Considering this correspondence, we will often refer to points $\vec{X} \in \Phi$ by their projection $\vec{x}$ in $\Theta$ and their distance $r$ from the camera center, as in $\vec{X}\!=\!(\vec{x},r)$. The domain $\Phi$ is thus formally defined as the set of points in 3D world space that are visible in the input image $\Theta$ and are at a certain distance range from the camera center, \ie, $\Phi \! \triangleq \! \left\{(\vec{x},r) \in {\mathbb R}^3: \vec{x} \in \Theta, R_{min} \le r \le R_{max}\right\}$ (Fig. \ref{fig:Setup}).

As mentioned at the beginning of this section the shape $q$ is a ``projection" of the continuous 3D shape $v$ in $\Phi$ onto $\Theta$. In other words, the state $q(\vec{x})$ (\texttt{Foreground} or \texttt{Background}) of the shape $q$ at a point $\vec{x} \in \Theta$ only depends on the states of the shape $v$ (\texttt{In} or \texttt{Out}) in the ray $R(\vec{x})$, and not on the states of $v$ in other points of $\Phi$.
To emphasize this fact we will write $P\left(q(\vec{x})|v\right) = P\left(q(\vec{x})|v_{R(\vec{x})} \right)$, where $v_{R(\vec{x})}$ denotes the part of $v$ in $R(\vec{x})$ and is referred to as the \emph{shape $v$ in the ray $R(\vec{x})$}.
Note that $v_{R(\vec{x})}:R(\vec{x}) \to \{0,1\}$ is itself a 1D continuous shape defined by $v_{R(\vec{x})}(r) \triangleq v(\vec{c}+r\vec{x})$.

Then given a 3D continuous shape $v$ in $\Phi$ our goal is to define a BF $B_g$ for $q$ in $\Theta$ (level 4 of Fig. \ref{fig:FactorGraph}), by ``projecting" the 3D shape $v$ into $\Theta$.
For notational convenience, given a point $\vec{x} \in \Theta$, its ray $R(\vec{x})$ and the shape $v$ in this ray, $v_{R(\vec{x})}$, let us define the \emph{failure rate} of the BF $B_g$ as
\begin{equation}
g\left(v_{R(\vec{x})}\right) \triangleq P\left(q(\vec{x})=0 | v_{R(\vec{x})}\right).
\label{eq:ProjectionFunction}
\end{equation}
This function, referred to as a \emph{projection function}, encodes the probability of ``seeing'' the \texttt{Background} (\ie, \emph{not} the shape $v$) at $\vec{x}$ given the shape $v_{R(\vec{x})}$ in $R(\vec{x})$.

We could simply define a projection function to be 1 when the measure of the set $v_{R(\vec{x})}$ is strictly zero (recall that continuous shapes can also be considered as sets, in this case $v_{R(\vec{x})}$ is the set $\{\vec{x} \in R(\vec{x}): v(\vec{x}) = 1\}$), and to be 0 otherwise, yielding the \emph{natural projection function}
\vspace{-5pt}
\begin{equation}
g_{Natural}\left(v_{R(\vec{x})}\right) \triangleq \left \{ \begin{array}{ll}
1, & \ \text{if} \ |v_{R(\vec{x})}| = 0, \\ 
0, & \ \text{otherwise}. \end{array}
\right.
\label{eq:NaturalProjection}
\end{equation}
However, this projection function leads to solutions that are not desirable, since these solutions can ``explain" the parts of the image that are much more likely \texttt{Fore\-ground} than \texttt{Background} (\ie, those where $p_{\vec{x}}(f(\vec{x}) |$ $q(\vec{x}) \!=\!1) \gg p_{\vec{x}}(f(\vec{x}) | q(\vec{x}) = 0)$) by placing an infinitesimal amount of ``mass" in the parts of 3D space that, according to the shape prior, are much more likely to be \texttt{Out} of the reconstruction than \texttt{In} it  (\ie, where $P(v(\vec{X}) = 0|H) \gg P(v(\vec{X}) = 1|H)$. Note that because the amount of mass placed is infinitesimal, these solutions do not ``pay the price" of living in the unlikely parts of 3D space, but still ``collect the rewards" of living in the likely parts of the 2D image. In order to avoid these undesirable solutions, we derive next, from first principles, a new expression for projection functions. 

Intuitively we expect $g(v_{R(\vec{x})} )$ to be lower when the measure of the set $v_{R(\vec{x})}$ is larger
(\ie, the larger the part of the object that intersects the ray, the least likely it is to see the background through that ray). In other terms, given two shapes in the same ray $R$, $v^1_R$ and $v^2_R$, such that $v^1_R(r) \le v^2_R(r) \ \forall r \in [R_{min},R_{max}]$, we expect
\begin{equation}
g \left(v^1_R\right) \ge g \left(v^2_R\right).
\label{eq:ProjectionMonotonicity}
\end{equation}
This property of the projection function, referred to as \emph{monotonicity}, guarantees that reconstructions that are intuitively ``worse'' are assigned lower log-probabilities.

While there are many constrains that can be imposed to enforce monotonicity, we will require that projection functions satisfy the \emph{independence property} defined next. In addition to monotonicity, this will yield a simple form for the projection function that has an intuitive interpretation.

\begin{definition}[Independence]
\label{def:ProjectionIndependence}
Given two continuous shapes in the same ray $R$, $v^1_R$ and $v^2_R$, that are disjoint (\ie, $v^1_R \cap v^2_R = \emptyset$), a projection function $g$ is said to have the \emph{independence property} if
\begin{equation}
g(v^1_R\cup v^2_R) = g(v^1_R) g(v^2_R).
\label{eq:ProjectionIndependence}
\end{equation}
In words, \eqref{eq:ProjectionIndependence} states that the events that the background is occluded by one shape or the other are independent. It can be seen that \eqref{eq:ProjectionIndependence} implies \eqref{eq:ProjectionMonotonicity}.
\end{definition}

Next consider two 3D continuous shapes $v^1$ and $v^2$ related by a central dilation $T_{\vec{c}}$ of scale $S > 1$, whose center is at the camera center $\vec{c}$ (\ie, $v^1(\vec{X}) = v^2(T_{\vec{c}}(\vec{X}))$ and $T_{\vec{c}}(\vec{X}) \triangleq \vec{c}+S(\vec{X}-\vec{c})$).
These two shapes are \emph{similar} (\ie, they are equal up to a change of scale) and produce the same projection on the camera retina, even though $v^1$ is smaller and closer to the camera than $v^2$.
In this situation we would like the framework to be agnostic as to which shape is present in the scene. Otherwise the framework would be either biased towards smaller shapes that are closer to the camera, or towards larger shapes that are farther away from it. This behavior is enforced by requiring projection functions to be scale invariant, as defined next.

\begin{definition}[Scale invariance]
\label{def:ProjectionScaleInvariance}
Given any shape in a ray $v_R$, a projection function $g$ is said to be \emph{scale invariant} if
\begin{equation}
g(v_R(r)) = g(v_R(Sr)), \quad \forall S > 0.
\label{eq:ProjectionScaleInvariance}
\end{equation}
\end{definition}

The following proposition provides a family of functions that satisfy the desired requirements \eqref{eq:ProjectionIndependence} and \eqref{eq:ProjectionScaleInvariance}. 

\begin{proposition}[Form of the projection function]
\label{prop:FormProjectionFunction}
Let us denote by $1_{(u,w)}$ a shape in a ray that consists on the interval $(u,w)$, and let $g$ be an independent and scale invariant projection function that satisfies the condition
\begin{equation}
g(1_{(1,e)}) = e^{\alpha}, \qquad \text{for some} \ \alpha < 0. 
\label{eq:BorderCondition}
\end{equation}
Let $v$ be a 3D continuous shape, and let 
\begin{equation}
\ell_v(\vec{x}) \triangleq \int_0^{\infty}{\frac{v_{R(\vec{x})}(r)}{r} \ dr} = \int_0^{\infty}{\frac{v(\vec{c}+r\vec{x})}{r} \ dr}
\label{eq:DefLx}
\end{equation}
be a measure of the ``mass" in the ray $R(\vec{x})$. Then the projection function $g$ must have the form
\begin{equation}
\label{eq:FormProjectionFunction}
g \left(v_{R(\vec{x})}\right) = e^{\alpha \ell_v(\vec{x})}.
\end{equation}
\begin{proof}
\hspace{-4pt}: See proof in the supplementary material.\qed
\end{proof}
\end{proposition}
It is interesting to note in \eqref{eq:FormProjectionFunction} that the scalar quantity $\ell_v(\vec{x})$ summarizes the relevant characteristics of the 1D shape $v_{R(\vec{x})}$. For this reason we will abuse the notation and write $P(q(\vec{x}) | v_{R(\vec{x})}) = P(q(\vec{x}) | \ell_v(\vec{x}))$.
 
Note that the natural projection function defined in \eqref{eq:NaturalProjection} also satisfies the independence and scale invariance requirements defined before. Moreover, the natural projection function is an extreme of the family of functions defined in Proposition \ref{prop:FormProjectionFunction} (for $\alpha \to -\infty$). For the reasons previously described, however, and as we empirically observed, this function is not convenient for our purposes. 

Using Proposition \ref{prop:FormProjectionFunction} we can now write an expression for the projection term as
\begin{align}
\log{P(q(\vec{x}) | v )} &= \log{P(q(\vec{x}) | v_{R(\vec{x})} )} = \log{P(q(\vec{x}) | \ell_v(\vec{x}) )} \nonumber \\
& = 
\left\{
\begin{array}{ll}
\alpha \ell_v(\vec{x}), & \ \text{if} \ q(\vec{x})=0, \\ 
\log{ \left(1-e^{\alpha \ell_v(\vec{x})}\right)}, & \ \text{if} \ q(\vec{x})=1.
\end{array}
\right.
\label{eq:ProjectionTermForm}
\end{align}
Therefore, using \eqref{eq:ContinuousLikelihood}, the projection term is given by
\begin{equation}
\log{P(q|v)} \triangleq \frac{1}{u_{\Theta}} \int_{\Theta}{\log{P\left(q(\vec{x}) | \ell_v(\vec{x}) \right)} \ d\vec{x}},
\label{eq:ProjectionTerm}
\end{equation}
where $u_{\Theta}$ is the equivalent unit size in $\Theta$ (the same that appeared in \eqref{eq:ImageTerm} before).

The definition of the projection function in \eqref{eq:FormProjectionFunction} contrasts with the choice made in \cite{RotICCV09}, where shift invariance rather than scale invariance was imposed. That choice biases the decision between two similar shapes with equal projection on the camera retina towards the shape that is further away from the camera.

\subsection{Definition of the evidence $L(H)$}
\label{sec:EvidenceDefinition}

In previous subsections we derived expressions for each of the terms in the system's posterior distribution given in \eqref{eq:JointProbability}. In this section we put them together to find an expression for the evidence $L(H)$ of a hypothesis $H$. This is the expression that the system will optimize.

Substituting the expressions for the image term, $\log{P(f|q)}$, the shape prior term, $\log{P(v|H)}$, and the projection term, $\log{P(q|v)}$ (given by \eqref{eq:ImageTerm}, \eqref{eq:ShapePriorTerm}, and \eqref{eq:ProjectionTerm}, respectively) into \eqref{eq:JointProbability}, the log-posterior is given by
\begin{align}
\log{P(f|q,v,H)} + \log{P(q,v,H)}= \log{P(H)} + \nonumber \\
\frac{1}{u_{\Theta}} \left[Z_{B_f} + \int_{\Theta}{q(\vec{x}) \delta_{B_f}(\vec{x}) \ d\vec{x}} \right] + C_1 + \nonumber \\
\frac{1}{u_{\Theta}} \int_{\Theta} {\log{P(q(\vec{x}) | \ell_v(\vec{x}))} \ d\vec{x}} + \nonumber \\
\frac{\lambda'}{|J(T)|} \left[Z_{B_H} + \int_{\Phi}{v(\vec{X}) \delta_{B_H}(\vec{X})\ d\vec{X}}\right] \!.
\label{eq:LPrime}
\end{align}

Our goal can now be formally stated as solving \linebreak[4] $\mathop{\sup}_{q,v,H} \big[\log{P(f|q,v,H)} + \log{P(q,v,H)}\big]$, which is \linebreak[4] equivalent to solving $\mathop{\max}_{H} L'(H)$, with
\begin{align}
L'(H) \triangleq \mathop{\sup}_{q, v} \Big[\log{P(f|q,v,H)} + \log{P(q,v,H)}\Big]
\label{eq:PreEvidence}
\end{align}
(the \emph{supremum} is used in optimizations over $q$ and $v$ since the set of continuous shapes might not contain a greatest element). However, instead of computing \eqref{eq:PreEvidence} directly, we will first derive an expression that is equal to it (up to a constant and a change of scale), but is simpler to work with.

In order to derive this expression, we disregard the terms $C_1$ and $Z_{B_f} / u_{\Theta}$ that do not depend on $H$, $q$, or $v$, disregard the term $\log{P(H)}$ that is assumed to be equal for all hypotheses, rearrange terms, multiply by $u_{\Theta}$, define $\lambda \triangleq \lambda' u_{\Theta}$ and substitute \eqref{eq:LPrime} into \eqref{eq:PreEvidence}, to obtain the final expression for the evidence,
\begin{align}
L(H) \triangleq & \mathop{\sup}_{q, v} \bigg\{
\int_{\Theta}{\bigg[q(\vec{x}) \delta_{B_f}(\vec{x}) + \log{P(q(\vec{x}) | \ell_v(\vec{x}))}} + \nonumber \\ 
& \frac{\lambda}{|J(T)|} \int_{R_{min}}^{R_{max}}{r^2 v(\vec{x},r) \delta_{B_H}(\vec{x},r)\ dr} \bigg] \ d\vec{x} + \nonumber \\
& \frac{\lambda Z_{B_H}}{|J(T)|} \bigg\}.
\label{eq:Evidence}
\end{align}

In finding the hypothesis $H\!=\!(K, T)$ that maximizes this evidence we are solving the classification problem (because $K$ is estimated) and the pose estimation problem (because $T$ is estimated). At the same time approximations to the segmentation $q$ and the reconstruction $v$ are obtained for the best hypothesis $H$. By construction the approximation for $v$ is a compromise between ``agreeing" with the shape prior of the estimated class $K$ when this prior is transformed by the estimated pose $T$, and ``explaining" the features observed in the input image $f$. 

As mentioned before, however, computing the evidence $L(H)$ for each hypothesis $H$ using \eqref{eq:Evidence} would be prohibitively expensive, because of the large number of pixels and voxels that need to be inspected to compute the integrals in that expression. For this reason we instead compute bounds for it and use an H\&B algorithm to select the best hypothesis. In the next two sections we describe how to compute those bounds.

%% file: LowerBound.tex
\section{Lower bound for $L(H)$}
\label{sec:LowerBound}

In this section we show how to efficiently compute lower bounds for the evidence $L(H)$ defined in \eqref{eq:Evidence}. Towards this end we first briefly review in \S \ref{sec:LowerBoundReview} the concept of a \emph{mean}-summary from \cite{Rot10} and two result concerning it. Then, in \S \ref{sec:LowerBoundDerivation}, we use these results to derive the lower bound for $L(H)$.

\subsection{Review: \emph{mean}-summaries}
\label{sec:LowerBoundReview}

\begin{definition}[\emph{Mean}-summary]
Given a BF $B$ defined on a set $\Omega$ and a partition $\Pi(\Omega)$ of this set, the \emph{mean-summary} is the functional $\hat{Y}_B = {\{ \hat{Y}_{B,\omega} \}}_{\omega \in \Pi(\Omega)}$ that assigns to each partition element $\omega \in \Pi(\Omega)$ the value $\hat{Y}_{B,\omega}$, defined by
\begin{align}
\hat{Y}_{B,\omega} \triangleq \int_{\omega}{\delta_B(\vec{x}) \ d\vec{x}}.
\label{eq:DefinitionMeanSummary}
\end{align}
\end{definition}
The name ``summary" is motivated by the fact that the ``infinite dimensional" BF is ``summarized" by just $n \triangleq |\Pi(\Omega)|$ values.

\emph{Mean}-summaries have two important properties: 1) for certain kinds of sets $\omega \in \Pi(\Omega)$, the values $\hat{Y}_{B,\omega}$ in the summary can be computed in constant time, regardless of the ``size" of the sets $\omega$ (using integral images \cite{Vio01}, see \S \ref{sec:ComputingSummaries} in the supplementary material); and 2) they can be used to obtain a lower bound for the evidence. 

It can be shown that the BFs that produce a given summary $\hat{Y}$ form an equivalence class. With an abuse of notation, we will use $B \sim \hat{Y}$ to denote the fact that a BF $B$ is in the equivalence class of the summary $\hat{Y}$. Next we prove two results that will be used to obtain a lower bound for the evidence.

\begin{lemma}[\emph{Mean}-summary identity]
\label{lem:MeanSummary}
Let $\Pi(\Omega)$ be a partition of a set $\Omega$, let $\hat{s}$ be a discrete shape defined on $\Pi(\Omega)$, let $B$ be a BF on $\Omega$, and let $\hat{Y}_B = \{ \hat{Y}_{B,\omega} \}$ ($\omega \in \Pi(\Omega)$) be the \emph{mean}-summary of $B$ in $\Pi(\Omega)$.
Then, for any continuous shape $s \sim \hat{s}$, it holds that
\begin{align}
\int_\Omega{\delta_B(\vec{x}) s(\vec{x}) \ d\vec{x}} = \sum_{\omega \in \Pi(\Omega)}{\hat{s}(\omega) \hat{Y}_{B,\omega}}.
\end{align}
\begin{proof}
\hspace{-4pt}: Immediate from the definitions. \qed
\end{proof}
\end{lemma}

\begin{lemma}[Relationship between the sets of continuous and discrete shapes]
\label{lem:SetsCDShapes}
Let $\Pi(\Omega)$ be a partition of a set $\Omega$, let $\mathbb{S}(\Omega)$ be the set of all continuous shapes in $\Omega$ and let $\hat{\mathbb{S}}(\Pi(\Omega))$ be the set of all discrete shapes in $\Pi(\Omega)$.
Then the set of continuous shapes that are produced by \emph{any} discrete shape in $\hat{\mathbb{S}}(\Pi(\Omega))$, $\mathbb{S}(\Pi(\Omega)) \! \triangleq\! \left\{s \! : s \sim \hat{s}, \hat{s} \in \hat{\mathbb{S}}(\Pi(\Omega)) \right\}$, is a subset of the set of all continuous shapes in $\Omega$, \ie,
$\mathbb{S}(\Pi(\Omega)) \subset \mathbb{S}(\Omega)$.
\begin{proof}
\hspace{-4pt}: Immediate from the definitions. \qed
\end{proof}
\end{lemma}

\subsection{Derivation of the lower bound}
\label{sec:LowerBoundDerivation}
In \cite{Rot10} we showed how to compute lower bounds for expressions that were much simpler than the evidence in \eqref{eq:Evidence}, by relying on partitions. To compute bounds for \eqref{eq:Evidence} we will also rely on a partition, namely, the standard partition. Thus, we define next the standard partition of $\left(\Theta, \Phi \right)$ and then proceed to derive the formulas for the bounds.

\begin{definition}[Standard partition]
\label{def:StandardPartition}
Let $\Theta \subset S^2$, $\Phi \subset \mathbb{R}^3$, and $\mathcal{R} \! \triangleq \! \left[R_{min},R_{max}\right]$ be three sets such that $\Phi \! = \! \Theta \times \mathcal{R}$ (as in Fig. \ref{fig:Setup}). Let $\Pi(\Theta) = \left\{\Theta_1, \dots, \Theta_{N_{\Theta}} \right\}$ be a partition of $\Theta$ and let $\Pi(\mathcal{R}) = \big\{[r_0,r_1), \dots, [r_{N_r-1},r_{N_r}) \big\}$ be a partition of $\mathcal{R}$ such that
\vspace{-5pt}
\begin{align}
\label{eq:PartitionR}
R_{min} &= r_0 < r_1 < \dots < r_{N_r} = R_{max} \quad \ \text{and} \\ 
r_i & = \beta r_{i-1} = \beta^i r_0, \ \ \text{for some} \ \beta > 1.
\label{eq:r_i}
\end{align}
The \emph{standard partition} for $\left(\Theta, \Phi \right)$ is defined to be $\big(\Pi(\Theta),$ $\Pi(\Phi )\big)$, where $\Pi(\Phi) \! \triangleq \! \Pi(\Theta) \times \Pi(\mathcal{R}) \! = \! \{\Phi_{1,1}, \Phi_{1,2}, \dots,$ $\Phi_{N_{\Theta} ,N_r}\}$ with $\Phi_{j,i} \triangleq \Theta_j \times [r_{i-1},r_i)$ (Fig. \ref{fig:StandardPartition}). Notice that given an arbitrary partition for the set $\Theta$ and a particular partition for the set $\mathcal{R}$, the standard partition defines a partition for the set $\Phi = \Theta \times \mathcal{R}$. 

\begin{figure}
\vspace{-5pt}
\begin{center}
\includegraphics[width=0.85\columnwidth]{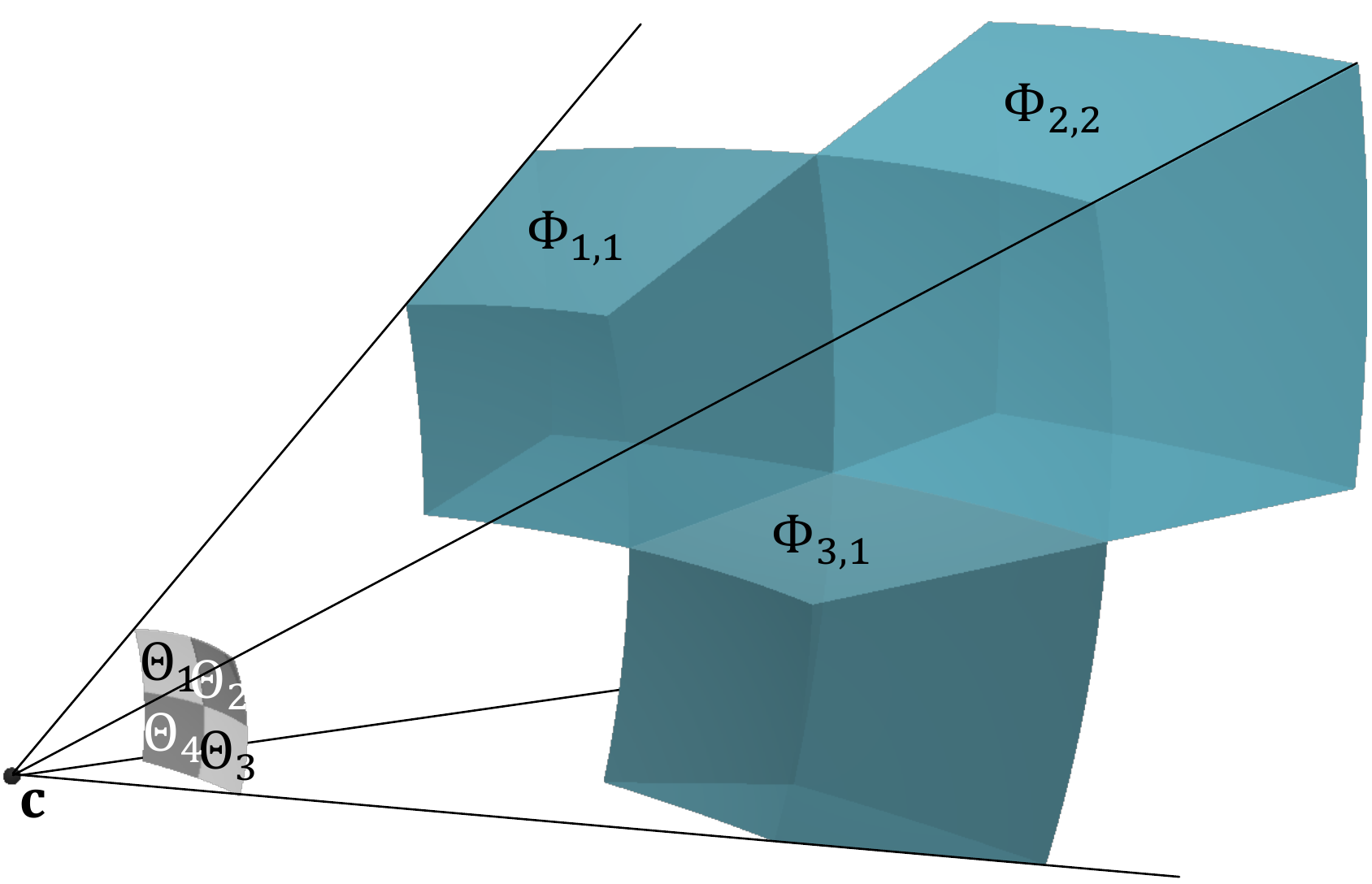}
\end{center}
\vspace{-15pt}
\caption{The standard partition for $(\Theta, \Phi)$, where $\Phi = \Theta \times \mathcal{R}$, $\mathcal{R} \triangleq [R_{min}, R_{max}]$, $\Pi(\Theta) = \{ \Theta_1, \Theta_2, \Theta_3, \Theta_4 \}$, and $\Pi(\mathcal{R}) = \{ [R_{min}, r_1), [r_1, R_{max}) \}$. For clarity only the voxels $\Phi_{1,1}, \Phi_{2,2}, \Phi_{3,1} \in \Pi(\Phi)$ are shown ($\Phi_{1,1} \triangleq \Theta_1 \times [R_{min}, r_1)$, $\Phi_{2,2} \triangleq \Theta_2 \times [r_1, R_{max})$, and $\Phi_{3,1} \triangleq \Theta_3 \times [R_{min}, r_1)$).}
\label{fig:StandardPartition}
\vspace{-15pt}
\end{figure}

In the next theorem we derive an expression to \linebreak[4] bound $L(H)$ from below. The main observation in this theorem is that, according to Lemma \ref{lem:SetsCDShapes}, the supremum in \eqref{eq:Evidence} for $q \in \mathbb{S}(\Pi(\Theta))$ and $v \in \mathbb{S}(\Pi(\Phi))$ is less than the supremum for $q \in \mathbb{S}(\Theta)$ and $v \in \mathbb{S}(\Phi)$. Moreover, since the continuous shapes in $\mathbb{S}(\Pi(\Theta))$ and $\mathbb{S}(\Pi(\Phi))$ are constant inside each partition element, evaluating this new supremum is easier.

\begin{theorem}[Lower bound for \emph{L(H)}]
\label{theo:LowerBound}
Let $\mathbf{\Pi} \triangleq \ $ $(\Pi(\Theta), \Pi(\Phi))$ be a standard partition and let $\hat{Y}_f = {\left\{\hat{Y}_{f,\theta}\right\}}_{\theta \in \Pi(\Theta)}$ and $\hat{Y}_H = {\left\{\hat{Y}_{H,\phi}\right\}}_{\phi \in \Pi(\Phi)}$ be the \emph{mean}-summaries of two unknown BFs in $\Pi(\Theta)$ and $\Pi(\Phi)$, respectively. Let $\psi_{j,k}$ be the set of the indices of the $k$ largest elements of $\left\{\hat{Y}_{H,\Phi_{j,1}}, \hat{Y}_{H,\Phi_{j,2}}, \dots, \hat{Y}_{H,\Phi_{j,N_r}}\right\}$, and let $\Psi_{j,k}$ be the sum of these elements, \ie,
\begin{equation}
\Psi_{j,k} \triangleq \sum_{i \in \psi_{j,k}}{ \hat{Y}_{H,\Phi_{j,i}} }.
\label{eq:SumLargestK}
\end{equation}

Then, for any BF $B_f \sim \hat{Y}_f$ and any BF $B_H \sim \hat{Y}_H$, it holds that $L(H) \ge \underline{L}_{\mathbf{\Pi}}(H)$, where
\begin{align}
\label{eq:LowerBound}
\underline{L}_{\mathbf{\Pi}}(H) & \triangleq \frac{\lambda Z_{B_H}}{|J(T)|} + \sum_{j=1}^{N_{\Theta}} \underline{\mathcal{L}}_{\Theta_j}(H),  \ \ \text{and} \\
\underline{\mathcal{L}}_{\Theta_j}(H) & \triangleq \mathop{\max}_{0 \le \mathsf{n}_j \le N_r} \bigg\{\mathop{\max}_{\mathsf{q} \in \{0,1\}} \bigg[\mathsf{q} \hat{Y}_{f, \Theta_j} + \nonumber \\
& \left|\Theta_j\right| \log{P\left(\mathsf{q}| \mathsf{n}_j \log{\beta} \right)} \bigg] + \frac{\lambda}{|J(T)|} \Psi_{j,\mathsf{n}_j} \bigg\}.
\label{eq:LowerBoundLocal}
\end{align}

Moreover, the 3D reconstruction and the 2D segmentation corresponding to this bound are given by the discrete shapes $\hat{v}$ and $\hat{q}$, respectively, defined by 
\begin{align}
\label{eq:ReconstructionLower}
\hat{v}(\Phi_{j,i}) & \triangleq \left \{
\begin{array}{ll}
1, & \ \ \text{if} \ i \in \psi_{j, \mathsf{n}_j^*}, \\
0, & \ \ \text{otherwise},
\end{array} \right. \qquad \text{and} \\
\label{eq:SegmentationLower}
\hat{q}(\Theta_j) & \! \triangleq \!
\mathop{\arg \max}_{\mathsf{q} \in \{0,1\}} \bigg[\mathsf{q} \hat{Y}_{f, \Theta_j} \! + \!
\left|\Theta_j\right| \log{P\left(\mathsf{q}| \mathsf{n}_j^* \log{\beta} \right)} \bigg] \!,
\end{align}
where $\mathsf{n}_j^*$ is the solution to \eqref{eq:LowerBoundLocal}.

\begin{proof}
\hspace{-4pt}: From Lemma \ref{lem:SetsCDShapes} it holds that $L(H)$ (defined in \eqref{eq:Evidence}) is greater than or equal to
\begin{align}
\frac{\lambda Z_{B_H}}{|J(T)|} & + \mathop{\max}_{\hat{q},\hat{v}} \bigg\{ \mathop{\sup}_{
\begin{array}{c}
\scriptstyle{q \sim \hat{q}} \\ 
\scriptstyle{v \sim \hat{v}} \end{array}} 
\int_{\Theta}{\bigg[\delta_{B_f}(\vec{x}) q(\vec{x}) + } \nonumber \\
& + \frac{\lambda}{|J(T)|} \int_{R_{min}}^{R_{max}}{r^2 v(\vec{x},r) \delta_{B_H}(\vec{x},r) \ dr} + \nonumber \\
& + \log{P\left(q(\vec{x}) | \ell_v(\vec{x})\right)} \bigg] \ d\vec{x} \bigg\}.
\label{eq:LB1}
\end{align}

Since $q \! \sim \! \hat{q}$ and $v \! \sim \! \hat{v}$, it follows from Lemma \ref{lem:MeanSummary} that
\begin{align}
\int_{\Theta}{\delta_{B_f}(\vec{x})q(\vec{x})\ d\vec{x}} = \sum_{j=1}^{N_{\Theta}}{\hat{q}\left(\Theta_j\right)\hat{Y}_{f,\Theta_j}}, \ \text{and}
\label{eq:LBMeanSummaryQ}
\end{align}
\begin{align}
\int_{\Theta}{\int_{R_{min}}^{R_{max}}{r^2 v(\vec{x},r) \delta_{B_H}(\vec{x},r)\ dr}\ d\vec{x}} = \nonumber \\
\sum_{j=1}^{N_{\Theta}}{\sum_{i=1}^{N_r}{\hat{v}\left(\Phi_{j,i}\right) \hat{Y}_{H,\Phi_{j,i}} }}.
\label{eq:LBMeanSummaryV}
\end{align}
On the other hand, $\ell_v(\vec{x})$ is constant inside each element of $\Pi(\Theta)$, because $\forall \vec{x} \in \Theta_j$,
\begin{align}
\ell_v(\vec{x}) & = 
\int_{r_0}^{r_{N_r}}{\frac{v(\vec{x},r)}{r}\ dr} =
\sum_{i=1}^{N_r}{\hat{v}\left(\Phi_{j,i}\right) \log{\left( \frac{r_i}{r_{i-1}} \right) }} = \nonumber \\
& = \log{(\beta)} \sum_{i=1}^{N_r}{\hat{v}\left(\Phi_{j,i}\right)}.
\label{eq:LBLV}
\end{align}

Then, substituting \eqref{eq:LBMeanSummaryQ}, \eqref{eq:LBMeanSummaryV} and \eqref{eq:LBLV} into \eqref{eq:LB1}, that expression reduces to
\begin{align}
\frac{\lambda Z_{B_H}}{|J(T)|} & + 
\mathop{\max}_{\hat{q},\hat{v}} \sum_{j=1}^{N_{\Theta}}{\bigg[\hat{q}\left(\Theta_j\right) \hat{Y}_{f,\Theta_j} + }\nonumber \\
& + \left|\Theta_j\right| \log{ P\left(\hat{q}\left(\Theta_j\right) \Big| \log{(\beta)} \sum_{i=1}^{N_r}{\hat{v}\left(\Phi_{j,i}\right)}\right)} + \nonumber \\
& + \frac{\lambda}{|J(T)|} \sum_{i=1}^{N_r}{\hat{v}\left(\Phi_{j,i}\right) \hat{Y}_{H,\Phi_{j,i}}} \bigg].
\label{eq:LB2}
\end{align}

Note that the first (leftmost) term inside the square brackets in \eqref{eq:LB2} does not depend on the states of the voxels $\hat{v}\left(\Phi_{j,i}\right)$, the third term does not depend on the state of the pixel $\hat{q}\left(\Theta_j\right)$, and the second term does not depend on which particular voxels are full, but on the number of full voxels $\mathsf{n}_j \triangleq \sum_{i=1}^{N_r}{\hat{v}\left(\Phi_{j,i}\right)}$. In contrast, the third term does depend on which voxels are full, and given the number $\mathsf{n}_j$ of full voxels, the third term will be maximum when the $\mathsf{n}_j$ voxels with the largest summary $\hat{Y}_{H,\Phi_{j,i}}$ are full. In this case the third term takes the value $\Psi_{j,\mathsf{n}_j}$, defined in \eqref{eq:SumLargestK}. Therefore \eqref{eq:LB2} is equal to,
\begin{align}
\frac{\lambda Z_{B_H}} {|J(T)|} & + 
\sum_{j=1}^{N_{\Theta}} \mathop{\max}_{0 \le \mathsf{n}_j \le N_r} \bigg\{
\mathop{\max}_{\mathsf{q} \in \{0,1\}} \bigg[\mathsf{q} \hat{Y}_{f,\Theta_j} + \\
& + \left|\Theta_j\right| \log{P\left(\mathsf{q} | \mathsf{n}_j \log{\beta} \right)} \bigg] + 
\frac{\lambda}{|J(T)|} \Psi_{j,\mathsf{n}_j} \bigg\}, \nonumber
\end{align}
which can be rearranged to yield \eqref{eq:LowerBound} and \eqref{eq:LowerBoundLocal}. That the discrete shapes $\hat{v}$ and $\hat{q}$ (defined respectively in \eqref{eq:ReconstructionLower} and \eqref{eq:SegmentationLower}) maximize \eqref{eq:LowerBoundLocal} follows immediately, proving the theorem. \qed
\end{proof}
\end{theorem}

Note that this bound is computed in $O(N_{\Theta} N_r \log{N_r})$.

%% file: UpperBound.tex
\section{Upper bound for $L(H)$}
\label{sec:UpperBound}

In this section we show how to efficiently compute upper bounds for the evidence $L(H)$ defined in \eqref{eq:Evidence}. Towards this end we first briefly review in \S \ref{sec:UpperBoundReview} some concepts and results from \cite{Rot10}, and then in \S \ref{sec:UpperBoundDerivation} we use these concepts and results to derive the upper bound for $L(H)$.

\subsection{Review: semidiscrete shapes and $m$-summaries}
\label{sec:UpperBoundReview}

In \S \ref{sec:LowerBound} we used discrete shapes and \emph{mean}-summa\-ries to get a lower bound for the evidence. Analogously, in this section we define and use the concepts of semidiscrete shapes and $m$-summaries to get an upper bound. Like discrete shapes and \emph{mean}-summaries, semidiscrete shapes and $m$-summaries ``condense" the ``infinite dimensional" continuous shapes and BFs, respectively, into a finite set of real numbers.

\begin{definition}[Semidiscrete shape]
Given a partition $\Pi(\Omega)$ of a set $\Omega \subset \mathbb{R}^d$, the \emph{semidiscrete shape} $\tilde{s}$ is the function $\tilde{s}: \Pi(\Omega) \to \mathbb{R}$ that associates to each element $\omega \in \Pi(\Omega)$ a value in the interval $[0, |\omega|]$.
Given a continuous shape $s$ and a semidiscrete shape $\tilde{s}$ we say that both shapes are \emph{equivalent} (denoted as $s \sim \tilde{s}$) if $\tilde{s}(\omega) = |s \cap \omega| \ \forall \omega \in \Pi(\Omega)$ (\ie, if the measure of $s$ in each set $\omega$ is equal to $\tilde{s}(\omega)$). Informally the semiscrete shape $ \tilde{s}$ ``remembers" \emph{how much} of each partition element is occupied by the shape $s$, but ``forgets" the state $s(\vec{x})$ of each particular point $\vec{x} \in \omega$.
\end{definition}

The following lemma explores the relationship between the sets of continuous and semidiscrete shapes. It simply states that, for any partition, every continuous shape is equivalent to some semidiscrete shape in the partition.

\begin{lemma}[Relationship between the sets of continuous and semidiscrete shapes]
\label{lem:SetsCSShapes}
Let $\Pi(\Omega)$ be a partition of a set $\Omega$, let $\mathbb{S}(\Omega)$ be the set of all continuous shapes in $\Omega$ and let $\tilde{\mathbb{S}}(\Pi(\Omega))$ be the set of all semidiscrete shapes in $\Pi(\Omega)$. Then
\begin{align}
\left\{ s \in \mathbb{S}(\Omega) : s \sim \tilde{s}, \tilde{s} \in \tilde{\mathbb{S}}(\Pi(\Omega)) \right\} = \mathbb{S}(\Omega).
\end{align}
\begin{proof}
\hspace{-4pt}: Immediate from the definitions. \qed
\end{proof}
\end{lemma}

\begin{definition}[\emph{m}-summary]
Given a BF $B$ defined on a set $\Omega$ and a partition $\Pi(\Omega)$ of this set, the \emph{m-summary} is the functional $\tilde{Y}_B = {\{ \tilde{Y}_{B,\omega} \}}_{\omega \in \Pi(\Omega)}$ that assigns to each partition element $\omega \in \Pi(\Omega)$ the $(2m+1)$-dimensional vector $\tilde{Y}_{B,\omega} = [\tilde{Y}^{-m}_{B,\omega}, \dots, \tilde{Y}^{m}_{B,\omega}]$, whose components are defined by
\begin{align}
\tilde{Y}^j_{B,\omega} \triangleq \Big|\Big\{ \vec{x} \in \omega : \delta_B(\vec{x}) < \frac{j \delta_{max}} {m}\Big\}\Big|,
\label{eq:DefMSummary}
\end{align}
for $j = -m, \dots, m$. In other words, the $m$-summary element $\tilde{Y}_{B,\omega}$ ``remembers" how the values of $\delta_B$ in the set $\omega$ are distributed, but ``forgets" \emph{where} those values are within the set. More specifically, the quantity $(\tilde{Y}^{j + 1}_{B,\omega} - \tilde{Y}^j_{B,\omega})$ indicates the measure of the subset of $\omega$ whose values of $\delta_B$ are in the interval $[j \delta_{max} / m, (j + 1) \delta_{max} / m)$.
Given a BF $B$ and an $m$-summary $\tilde{Y}$ defined on a partition $\Pi(\Omega)$, we say that they are \emph{equivalent} (denoted as $B \sim \tilde{Y}$) if they satisfy \eqref{eq:DefMSummary} for each set $\omega \in \Pi(\Omega)$. Throughout this work we use $m=6$.
\end{definition}

\emph{M}-summaries, like \emph{mean}-summaries, have two important properties: 1) for certain kinds of sets $\omega \in \Pi(\Omega)$, the values $\tilde{Y}_{B,\omega}$ in the summary can be computed in constant time, regardless of the ``size" of the sets $\omega$ (using integral images \cite{Vio01}, see \S \ref{sec:ComputingSummaries} in the supplementary material); and 2) they can be used to obtain an upper bound for the evidence. Lemma \ref{lem:mSummary}, below, will be used to obtain this upper bound. This lemma tells us how to bound the integral $\int_{\Omega} {\delta_B(\vec{x}) s(\vec{x})}$ $d\vec{x}$ when the BF $B$ is only known to be equivalent to an $m$-summary $\tilde{Y}$, and the continuous shape $s$ is only known to be equivalent to a semidiscrete shape $\tilde{s}$.

\begin{lemma}[\emph{m}-summary bound]
\label{lem:mSummary}
Let $\Pi(\Omega)$ be a partition of a set $\Omega$, let $\tilde{s}$ be a semidiscrete shape defined on $\Pi(\Omega)$, and let $\tilde{Y} = {\{ \tilde{Y}_{\omega} \}}_{\omega \in \Pi(\Omega)}$ be an $m$-summary in $\Pi(\Omega)$.
Then for any continuous shape $s$ on $\Omega$ such that $s \sim \tilde{s}$, it holds that
\begin{align}
\mathop{\sup}_{B \sim \tilde{Y}} \left[\int_{\Omega} {\delta_B(\vec{x}) s(\vec{x}) \ d\vec{x}}\right] \le
\sum_{\omega \in \Pi(\Omega)} {F_{\tilde{Y},\omega}(\tilde{s}(\omega))},
\label{eq:LemMSummaryThesis}
\end{align}
where $B$ is a BF,
\begin{align}
\label{eq:F}
F_{\tilde{Y},\omega}(S) & \triangleq \frac{\delta_{max}}{m} \Bigg[\sum_{j = J(S)}^{m - 1}{\left(\tilde{Y}^{j + 1}_{\omega} - \tilde{Y}^j_{\omega}\right) (j + 1)} + \nonumber \\
& + J(S) \left(\tilde{Y}^J_{\omega} - |\omega| + S \right) \Bigg], \quad \text{and} \\
\label{eq:J}
J(S) & \triangleq \min \left\{ j: \tilde{Y}^j_{\omega} \ge |\omega| - S \right\}.
\end{align}
\begin{proof}
\hspace{-4pt}: See \cite[Lemma 1]{Rot10}. \qed
\end{proof}
\end{lemma}

Intuitively, the continuous shape $s$ that yields the supremum of the integral in \eqref{eq:LemMSummaryThesis} contains the parts of each $\omega$ where $\delta_B$ is greatest and has a ``mass" of $\tilde{s}(\omega)$ in each $\omega \in \Pi(\Omega)$. Hence, the supremum within each set $\omega$, $F_{\tilde{Y}, \omega}(\tilde{s}(\omega))$, is obtained by adding the ``mass" of the subset of $\omega$ where the value of $\delta_B$ is in the interval $[j \delta_{max} / m, (j + 1) \delta_{max} / m)$, times the maximum value of $\delta_B$ in this interval, $(j + 1) \delta_{max} / m$. These terms are added in descending order of $j$ until the total mass allocated in $\omega$ is $\tilde{s}(\omega)$.

\subsection{Upper bound for $L(H)$}
\label{sec:UpperBoundDerivation}

In this subsection we will derive a formula to compute an upper bound for the evidence $L(H)$ defined in \eqref{eq:Evidence}. Towards this end recall that $L(H)$ is computed by solving an optimization problem of the form 
\begin{align}
L(H) = \mathop{\sup}_{q, v} E_1(B_f, B_H, q, v),
\label{eq:OptimizationProblemOriginal}
\end{align}
where $B_f$ and $B_H$ are two BFs obtained from the input image $f$ and the hypothesis $H$, respectively, and $q$ and $v$ are two continuous shapes. 
To derive the bound, we proceed in three steps.

\paragraph{Step 1.}
We reduce the ``amount  of information" to be processed in the computation of $L(H)$ by considering not only the given BFs $B_f$ and $B_H$, but also all the BFs ${B'}_{\! f}$ and ${B'}_{\! H}$ that have the \emph{same} $m$-summaries $\tilde{Y}_f$ and $\tilde{Y}_H$, respectively. In other words, $B_f \sim \tilde{Y}_f \sim {B'}_{\! f}$ and $B_H \sim \tilde{Y}_H \sim {B'}_{\! H}$. In doing so we obtain an upper bound for $L(H)$,
\begin{align}
\label{eq:OptimizationProblemSummaries}
L(H) & \le \mathop{\sup}_{q, v} E_2(\tilde{Y}_f, \tilde{Y}_H, q, v), \quad \text{where} \\
E_2(\tilde{Y}_f, \tilde{Y}_H, q, v) & \triangleq \mathop{\sup}_{
\begin{array}{c}
\scriptstyle{{B'}_{\! f} \sim \tilde{Y}_f} \\ 
\scriptstyle{{B'}_{\! H} \sim \tilde{Y}_H} 
\end{array}} 
E_1({B'}_{\! f}, {B'}_{\! H}, q, v).
\label{eq:G2}
\end{align}
Therefore, we can disregard the details about each BF and only consider the information in their $m$-summaries. Moreover, using Lemma \ref{lem:mSummary}, we can bound the first and third terms in \eqref{eq:Evidence} as a function of the semidiscrete shapes $\tilde{q}$ and $\tilde{v}$ defined on the standard partition $\big(\Pi(\Theta),$ $\Pi(\Phi)\big)$.
 
\paragraph{Step 2.}
The second term in \eqref{eq:Evidence} can also be bounded in terms of the shapes $\tilde{q}$ and $\tilde{v}$.
In Lemma \ref{lem:UBProjectionTerm} we will show that it is possible to write in closed form (up to a permutation of the rays within each partition element) the continuous shapes $q_*$ and $v_*$ that maximize this term and are respectively equivalent to $\tilde{q}$ and $\tilde{v}$. We denote this fact as $q_* = f_1(\tilde{q})$ and $v_* = f_2(\tilde{q}, \tilde{v})$. Therefore, for each pair of semidiscrete shapes $\tilde{q}$ and $\tilde{v}$, we do not need to consider all possible continuous shapes $q$ and $v$ such that $q \sim \tilde{q}$ and $v \sim \tilde{v}$, we only need to consider $f_1(\tilde{q})$ and $f_2(\tilde{q}, \tilde{v})$. Hence, the optimization problem in \eqref{eq:OptimizationProblemSummaries} can be simplified to a problem of the form
\begin{align}
\mathop{\max}_{\tilde{q}, \tilde{v}} E_2(\tilde{Y}_f, \tilde{Y}_H, f_1(\tilde{q}),f_2(\tilde{q}, \tilde{v})),
\label{eq:OptimizationProblemSimplified}
\end{align}
\ie, to estimate a pair of semidiscrete shapes rather than a pair of discrete shapes.
This means that in \eqref{eq:OptimizationProblemSimplified} only a finite number of quantities needs to be estimated ($\tilde{q}(\Theta_j)$ and $\tilde{v}(\Phi_{j,i})$, for $j = 1, \dots, N_\Theta$ and $i = 1, \dots, N_r$), whereas in \eqref{eq:OptimizationProblemSummaries} an ``infinite number of quantities" needs to be estimated ($q(\vec{x})$ and $v(\vec{X})$, for $\vec{x} \in \Theta$ and $\vec{X} \in \Phi$).

\paragraph{Step 3.}
As it turns out, the problem in \eqref{eq:OptimizationProblemSimplified} can be further simplified because it is possible to efficiently compute the optimal semidiscrete shape $\tilde{v}_*$ that corresponds to any semidiscrete shape $\tilde{q}$. We denote this fact as $\tilde{v}_* = f_3(\tilde{q})$. Therefore the problem in \eqref{eq:OptimizationProblemSimplified} can be simplified into a problem of the form
\begin{align}
\mathop{\max}_{\tilde{q}} E_2(\tilde{Y}_f, \tilde{Y}_H, f_1(\tilde{q}),f_2(\tilde{q}, f_3(\tilde{q}))).
\label{eq:SimplifiedOptimizationProblem2}
\end{align}
That is, for each element $\Theta_j$ of the partition $\Pi(\Theta)$, we need to solve, independently, an optimization over the scalar parameter $\mathsf{q}_j \triangleq \tilde{q}(\Theta_j)$. Each of these optimizations is solved using grid search.

\begin{figure}
\vspace{-5pt}
\begin{center}
\includegraphics[width=0.95\columnwidth]{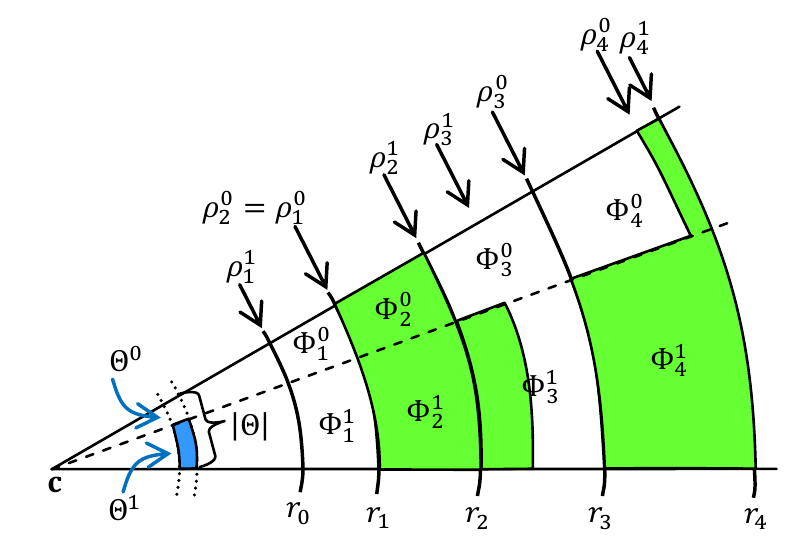}
\end{center}
\vspace{-15pt}
\caption{Continuous shapes  $q_*$ (in blue) and $v_*$ (in green) that maximize \eqref{eq:DefBigLambda} when the mass in a pixel $\Theta$ is constrained to be $\tilde{q}(\Theta)$ and the mass in each voxel $\Phi_i$ is constrained to be $\tilde{v}(\Phi_i)$ (for convenience the pixel subscript $j$ has been omitted). $\Theta^0$ and $\Theta^1$ are, respectively, the parts of the pixel $\Theta$ where $q_*(\vec{x})\!=\!0$ and $q_*(\vec{x}) \!=\! 1$. The side of each voxel $\Phi_i$ that projects to $\Theta^1$, $\Phi_i^1$, is filled first. In this part the mass is concentrated on the inner side of the voxel, between $r_{i-1}$ and $\rho^1_i$. Only when this part is full, the other part of the voxel (\ie, the part that projects to $\Theta^0$), $\Phi_i^0$, starts to be filled. On that part of the voxel the mass is concentrated in the outer part, between $\rho^0_i$ and $r_i$. $\rho^1_i$ and $\rho^0_i$ are computed using the formulas in \eqref{eq:Rho1} and \eqref{eq:Rho0}.}
\vspace{-5pt}
\label{fig:BoundProjection}
\end{figure}

The functions $f_1$ and $f_2$ mentioned in Step 2 above are informally defined in Fig. \ref{fig:BoundProjection}. Given the semidiscrete shape $\tilde{q}$, the function $f_1$ returns a continuous shape $q_*$ in $\Theta$ such that $q_* \sim \tilde{q}$. From Lemma \ref{lem:mSummary}, any continuous shape in the set $\{q: q \sim \tilde{q} \}$ is ``equally good," providing the same bound for the first term in \eqref{eq:Evidence}. This is what we meant by ``up to a permutation of the rays within each partition element."

The function $f_2$ is somewhat more complex. To understand it, we need to define the sets $\Theta^0_j$, $\Theta^1_j$, $\Phi^0_{j,i}$ and $\Phi^1_{j,i}$. Given the continuous shape $q_*$ in $\Theta$, the sets $\Theta^0_j$ and $\Theta^1_j$ are the parts of pixel $\Theta_j$ where $q_*(\vec{x})=0$ and $q_*(\vec{x})=1$, respectively, and $\Phi^0_{j,i}$ and $\Phi^1_{j,i}$ are the parts of voxel $\Phi_{j,i}$ that project to $\Theta^0_j$ and $\Theta^1_j$, respectively (Fig. \ref{fig:BoundProjection}). More formally,
\begin{align}
\label{eq:Theta0}
\Theta^0_j & \triangleq \left\{\vec{x} \in \Theta_j: q_*(\vec{x})=0\right\}, \\
\label{eq:Theta1}
\Theta^1_j & \triangleq \left\{\vec{x} \in \Theta_j: q_*(\vec{x})=1\right\},
\end{align}
\begin{align}
\Phi^0_{j,i} & \triangleq \Theta^0_j \times \left[r_{i-1}, r_i\right], \ \ \text{and}\\
\label{eq:Phi1}
\Phi^1_{j,i} & \triangleq \Theta^1_j \times \left[r_{i-1}, r_i\right].
\end{align}
\end{definition}

The continuous shape $v_*$ returned by $f_2$ has a ``mass" of $\tilde{v}(\Phi_{j,i})$ inside each voxel $\Phi_{j,i}$ (because $v_* \sim \tilde{v}$). As demonstrated in Lemma \ref{lem:UBProjectionTerm} below, within each voxel $\Phi_{j,i}$, the mass is preferentially allocated on $\Phi^1_{j,i}$ rather than on $\Phi^0_{j,i}$. To make this more precise, let us define the quantities $\mathsf{v}^1_{j,i}$ and $\mathsf{v}^0_{j,i}$ to be the mass of $v_*$ in $\Phi^1_{j,i}$ and $\Phi^0_{j,i}$, respectively, that is
\begin{align}
\label{eq:v1j}
\mathsf{v}^1_{j,i} & \triangleq \left| \left\{ \vec{X} \in \Phi^1_{j,i}: v_*(\vec{X}) = 1 \right\} \right|, \ \ \text{and}\\
\label{eq:v0j}
\mathsf{v}^0_{j,i} & \triangleq \left| \left\{ \vec{X} \in \Phi^0_{j,i}: v_*(\vec{X}) = 1 \right\} \right|.
\end{align}
Since the mass is allocated preferentially in $\Phi^1_{j,i}$ rather than in $\Phi^0_{j,i}$, these quantities can be computed with the formulas
\begin{align}
\label{eq:Computev1}
\mathsf{v}^1_{j,i} & \triangleq \min \{ \tilde{v}(\Phi_{j,i}), |\Phi^1_{j,i}| \} \qquad \text{and} \\
\label{eq:Computev0}
\mathsf{v}^0_{j,i} & \triangleq \max \{ \tilde{v}(\Phi_{j,i}) - |\Phi^1_{j,i}|, 0 \} = \tilde{v}(\Phi_{j,i}) - \mathsf{v}^1_{j,i}.
\end{align}

It is also proved in Lemma \ref{lem:UBProjectionTerm} that the mass in $\Phi^1_{j,i}$ lies on the inner side of the voxel, while the mass in $\Phi^0_{j,i}$ lies on the outer side of the voxel (Fig. \ref{fig:BoundProjection}).
To make this statement more precise, recall that $\mathsf{q}_j \triangleq \tilde{q}(\Theta_j)$ and define the quantities $\rho^1_{j,i} = \rho^1_{j,i}(\mathsf{q}_j, \mathsf{v}^1_{j,i})$ and $\rho^0_{j,i} = \rho^0_{j,i}(\mathsf{q}_j, \mathsf{v}^0_{j,i})$ to be the radius of the outer part of $\Phi^1_{j,i}$ that is full, and the the radius of inner part of $\Phi^0_{j,i}$ that is full, respectively. In order to compute these quantities, we define the \emph{voxel volume} function $\Upsilon$.
Given a voxel $\phi \triangleq \theta \times [\rho_0, \rho_1)$ defined by the Cartesian product between a set $\theta$ in the camera retina and an interval $[\rho_0, \rho_1)$ in the real line (as in Definition \ref{def:StandardPartition}), its volume is given by $\Upsilon(|\theta|, \rho_0, \rho_1)$, where $|\theta|$ is the solid angle on the camera retina subtended by $\theta$ (which is equal to its measure) and the function $\Upsilon$ is defined by
\begin{equation}
\Upsilon \left(a,\rho_0, \rho_1\right) \triangleq \frac{a} {3} \left({\rho_1}^3 - {\rho_0}^3\right).
\end{equation}
Using this definition the volume of a voxel $\Phi_{j,i}$ in the standard partition is $\Upsilon(\left|\Theta_j\right|,$ $r_{i-1},r_i)$, the volume of the set $\Phi^k_{j,i}$ ($k \in \{0,1\}$) is 
\begin{align}
|\Phi^k_{j,i}| = \Upsilon \big(\left|\Theta^k_j\right|, r_{i-1}, r_i\big),
\label{eq:VolumePhik}
\end{align}
and the mass of $v_*$ in $\Phi^1_{j,i}$ and $\Phi^0_{j,i}$ is, respectively,
\begin{align}
\label{eq:Rho1}
\mathsf{v}^1_{j,i} & = \Upsilon(\mathsf{q}_j,r_{i-1},\rho^1_{j,i}), \qquad \text{and} \\
\label{eq:Rho0}
\mathsf{v}^0_{j,i} & = \Upsilon(|\Theta_j| - \mathsf{q}_j,\rho^0_{j,i}, r_i).
\end{align}
Hence, given $\mathsf{q}_j$ and $\tilde{v}(\Phi_{j,i})$, it holds that $|\Theta_j^1| = \mathsf{q}_j$, $|\Phi^1_{j,i}|$ can be found using \eqref{eq:VolumePhik}, 
$\mathsf{v}^1_{j,i}$ and $\mathsf{v}^0_{j,i}$ can be found using \eqref{eq:Computev1} and \eqref{eq:Computev0}, and $\rho^1_{j,i}$ and $\rho^0_{j,i}$ can be found as the solutions to \eqref{eq:Rho1} and \eqref{eq:Rho0}.

The previous statements about $v_*$ are proved in the following lemma. For notational convenience we group the quantities $\{\mathsf{v}^0_{j,i} \}$ and $\{\mathsf{v}^1_{j,i} \}$ that project to the same pixel $\Theta_j$ into the vectors
\begin{align}
\mathsf{V}^0_j \triangleq \left[\mathsf{v}^0_{j,1}, \dots , \mathsf{v}^0_{j,N_r}\right] \ \text{and} \ 
\mathsf{V}^1_j \triangleq \left[\mathsf{v}^1_{j,1}, \dots , \mathsf{v}^1_{j,N_r}\right].
\label{eq:V01}
\end{align}

\begin{lemma}[Upper bound for the projection term]
\label{lem:UBProjectionTerm}
Let $\left(\Pi(\Theta), \Pi(\Phi)\right)$ be a standard partition and let $\tilde{q}$ and $\tilde{v}$ be two semidiscrete shapes in $\Pi(\Theta)$ and $\Pi(\Phi)$, respectively. Let $q$ and $v$ be two continuous shapes in $\Theta$ and $\Phi$, respectively, and define  
\begin{equation}
\Lambda_{\Theta}(q,v) \triangleq \int_{\Theta}{\log{P\left(q(\vec{x})|\ell_v(\vec{x})\right)}\ d\vec{x}},
\label{eq:DefBigLambda}
\end{equation}
with the integrand of this expression defined as in \eqref{eq:ProjectionTermForm}. Then:

1. Any continuous shape $q_*$ such that $q_* \sim \tilde{q}$ and the continuous shape $v_*$ defined by 
\begin{align}
v_*(\vec{X}) \!= \!\left\{\begin{array}{cc}
\!\!1 & \text{if } \vec{X} \! \in \bigcup_{j,i}{\Big[\Theta_j^1 \! \times \! [r_{i-1}, \rho_{j,i}^1] \cup \Theta_j^0 \!\times \! [\rho_{j,i}^0, r_i] \Big]} \\
\!\!0 & \text{otherwise},
\end{array} \right.
\label{eq:VStar}
\end{align}
(Fig. \ref{fig:BoundProjection}), where the quantities $\Theta_j^0$, $\Theta_j^1$, $r_i$, $\rho^1_{j,i}$ and $\rho^0_{j,i}$ are defined in \eqref{eq:Theta0}-\eqref{eq:Theta1}, \eqref{eq:r_i}, \eqref{eq:Rho1} and \eqref{eq:Rho0}, respectively, are a solution to the problem
\begin{align}
\label{eq:DefSupLambda}
\overline{\Lambda}_{\Theta} (\tilde{q},\tilde{v}) \triangleq \mathop{\sup}_{ \begin{array}{c}
\scriptstyle{q \sim \tilde{q}} \\ 
\scriptstyle{v \sim \tilde{v}} \end{array}}
\Lambda_{\Theta}(q,v).
\end{align}

2. The optimal value can be computed as
\begin{align}
\overline{\Lambda}_{\Theta} (\tilde{q},\tilde{v}) \triangleq & \sum_{j}{ \Big[ (|\Theta_j|-\mathsf{q}_j) \log{P\left(q=0|\ell^0(\mathsf{q}_j,\mathsf{V}^0_j)\right)}} \ + \nonumber \\
& \ \mathsf{q}_j \log{P\left(q=1|\ell^1(\mathsf{q}_j,\mathsf{V}^1_j)\right) \Big]},
\label{eq:SolutionBigLambda}
\end{align}
where
\begin{align}
\label{eq:DefEll0}
\ell^0(\mathsf{q}_j,\mathsf{V}^0_j) & \triangleq \sum_{i=1}^{N_r}{\log{\frac{r_i}{\rho^0_{j,i}(\mathsf{q}_j,\mathsf{v}^0_{j,i})}} },\ \text{and} \\
\label{eq:DefEll1}
\ell^1(\mathsf{q}_j,\mathsf{V}^1_j) & \triangleq \sum_{i=1}^{N_r}{\log{\frac{\rho^1_{j,i}(\mathsf{q}_j,\mathsf{v}^1_{j,i})}{r_{i-1}}} }. 
\end{align}
\begin{proof}
\hspace{-4pt}: See proof in the appendix.\qed
\end{proof}
\end{lemma}

Notice that the quantities ${\{\mathsf{q}_j \}}$ ($j=1, \dots, N_{\Theta}$) simply relabel the quantities in the semidiscrete shape $\tilde{q}$. Notice also that $\tilde{v}(\Phi_{j,i}) = \mathsf{v}^0_{j,i} + \mathsf{v}^1_{j,i}$ for every element $\Phi_{j,i} \in \Pi(\Phi)$. Therefore, the quantities ${\{\mathsf{v}^0_{j,i} + \mathsf{v}^1_{j,i} \}}$ ($j=1, \dots, N_{\Theta},\ i=1, \dots, N_r$) define the semidiscrete shape $\tilde{v}$. 
This notation emphasizes the fact that to estimate the semidiscrete shape $\tilde{q}$, we need to estimate $N_{\Theta}$ scalar quantities, and to estimate the semidiscrete shape $\tilde{v}$, we need to estimate $2 N_{\Theta} N_r$ scalar quantities. These quantities are estimated in the process of computing the upper bound in the next theorem.

\begin{theorem}[Upper bound for \emph{L(H)}]
\label{theo:UpperBound}
Let $\mathbf{\Pi} \triangleq \ $ $(\Pi(\Theta), \Pi(\Phi))$ be a standard partition and let $\tilde{Y}_f \triangleq {\left\{\tilde{Y}_{f,\theta}\right\}}_{\theta \in \Pi(\Theta)}$ and $\tilde{Y}_H \triangleq {\left\{\tilde{Y}_{H,\phi}\right\}}_{\phi \in \Pi(\Phi)}$ be the $m$-su\-mma\-ries of two unknown BFs in $\Pi(\Theta)$ and $\Pi(\Phi)$, respectively. 
Then for any BFs $B_f \sim \tilde{Y}_f$ and $B_H \sim \tilde{Y}_H$, it holds that $L(H) \le \overline{L}_{\mathbf{\Pi}}(H)$, where
\begin{align}
\label{eq:UpperBound}
\overline{L}_{\mathbf{\Pi}}(H) & \triangleq \frac{\lambda Z_{B_H}}{|J(T)|} + \sum_{j=1}^{N_{\Theta}}{\overline{\mathcal{L}}_{\Theta_j}(H)},  \\
\overline{\mathcal{L}}_{\Theta_j}(H) & \triangleq \mathop{\max}_{0 \le \mathsf{q}_j \le |\Theta_j|} \left[F_{\tilde{Y}_f,\Theta_j}(\mathsf{q}_j) + \Gamma_j(\mathsf{q}_j)\right],
\label{eq:UpperBoundLocal}
\end{align}
and $\Gamma_j(\mathsf{q}_j)$ is the solution to the problem
\begin{align}
\Gamma_j(\mathsf{q}_j) \triangleq \left\{
\begin{array}{lll}
\mathop{\sup}_{\mathsf{V}^0_j, \mathsf{V}^1_j} 
\gamma_j\left(\mathsf{q}_j, \mathsf{V}^0_j, \mathsf{V}^1_j\right),  \\ 
\text{subject to:} \\
\quad \quad 0 \le \mathsf{v}^1_{j,i}\le \left|\Phi^1_{j,i}\right| \ (i=1, \dots, N_r), \\ 
\quad \quad 0 \le \mathsf{v}^0_{j,i}\le \left|\Phi^0_{j,i}\right| 
\end{array} \right.
\label{eq:BigGamma}
\end{align}
with 
\begin{align}
\gamma_j \left(\mathsf{q}_j, \mathsf{V}^0_j, \mathsf{V}^1_j\right) & \triangleq \left(|\Theta_j|-\mathsf{q}_j\right) \alpha \ell^0_j(\mathsf{q}_j, \mathsf{V}^0_j) + \nonumber \\
& + \mathsf{q}_j \log{\left(1 - e^{\alpha \ell^1_j(\mathsf{q}_j, \mathsf{V}^1_j)}\right)} + \nonumber \\
& + \frac{\lambda }{|J(T)|} \sum_{i=1}^{N_r}{F_{\tilde{Y}_H, \Phi_{j,i}}(\mathsf{v}^0_{j,i} + \mathsf{v}^1_{j,i})}
\label{eq:SmallGamma}
\end{align}
($\ell^0_j$ and $\ell^1_j$ are respectively defined in \eqref{eq:DefEll0} and \eqref{eq:DefEll1}).

\begin{proof}
\hspace{-4pt}: It follows from Lemma \ref{lem:SetsCSShapes} and from \eqref{eq:OptimizationProblemSummaries} that $L(H)$ is less than or equal to
\begin{align}
\frac{\lambda Z_{B_H}}{|J(T)|} & +
\mathop{\sup}_{ \begin{array}{c}
\scriptstyle{B_f \sim \tilde{Y}_f} \\ 
\scriptstyle{B_H \sim \tilde{Y}_H} \end{array} } 
\Bigg\{\mathop{\sup}_{\tilde{q},\tilde{v}} \Bigg[\mathop{\sup}_{ \begin{array}{c}
\scriptstyle{q \sim \tilde{q}} \\ 
\scriptstyle{v \sim \tilde{v}} \end{array}} 
\int_{\Theta} {\bigg[q(\vec{x}) \delta_{B_f}(\vec{x}) + } \nonumber \\
& + \frac{\lambda} {|J(T)|} \int_{R_{min}}^{R_{max}}{r^2 v(\vec{x},r) \delta_{B_H}(\vec{x},r) \ dr} + \nonumber \\
& + \log{P\left(q(\vec{x}) | \ell_v(\vec{x})\right)} \bigg] \ d\vec{x} \Bigg] \Bigg\}.
\label{eq:UB1}
\end{align}
Exchanging the order of the \emph{sup} and \emph{max} operations, the second term in \eqref{eq:UB1} is equal to
\begin{align}
\mathop{\max}_{\tilde{q},\tilde{v}} \Bigg\{\mathop{\sup}_{ \begin{array}{c}
\scriptstyle{q \sim \tilde{q}} \\ 
\scriptstyle{v \sim \tilde{v}} \end{array} } 
& \Bigg(\mathop{\sup}_{B_f \sim \tilde{Y}_f} \left[\int_{\Theta} {\delta_{B_f}(\vec{x}) q(\vec{x})\ d\vec{x}} \right] + \nonumber \\
\frac{\lambda} {|J(T)|} \mathop{\sup}_{B_H \sim \tilde{Y}_H} \Bigg[
& \int_{\Theta} {\int_{R_{min}}^{R_{max}} {r^2 v(\vec{x},r) \delta_{B_H}(\vec{x},r) \ dr} \ d\vec{x}}
\Bigg] + \nonumber \\
& \int_{\Theta} {\log{P\left(q(\vec{x})|\ell_v(\vec{x})\right)} \ d\vec{x}} \Bigg) \Bigg\}.
\label{eq:UB2}
\end{align}
Using Lemma \ref{lem:mSummary} the following inequalities are obtained for the first and second terms in \eqref{eq:UB2}:
\begin{align}
\label{eq:UBInequality1}
\mathop{\sup}_{B_f \sim \tilde{Y}_f} & \left[\int_{\Theta} {\delta_{B_f}(\vec{x}) q(\vec{x}) \ d\vec{x}}\right] \le \sum_{j=1}^{N_{\Theta}} {F_{\tilde{Y}_f,\Theta_j}(\mathsf{q}_j)},
\end{align}
\begin{align}
\mathop{\sup}_{B_H \sim \tilde{Y}_H} & \left[\int_{\Theta} {\int_{R_{min}}^{R_{max}}{r^2 v(\vec{x},r) \delta_{B_H}(\vec{x},r) \ dr} \ d\vec{x}}\right] \le \nonumber \\
& \sum_{j=1}^{N_{\Theta}}{\sum_{i=1}^{N_r}{F_{\tilde{Y}_H, \Phi_{j,i}}(\mathsf{v}^0_{j,i} + \mathsf{v}^1_{j,i})}}.
\label{eq:UBInequality2}
\end{align}
And since the rhs's of \eqref{eq:UBInequality1} and \eqref{eq:UBInequality2} do not depend on $q$ or $v$, these terms can be moved out of the supremum on $q$ and $v$ in \eqref{eq:UB2}.

In Lemma \ref{lem:UBProjectionTerm}, on the other hand, we have shown that the last term in \eqref{eq:UB2}, referred to as $\overline{\Lambda}_{\Theta_j}(\tilde{q},\tilde{v})$, can be computed explicitly, and an expression for it is given in \eqref{eq:SolutionBigLambda}. Substituting \eqref{eq:UBInequality1}, \eqref{eq:UBInequality2} and \eqref{eq:SolutionBigLambda} into \eqref{eq:UB2} and rearranging, we obtain
\begin{align}
\sum_{j=1}^{N_{\Theta}} & \ \mathop{\max}_{0 \le \mathsf{q}_j \le |\Theta_j|} \Bigg\{F_{\tilde{Y}_f,\Theta_j}(\mathsf{q}_j) + \\
& \mathop{\sup}_{\tilde{v}} \Bigg[\frac{\lambda} {|J(T)|} \sum_{i=1}^{N_r}{F_{\tilde{Y}_H, \Phi_{j,i}}(\mathsf{v}^0_{j,i} + \mathsf{v}^1_{j,i})} +
\overline{\Lambda }_{\Theta_j}(\mathsf{q}_j,\tilde{v}) \Bigg] \Bigg\} \nonumber. 
\end{align}
which using the definitions in \eqref{eq:BigGamma}, \eqref{eq:SmallGamma}, and \eqref{eq:UpperBoundLocal} is equal to \eqref{eq:UpperBound}, completing the proof. \qed
\end{proof}
\end{theorem}

This theorem says that to compute the upper bound in \eqref{eq:UpperBound} we must maximize the function of the scalar variable $\mathsf{q}_j$ defined in \eqref{eq:UpperBoundLocal}, for each pixel $\Theta_j \in \Pi(\Theta)$. This function is the sum of two terms. In \eqref{eq:F} we provided a formula to efficiently compute the left term. We will show in \S \ref{sec:ComputationGamma} (in the supplementary material) how to efficiently compute the other term (defined by \eqref{eq:BigGamma} and \eqref{eq:SmallGamma}).

The semidiscrete 2D shape $\tilde{q}$, encoded by the quantities $\{ \mathsf{q}_j \}$ estimated in the previous theorem, is the 2D segmentation of the object in the image sphere. Similarly, the semidiscrete 3D shape $\tilde{v}$, encoded by the quantities $\{ \mathsf{v}^0_{j,i} + \mathsf{v}^1_{j,i} \}$ estimated in the previous theorem, is the 3D reconstruction of the object. Note that this segmentation/reconstruction pair corresponds to the upper bound. Recall that another such pair, encoded however by discrete shapes, was obtained jointly with the lower bound in Theorem \ref{theo:LowerBound}.

This concludes the derivation of the upper bound. In the next section we show how to use this bound and the lower bound in Theorem \ref{theo:LowerBound} to solve our problem of interest.

%% file: BoundingMechanism.tex
\section{Bounding mechanism}
\label{sec:BoundingMechanism}

Given a standard partition $(\Pi(\Theta), \Pi(\Phi))$, theorems \ref{theo:LowerBound} and \ref{theo:UpperBound} describe how to compute the bounds corresponding to this partition. These theorems, however, do not say how to progressively refine these bounds. In this section we explain how to construct a sequence of progressively finer partitions for a hypothesis, that will yield, in turn, a sequence of tigh\-ter bounds for the hypothesis.

For each hypothesis $H \in \mathbb{H}$ we define a \emph{pair} of progressively finer sequences of partitions of $\Theta$ and $\Phi$, $\left\{ \Pi^H_{\Theta, k} \right\}$ and $\left\{ \Pi^H_{\Phi, k} \right\}$, respectively. These sequences, in general, \emph{are different for different hypotheses}.
The sequence $\left\{ \Pi^H_{\Theta, k} \right\}$ is defined inductively by
\begin{align}
\label{eq:Theta0H}
\Pi^H_{\Theta, 0} & \triangleq \left\{ \theta^H_0 \right\}, \qquad \text{and} \\
\Pi^H_{\Theta, k+1} & \triangleq \left[\Pi^H_{\Theta, k} \setminus \theta^H_k\right] \cup \pi(\theta^H_k) \quad (k \ge 0),
\label{eq:InductionPartition}
\end{align}
where $\theta^H_k \in \Pi^H_{\Theta, k}$, $\pi(\theta^H_k)$ is a partition of $\theta^H_k$, and the set $\theta^H_0 \subset \Theta$ is the smallest axis-aligned rectangle that contains the projection on the camera retina of the support $\Phi_H$ of the hypothesis $H$ (defined in \S \ref{sec:ShapePriorTerm}).

In order to define the partition $\pi(\theta^H_k)$ in \eqref{eq:InductionPartition} we adopt the following rules. When the ratio of $\theta^H_k$'s height over its width is close to 1, $\pi(\theta^H_k)$ consists of 4 (approximately) equal rectangles obtained by splitting $\theta^H_k$ (ap\-prox\-i\-mate\-ly) in half along each dimension. When this ratio is not close to 1, $\theta^H_k$ is split in such a way as to obtain ``rectangles" that will have a ratio closer to 1 in the next iteration.

The sequence $\left\{ \Pi^H_{\Phi, k} \right\}$, on the other hand, is constructed from the sequence $\left\{\Pi^H_{\Theta, k}\right\}$ and the quantities $\left\{ N^H_{r, k}(\theta) \right\}$ (with $\theta \in \Pi^H_{\Theta, k}$). The quantity $N^H_{r, k}(\theta)$ indicates the number of voxels ``behind" the pixel $\theta \in \Pi^H_{\Theta, k}$ at the $k$-th step (note that different pixels in $\Pi^H_{\Theta, k}$ might have different numbers of voxels behind them). This quantity is initialized to 1, and then each time a pixel is subdivided, its number of voxels is doubled. In other terms, if the pixel chosen to be split in the $k$-th refinement cycle is $\theta^H_k$ (see \eqref{eq:InductionPartition}), the number of voxels behind the different pixels are computed using the following recursion:
\begin{align}
N^H_{r, 0}(\theta^H_0) & \triangleq 1, \\
N^H_{r, k + 1}(\theta) & \triangleq \left\{
\begin{array}{ll}
2 N^H_{r, k}(\theta^H_k), & \quad \text{if} \ \theta \in \pi(\theta^H_k), \\
N^H_{r, k}(\theta), & \quad \text{otherwise}.
\end{array} \right.
\end{align}

Thus, each partition $\Pi^H_{\Phi, k}$ is defined as
\begin{equation}
\Pi^H_{\Phi, k} \triangleq \bigcup_{\theta \in \Pi^H_{\Theta, k}}{\theta \times \Pi^H_{\mathcal{R}, N^H_{r, k}(\theta)}},
\end{equation}
where $\Pi^H_{\mathcal{R}, N_r}$ is as defined in \eqref{eq:PartitionR}-\eqref{eq:r_i}.

Let us now define the sequence of partition pairs as $\left\{ \mathbf{\Pi}^H_k \right\} \triangleq \left\{ \left(\Pi^H_{\Theta, k}, \Pi^H_{\Phi, k}\right) \right\}$. The first pair in this sequence, $\mathbf{\Pi}^H_1$, consists of a single voxel that projects to a single pixel ($\theta^H_0$). Then during each refinement cycle a pixel is split (in general) into four subpixels, and its voxels are split (in general) into eight subvoxels, to generate a new pair $\mathbf{\Pi}^H_k$ of the sequence.

For this new pair, lower and upper bounds for the evidence $L(H)$, respectively $\underline{L}_{\mathbf{\Pi}^H_k}(H)$ and $\overline{L}_{\mathbf{\Pi}^H_k}(H)$, \linebreak[4] could be computed by adding the $N_{\Theta}=|\Pi^H_{\Theta, k}|$ terms in \eqref{eq:LowerBound} and \eqref{eq:UpperBound}, respectively. However, these bounds can be computed more efficiently by exploiting the form of \eqref{eq:InductionPartition}, as
\begin{equation}
\label{eq:LowerBoundInductive}
\underline{L}_{\mathbf{\Pi}^H_{k+1}}(H) = \underline{L}_{\mathbf{\Pi}^H_{k}}(H) - \underline{\mathcal L}_{\theta^H_k}(H) + \sum_{\theta \in \pi(\theta^H_k)} {\underline{\mathcal L}_{\theta}(H)}.
\end{equation}
(A similar expression for the upper bound $\overline{L}_{\mathbf{\Pi}^H_{k+1}}(H)$ can be derived.) Since the partition $\pi(\theta^H_k)$ in general contains just 4 sets, only 4 evaluations of $\underline{\mathcal L}_{\theta}$ and $\overline{\mathcal L}_{\theta}$ are required, using \eqref{eq:LowerBoundLocal} and \eqref{eq:UpperBoundLocal} respectively, to compute the new bounds $\underline{L}_{\mathbf{\Pi}^H_{k+1}}(H)$ and $\overline{L}_{\mathbf{\Pi}^H_{k+1}}(H)$ for $L(H)$.

However, since the number of voxels in a pixel is doubled each time a pixel is subdivided, the cost of computing $\underline{\mathcal L}_{\theta}$ and $\overline{\mathcal L}_{\theta}$ correspondingly increases. To avoid this, we do not subdivide \emph{uniform voxels}, defined as those where the function $\delta_{B_H}$ is uniform in the voxel. Moreover, we join consecutive uniform voxels where $\delta_{B_H}$ takes the same value. In this way the number of voxels in a ray does not grow unboundedly, but rather it remains on the order of the number of consecutive uniform regions in a ray (typically three, namely: empty space, object, empty space). Consequently, the cost of computing $\underline{\mathcal L}_{\theta}$ and $\overline{\mathcal L}_{\theta}$ is also bounded, as is the total cost of a refinement cycle.

While any choice of $\theta^H_k$ from $\Pi^H_{\Theta, k}$ in \eqref{eq:InductionPartition} would result in a new partition pair $\mathbf{\Pi}^H_{k+1}$ that is finer than $\mathbf{\Pi}^H_{k}$, it is natural to choose $\theta^H_k$ to be the set in $\Pi^H_{\Theta, k}$ with the greatest local margin ($\overline{\mathcal L}_{\theta^H_k}(H) - \underline{\mathcal L}_{\theta^H_k}(H)$) since this is the set responsible for the largest contribution to the total margin of the hypothesis ($\overline{L}_{\mathbf{\Pi}^H_k}(H) - \underline{L}_{\mathbf{\Pi}^H_k}(H)$). In order to efficiently find the set $\theta^H_k$ with the greatest local margin, we store the elements of the partition $\Pi^H_{\Theta, k}$ in a priority queue, using their local margin as the priority (a different priority queue is used for each hypothesis). For further details about the refinement of the bounds see \cite[Section 6.3]{Rot10}.

This concludes the description of the BM used to solve our problem. Before presenting in \S \ref{sec:Results} the results obtained with this BM integrated in a H\&B algorithm, we review next the steps of the proposed approach.

%% file: Summary.tex
\section{Summary of the proposed method}
\label{sec:MethodSummary}

Having completed the description of all the components of the proposed approach, we now summarize the steps involved in its execution.
First, during the \emph{initialization stage}, the bounds corresponding to all the hypotheses are initialized. For this purpose, for each hypothesis $H \in \mathbb{H}$, the following steps are performed: 1) the set $\theta_0^H$ defined in \eqref{eq:Theta0H} is estimated; 2) the lower and upper bounds corresponding to this set, $\underline{L}(H)$ and $\overline{L}(H)$, respectively,  are computed using \eqref{eq:LowerBound}-\eqref{eq:LowerBoundLocal} and \eqref{eq:UpperBound}-\eqref{eq:UpperBoundLocal}, respectively; and 3) the set $\theta_0^H$ is inserted in an empty priority queue $\mathbf{\Pi}^H$ using the margin $\overline{L}(H)-\underline{L}(H)$ as the priority.

Then, during each cycle of the the \emph{refinement stage}, the following steps are performed: 1) a single hypothesis $H$ is selected for refinement (as mentioned in \S \ref{sec:SCAlgorithms}); 2) the pixel $\theta^H$ with the largest margin is extracted from the priority queue $\mathbf{\Pi}^H$; 3) this pixel $\theta^H$ is divided into the subpixels $\pi(\theta^H)$; 4) the bounds $\underline{\mathcal{L}}_{\theta}$ and $\overline{\mathcal{L}}_{\theta}$ are computed for each subpixel $\theta \in \pi(\theta^H)$, using \eqref{eq:LowerBoundLocal} and \eqref{eq:UpperBoundLocal}, respectively; 5) each set $\theta \in \pi(\theta^H)$ is reinserted in the priority queue $\mathbf{\Pi}^H$ using the local margin $\overline{\mathcal{L}}_{\theta} - \underline{\mathcal{L}}_{\theta}$ as the priority; and lastly 6) the bounds corresponding to the hypothesis $H$ are updated according to \eqref{eq:LowerBoundInductive} using the subpixel bounds $\underline{\mathcal{L}}_{\theta}$ and $\overline{\mathcal{L}}_{\theta}$. The refinement stage concludes when a hypothesis is proved optimal, or a set of hypotheses is proved indistinguishable.

%% file: Results.tex
\section{Experimental results}
\label{sec:Results}

In this section we show results obtained with the framework described in previous sections. To illustrate the process of discarding hypotheses we first show experiments on a single image (\S \ref{sec:IndividualExperiments}) and then present a quantitative analysis of the results obtained on a dataset containing multiple images (\S \ref{sec:PerformanceAssesment}).

\subsection{Experiments on a single image}
\label{sec:IndividualExperiments}

In order to highlight several of the method's unique characteristics, we first describe the results of three experiments. In these experiments a \emph{known object} is present in the input image and our goal is to estimate its pose (its class is known in this case). For this purpose we define the hypothesis spaces $\mathbb{H}_1$ (used in Experiment 1) and $\mathbb{H}_2$ (used in experiments 2 and 3) and use our framework to select the hypothesis $H \in \mathbb{H}_i$ ($i = 1, 2$) that maximizes the evidence $L(H)$.

\begin{figure}[b]
\begin{center}
\includegraphics[width=\columnwidth]{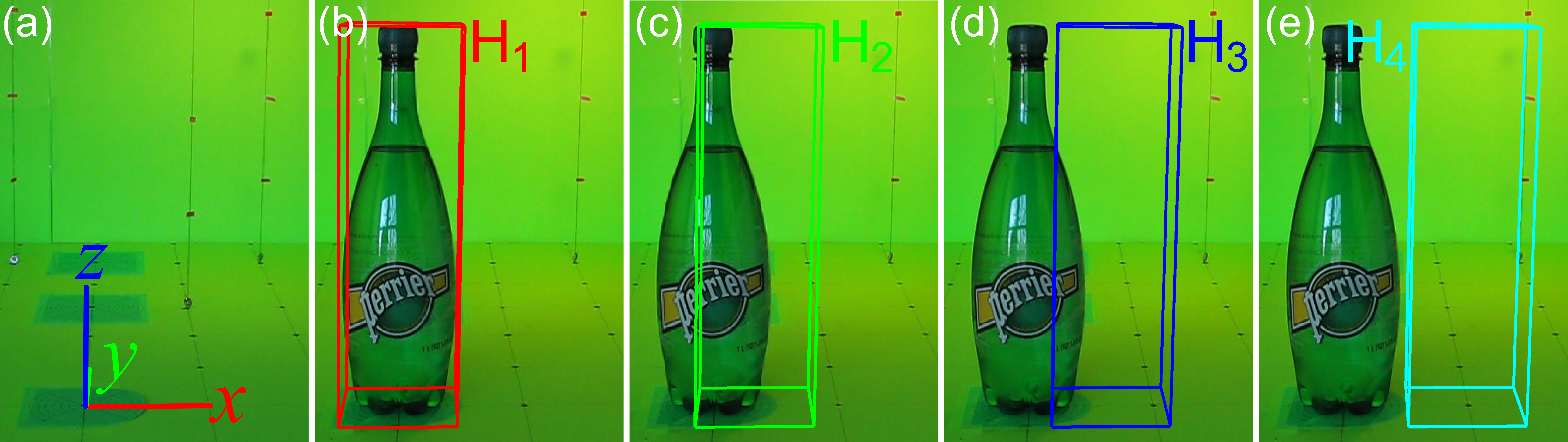}
\end{center}
\vspace{-5pt}
\caption{(a) Coordinate axes in the WCS. Each axis starts at the origin and is 10cm long. (b-e) The four hypotheses in $\mathbb{H}_1$ proposed to ``explain" the (same) input image. The support $\Phi_H$ of each hypothesis is indicated by the overlaid 3D box. Only a part of the image is shown, the actual image is larger. All images used in this work contain $640 \times 480$ pixels.}
\vspace{-0pt}
\label{fig:SetupSimpleExperiment}
\end{figure}

As mentioned before a hypothesis consists of an object class $K$ and a transformation (or pose) $T$. Hence, the sets
$\mathbb{H}_1$ and $\mathbb{H}_2$ are defined as $\mathbb{H}_i \triangleq \{(K_{object}, T): T \in \mathbb{T}_i\}$ ($i = 1, 2$), where $K_{object}$ denotes the class comprising only the known object (\ie, there is no uncertainty in the shape prior), and the $\mathbb{T}_i$'s are sets of transformations containing different horizontal translations. 
To formally define these sets, let us denote by $\vec{i}$, $\vec{j}$ and $\vec{k}$ the vectors that are 1cm long in the direction of the $x$, $y$ and $z$ axes (Fig. \ref{fig:SetupSimpleExperiment}a), respectively, and define the transformation 
\begin{align}
\label{eq:TranslationXY}
T_{t_x t_y} (\vec{X}) \triangleq \vec{X} + t_x \ \vec{i} + t_y \ \vec{j}.
\end{align}
The sets $\mathbb{T}_1$ and $\mathbb{T}_2$ are then defined as $\mathbb{T}_1 \! \triangleq \! \big\{T_{t_x t_y}\!: t_x \! \in \! \{0\!:\!3.2\!:\!9.6\}, t_y \! = \! 0 \big\}$ and $\mathbb{T}_2 \! \triangleq \! \big\{T_{t_x t_y}\!: t_x \! \in \! \{-15\!:\!0.5\!:\!15 \}, t_y \! \in \! \{-35\!:\!0.5\!:\!20\} \big\}$, where $\{E_L\!:\!\Delta\!:\!E_R\}$ denotes the set $\{E_L \! + \! k \Delta \!: \! k \! \in \! \mathbb{N}, 0 \!\le\! k \!\le\! (E_R \!-\! E_L) / \Delta \}$.
In other words, $\mathbb{H}_1$ contains four equispaced hypotheses along the $x$ direction (depicted in Fig. \ref{fig:SetupSimpleExperiment}b-e) and $\mathbb{H}_2$ contains $6,771$ hypotheses produced by combining 61 translations in the $x$ direction and 111 translations in the $y$ direction. 
The transformations must be interpreted as ``moving'' the object away from the ground truth position (hence, the method should ideally select the hypothesis corresponding to the identity transformation).

\begin{figure}
\vspace{-0pt}
\begin{center}
\includegraphics[width=\columnwidth]{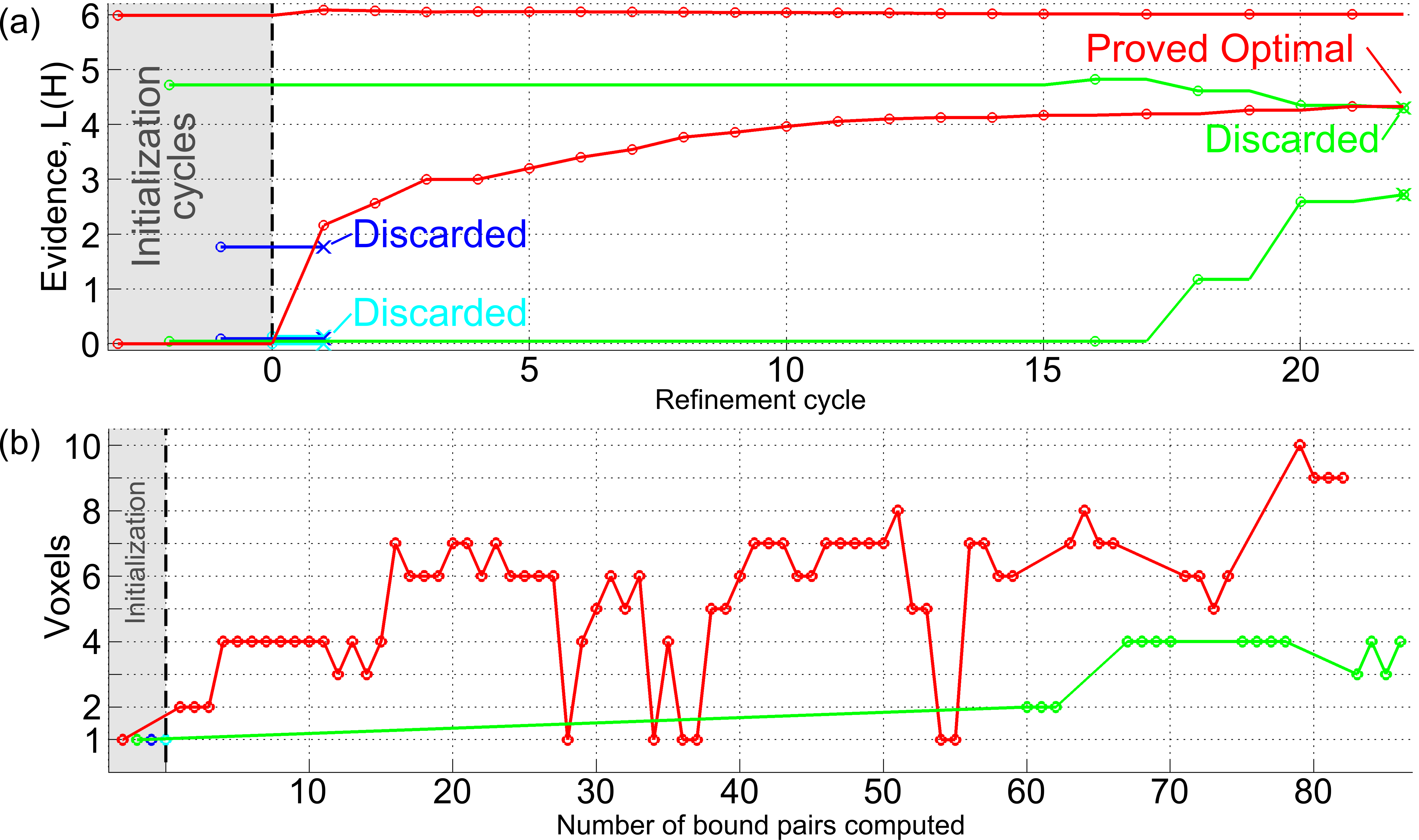}
\end{center}
\vspace{-10pt}
\caption{(a) Evolution of the bounds of the four hypotheses in Fig. \ref{fig:SetupSimpleExperiment}b-e. The bounds of one hypothesis are represented by the two lines of the same color, which is also the color of the hypothesis in Fig. \ref{fig:SetupSimpleExperiment}. Cycles in which a hypothesis is selected for refinement or discarded are indicated by the markers `$\circ$' and `$\times$', respectively. Initialization cycles (on or before cycle 0) are indicated by the gray background. The \emph{blue} and \emph{cyan} hypotheses are discarded during the $1^{st}$ refinement cycle. The \emph{green} hypothesis is discarded during the $22^{nd}$ cycle, proving then that the \emph{red} hypothesis is optimal. (b) Number of voxels processed, $N^H_{r,k}(\theta^H_k)$, for each pixel $\theta^H_k$ processed (see \S \ref{sec:BoundingMechanism}).}
\label{fig:SimpleEvolutionBounds}
\vspace{-15pt}
\end{figure}

\paragraph{\textbf{Experiment 1.}}
When the proposed method was applied to a noiseless BF with the hypotheses in $\mathbb{H}_1$, the bounds corresponding to these hypotheses evolved as depicted in Fig. \ref{fig:SimpleEvolutionBounds}a. In this case one hypothesis was proven to be optimal after 22 refinement cycles. 
Fig. \ref{fig:SimpleEvolutionBounds}b shows the number of voxels processed during each computation of bounds (\ie, each time \eqref{eq:LowerBoundLocal} and \eqref{eq:UpperBoundLocal} were evaluated). Since elements in the partition $\Pi^H_{\Theta, k}$ are split (in general) in 4 during each refinement cycle (as explained in \S \ref{sec:BoundingMechanism}), the number of bound computations is roughly four times the number of refinement cycles. It can be seen that as the bounds of one hypothesis $H$ are refined, the elements of $\Pi^H_{\Theta, k}$ become smaller and tend to initially have a larger number of voxels projecting to them. As explained in \S \ref{sec:BoundingMechanism}, the number of voxels initially doubles (steps 0-11 in the figure), and then it grows more slowly or even decreases (steps 12 and after) whenever we merge uniform voxels. For these reasons refinement cycles became more costly up to a point, and then they plateau.

\begin{figure}[th]
\vspace{-10pt}
\begin{center}
\includegraphics[width=\columnwidth]{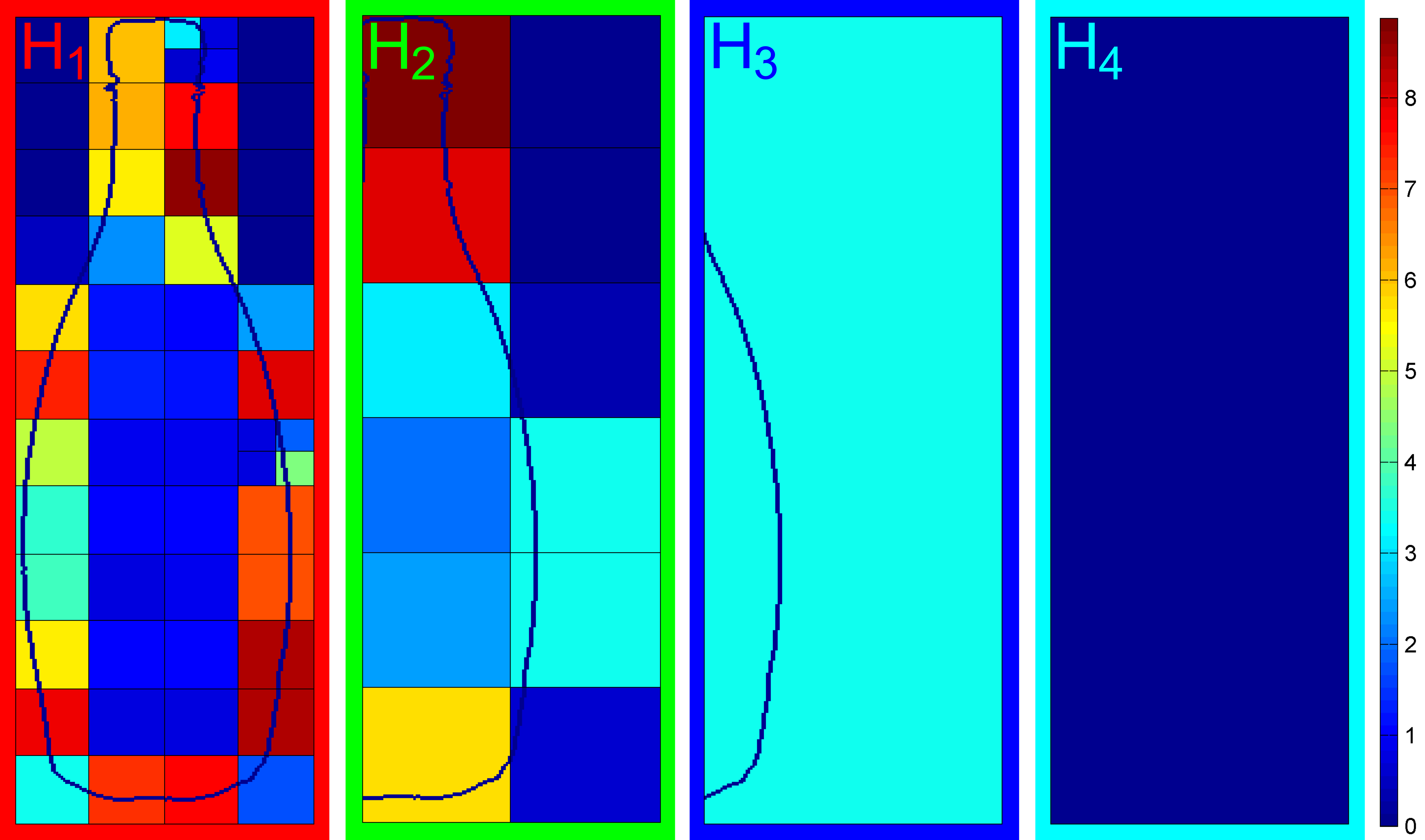}
\end{center}
\vspace{-10pt}
\caption{Partitions obtained after 22 refinement cycles. These partitions are sufficient to find the best hypothesis among the four defined in Fig. \ref{fig:SetupSimpleExperiment}. The color of each partition element represents the margin of the element divided by its area. The silhouettes of the object in the input image are displayed for reference only. Note that the blue and cyan hypotheses are discarded by looking at a single partition element (and computing a single pair of bounds).}
\label{fig:SimpleFinalPartitions}
\vspace{-20pt}
\end{figure}

Fig. \ref{fig:SimpleFinalPartitions} shows the state of the partitions $\Pi^1_{\Theta, end}, \dots,$ $ \Pi^4_{\Theta, end}$, corresponding to the four hypotheses in $\mathbb{H}_1$, when the process terminated. These partitions are sufficient to discriminate between the hypotheses and the additional available resolution does not influence the computational load. Thus \emph{the computational load depends on the task at hand (through the set of hypotheses to be discriminated) and not on the resolution of the input image}.

\begin{figure} \sidecaption
\vspace{10pt}
\includegraphics[width=0.55\columnwidth]{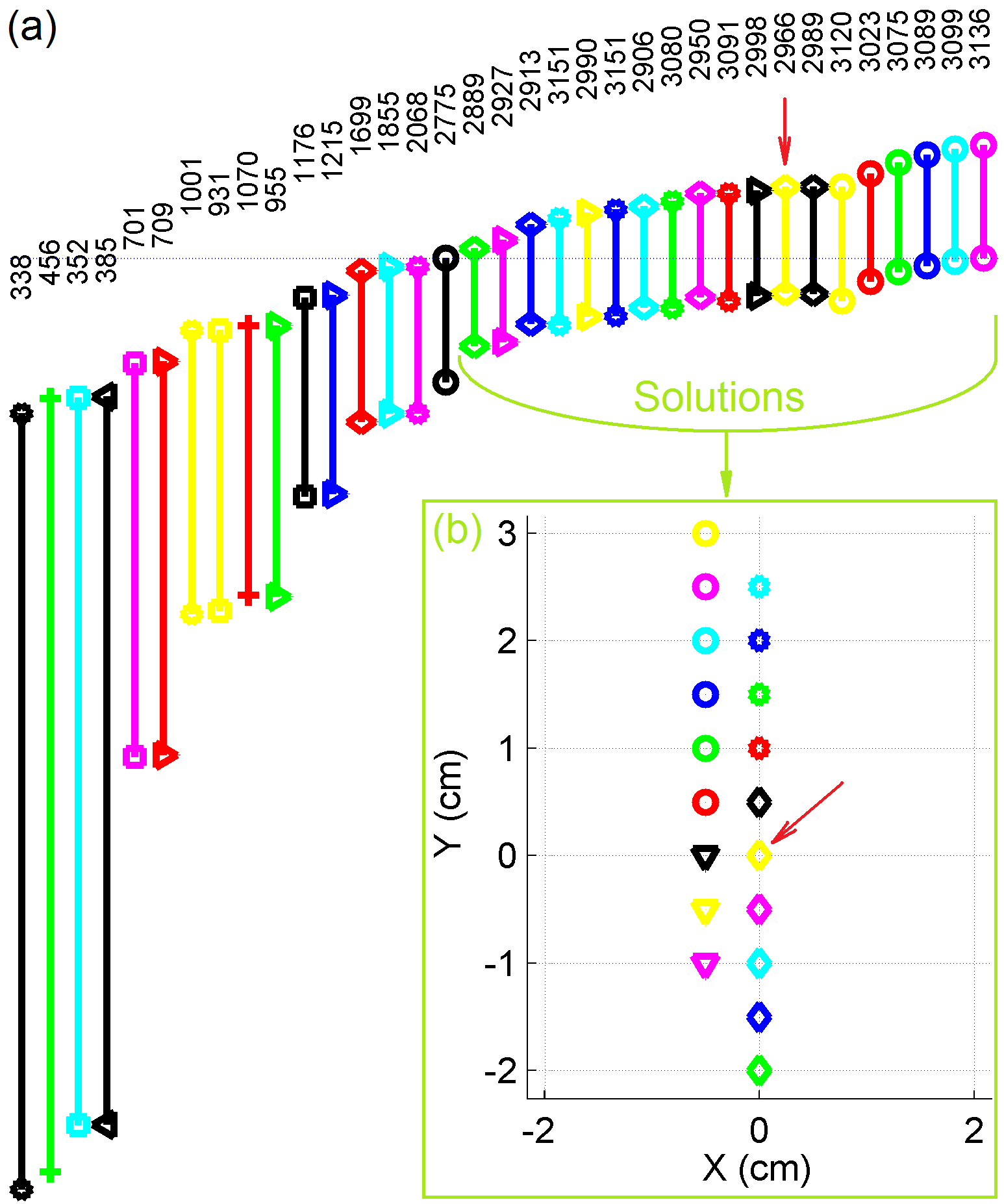}
\caption{(a) Final bounds for the subset of the best hypotheses. The number of refinement cycles allocated to each hypothesis is indicated above each hypothesis. (b) The translation corresponding to each solution. Hypotheses can be identified by their color and marker. The ground truth is indicated by the red arrow.}
\label{fig:ComplexFinalBounds}
\vspace{-15pt}
\end{figure}

\begin{figure}[b]
\begin{center}
\includegraphics[width=0.65\columnwidth]{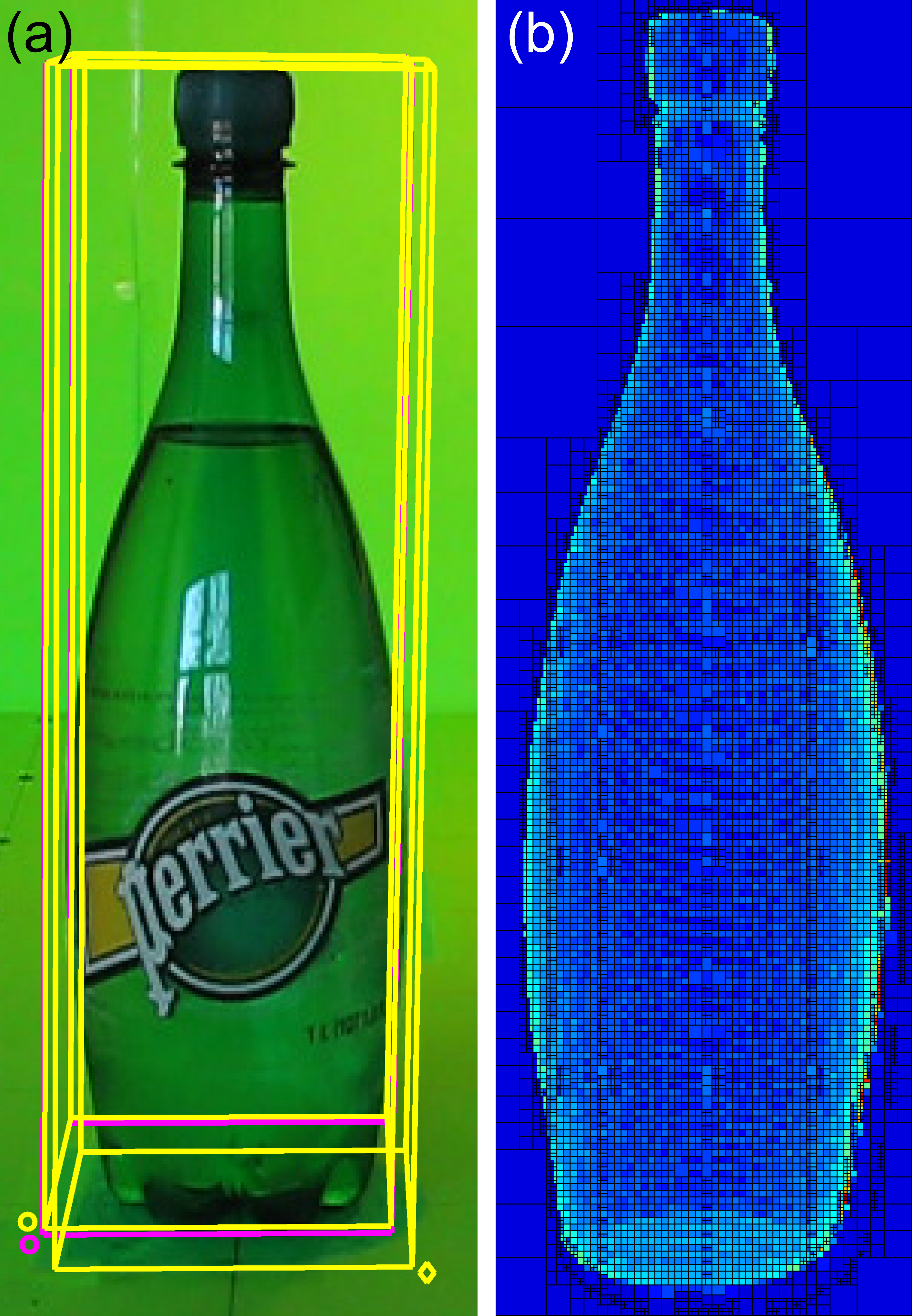}
\end{center}
\vspace{-5pt}
\caption{(a) Three of the hypotheses in $\mathbb{H}_2$ that could not be distinguished given the current input: the ground truth (\textcolor{yellow}{$\diamond$}), the best hypothesis (having the greatest upper bound, \textcolor{pink}{$\circ$}), and the one farthest away from the ground truth (\textcolor{yellow}{$\circ$}). (b) The final partition obtained for the best hypothesis (the one with the highest upper bound, indicated by the marker \textcolor{pink}{$\circ$} in Fig. \ref{fig:ComplexFinalBounds}). Colors indicate the margin of each partition element. Note that most of the work is performed around the edges of the image or the prior, and that uniform pixels/voxels are not further subdivided.}
\label{fig:SolutionsExp2_57}
\vspace{-17pt}
\end{figure}

\paragraph{\textbf{Experiment 2.}}
Fig. \ref{fig:ComplexFinalBounds} shows the final bounds obtained for the best hypotheses in $\mathbb{H}_2$ when our method was used to compute the bounds for this set. Note that at termination time there are still 19 hypotheses in the active set $\mathbb{A}$ that cannot be further refined (recall that a hypothesis is in $\mathbb{A}$ if its upper bound is greater than the maximum lower bound). These hypotheses are indistinguishable given the current input (as defined in \S \ref{sec:SCAlgorithms}) and will be referred to as \emph{solutions}. Three solutions are depicted in Fig. \ref{fig:SolutionsExp2_57}a: the ground truth solution; the best solution (\ie, the one having the greatest upper bound); and the solution farthest away from the ground truth.

In order to quantify the quality of the set of solutions $\mathbb{A}$ we define for each parameter $t$ of the transformations the \emph{bias of $t$} and the \emph{standard deviation of $t$}, respectively as
\begin{align}
\mu_t & \triangleq \frac{1}{|\mathbb{A}|} \sum_{H \in \mathbb{A}}{t_H - t_{true}} , \quad \quad \text{and} \\
\sigma_t & \triangleq \sqrt{ \frac{1}{|\mathbb{A}|} \sum_{H \in \mathbb{A}}{(t_H - t_{true})^2}},
\end{align} 
where $t_H$ is the value of the parameter $t$ corresponding to the hypothesis $H \in \mathbb{A}$ and $t_{true}$ is the true value of the parameter $t$. The values of these quantities obtained for this experiment are summarized in Table \ref{tab:Experiment2Errors}.

\begin{table}[h]
\vspace{-10pt}
\caption{Pose estimation errors for a \emph{known} object in a \emph{noiseless} image.}
\label{tab:Experiment2Errors}
\begin{tabular}{|
@{}>{\centering\arraybackslash}m{0.7in}@{}|
@{}>{\centering\arraybackslash}m{0.7in}@{}|
@{}>{\centering\arraybackslash}m{0.7in}@{}|
@{}>{\centering\arraybackslash}m{0.7in}@{}|
@{}>{\centering\arraybackslash}m{0.47in}@{}|} \hline 
$\mu_{t_x}$ (cm) & $\mu_{t_y}$ (cm) & $\sigma_{t_x}$ (cm) & $\sigma_{t_y}$ (cm) & $|\mathbb{A}|$ \\ \hline 
-0.24 & 0.61 & 0.34 & 1.54 & 19 \\ \hline
 \end{tabular}
\vspace{-10pt}
\end{table}

In the ideal case the indistinguishability of a group of hypotheses is a consequence solely of the resolution of the input image. In practice, however, other factors enter into play, including the inaccuracy of the camera model and calibration, the noise corrupting the input image, the fact that $m$ (in the $m$-summaries) is finite, and the approximation in the computation of the summaries (explained in \S \ref{sec:ComputingSummaries} in the supplementary material).

\begin{figure}
\includegraphics[width=\columnwidth]{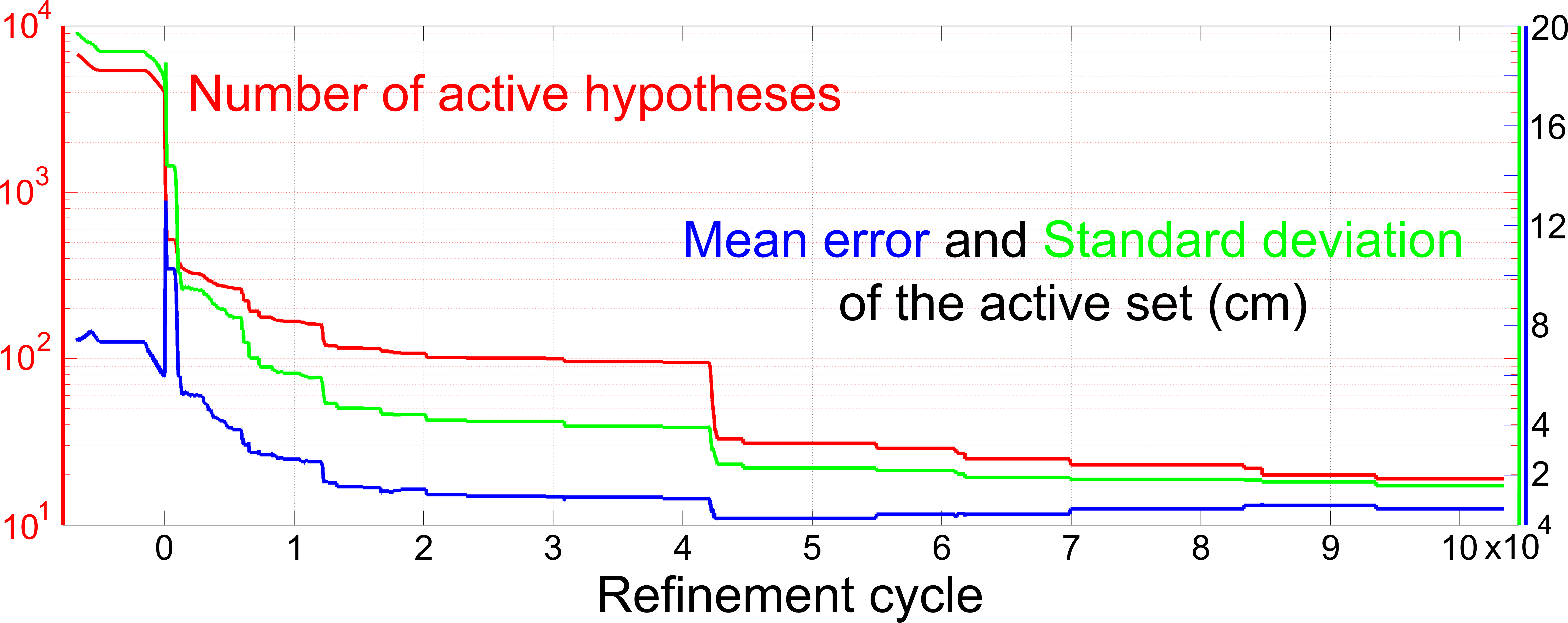}
\caption{Number of active hypotheses (red), mean error (blue) and standard deviation (green) of the active set vs. refinement cycles performed.}
\vspace{-10pt}
\label{fig:SolutionsExp2_4}
\end{figure}

As the bounds are refined some hypotheses are discarded, while others remain in the active set. Fig. \ref{fig:SolutionsExp2_4} shows the number of active hypotheses remaining after each refinement cycle, as well as the \emph{mean error} and the \emph{standard deviation} ($\sqrt{{\mu_{t_x}}^2 + {\mu_{t_y}}^2}$ and $\sqrt{{\sigma_{t_x}}^2 + {\sigma_{t_y}}^2}$, respectively) of the active set after each refinement cycle. Note that both the number of hypotheses in the active set and its standard deviation are non-increasing functions. The mean error of the active set, however, increases at times because the hypotheses ``on one side" or ``on the other side" of the ground truth are not discarded at exactly the same time.

Fig. \ref{fig:SolutionsExp2_57}b shows the final partition obtained for the best hypothesis. Note that the partition is finest in the area around the edges of the bottle, and coarsest in the area ``outside'' the bottle. This behavior emerges automatically (\ie, it does not have to be explicitly encoded in the framework) as the algorithm greedily reduces the uncertainty of each hypothesis by subdividing the partition elements with the greatest margin. Note also that the partition inside the silhouette of the bottle is finer than outside of it. This is because pixels ``inside" the bottle, even if they are not near the edges, still have to be divided in order to divide their voxels and obtain an accurate reconstruction (since in the current implementation voxels are divided in depth only when the pixel they project to is also divided).

\begin{figure}[t]
\includegraphics[width=\columnwidth]{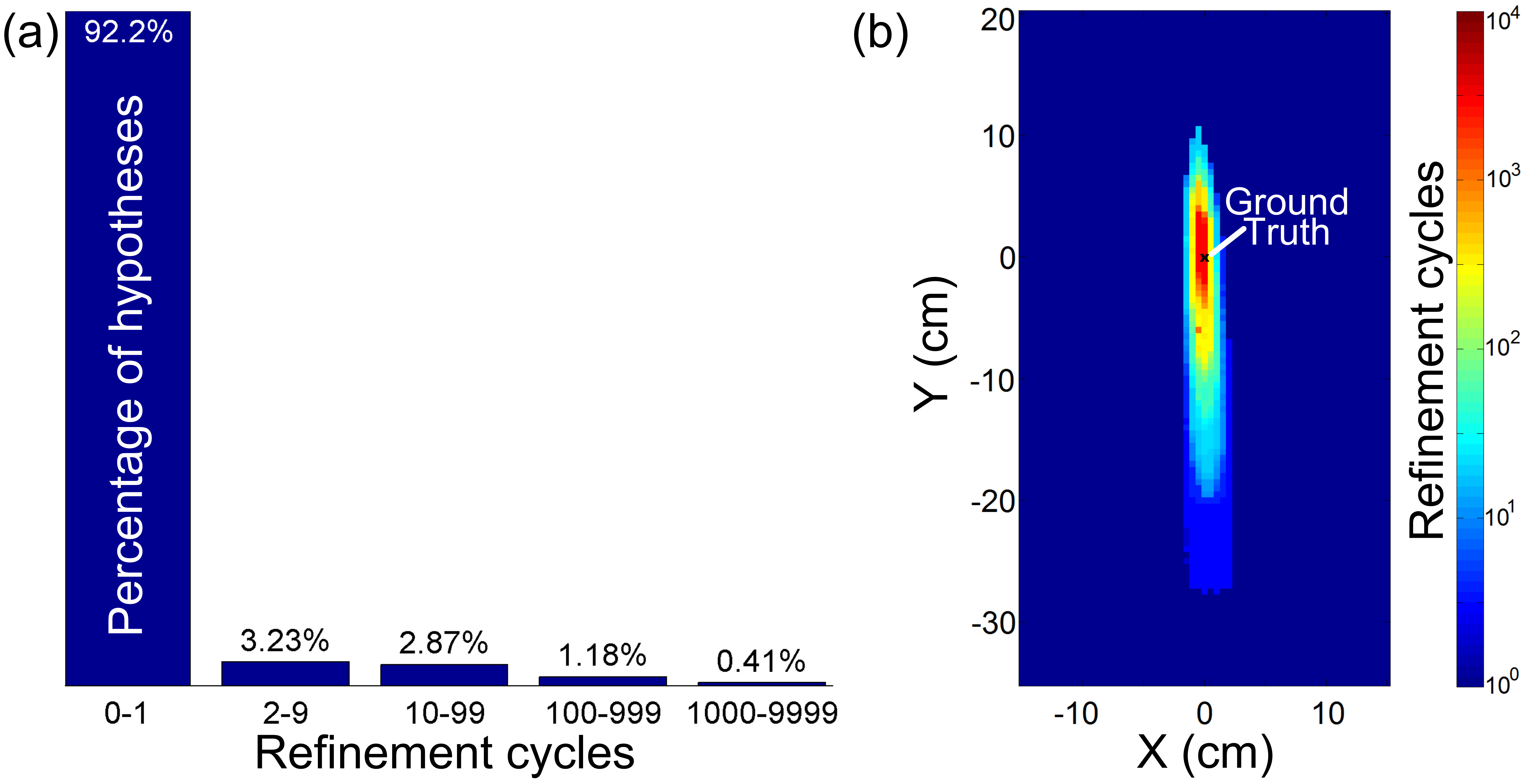}
\vspace{-10pt}
\caption{(a) Histogram of refinement cycles allocated per hypothesis. Most computation is allocated to a few hypotheses. (b) Refinement cycles allocated to each hypothesis in the hypothesis space $\mathbb{H}_2$ (note the logarithmic scale). The position of the true hypothesis is indicated by the marker `$\times$'. The hypotheses gathering most computation surround the true hypothesis.}
\vspace{-15pt}
\label{fig:ComplexWorkPerHypothesis}
\end{figure}

Fig. \ref{fig:ComplexWorkPerHypothesis} shows how the computation is distributed among the hypotheses in $\mathbb{H}_2$. It can be seen in Fig. \ref{fig:ComplexWorkPerHypothesis}a that most hypotheses (92.2\%) only require 0/1 refinement cycles, while only a few (0.41\%) had to be processed at the finest resolution. (The exact number of refinement cycles allocated to each hypothesis in the set $\mathbb{A}$ is indicated above each hypothesis in Fig. \ref{fig:ComplexFinalBounds}a.)
Fig. \ref{fig:ComplexWorkPerHypothesis}b shows that the hypotheses that require most computation surround the ground truth, however, not isotropically: hypotheses in the ground truth's line of sight are harder to distinguish from it (compare $\sigma_{t_x}$ and $\sigma_{t_y}$ in Table \ref{tab:Experiment2Errors} and look at the shape of the active set in Fig. \ref{fig:ComplexFinalBounds}b).
Hence, most hypotheses are discarded with minimal computation while only the most promising hypotheses gather most of the computation. This is the source of the computational gain of our algorithm.

It is difficult to accurately quantify this computational gain \emph{in general}, since it strongly depends on the position of the object in the input image, the level and type of noise affecting this image (to be discussed later in this section), the task at hand (through the set of hypotheses that must be discriminated) and the shape priors used. However, for illustration purposes only, it is possible to quantify this gain for the current experiment by comparing the number of voxels processed by our approach, versus a na\"{i}ve approach defined as follows. Suppose that to select the best hypothesis we directly compute the evidence $L(H)$, using \eqref{eq:Evidence}, for each one of the $6,771$ hypotheses in $\mathbb{H}_2$. Note that this entails processing, \emph{for each hypothesis}, all the pixels and voxels in the relevant parts of the image and world space (\ie, those presumed to contain the object). In this particular example the relevant part of the image contains approximately 50k pixels, and the relevant part of the world contains approximately 12.8M voxels ($= 50\text{k pixels} \times 256 \text{ radii}$). Hence the na\"{i}ve approach would need to process approximately 339M pixels ($=6,771 \times 50\text{k}$) and 86.7G voxels ($=6,771 \times 50\text{k} \times 256$).

In contrast, in the proposed approach only 440k pixels (\ie, elements of $\Pi(\Theta)$) and 3.6M voxels (\ie, elements of $\Pi(\Phi)$) are processed. This is a 770-fold reduction of the number of pixels processed, and a 24k-fold reduction of the number of voxels processed. On the other hand, if the accuracy given by the set of hypotheses $\mathbb{H}_1$ is sufficient for a particular application, our method would only need to process 90 pixels and 437 voxels (these voxels can be directly counted in Fig. \ref{fig:SimpleEvolutionBounds}b). This yields a 4M-fold reduction in the number of pixels to process and a 200M-fold reduction in the number of voxels to process, a significant efficiency gain. For this reason we said that the computation depends on the task, in particular in the precision (in the class or pose) required by the task. Moreover, to obtain this gain it is not necessary to down-sample the input image \emph{a priori} when the task might not even be defined yet; the framework automatically uses the appropriate resolution. Interestingly, the pixels and voxels processed for one hypothesis in the na\"{i}ve approach are all disjoint, while those processed in the proposed approach are not: pixels and voxels processed later lie within pixels and voxels processed earlier.

\begin{figure}[b]
\vspace{-10pt}
\includegraphics[width=\columnwidth]{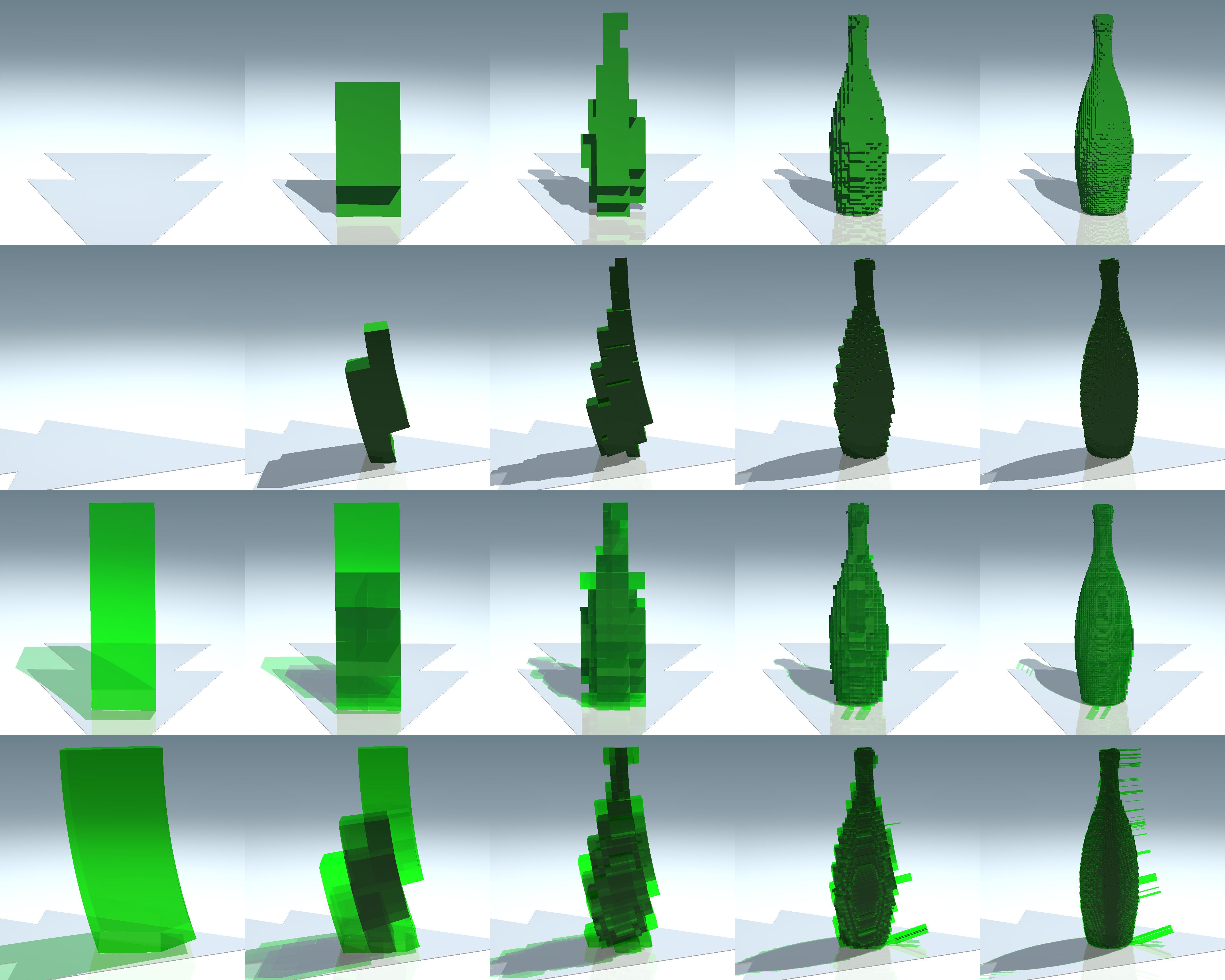}
\vspace{-10pt}
\caption{Different stages of the reconstructions obtained while computing the bounds of the best hypothesis. These reconstructions are given by the discrete shape $\hat{v}$ ($1^{st}$ and $2^{nd}$ rows) and the semidiscrete shape $\tilde{v}$ ($3^{rd}$ and $4^{th}$ rows). Each column contains two renderings of each of these shapes after (from left to right) 1, 10, 100, $1,000$ and $5,000$ refinement cycles. Each rendering was obtained from a different viewpoint.  In one rendering the camera was located in the same pose as in the original input image ($1^{st}$ and $3^{rd}$ rows), while in the other the rendering camera was rotated $90^\circ$ to the left of the object ($2^{nd}$ and $4^{th}$ rows). The triangles on the floor point in both cases towards the original camera. In the case of $\tilde{v}$ the transparency of each voxel indicates the fraction of the voxel that is full, \ie, $\tilde{v}(\Phi_{j,i}) / |\Phi_{j,i}|$. A perfectly transparent voxel indicates 0\%, while a perfectly opaque voxel indicates 100\%. Shadows and reflections were added for visualization purposes only.}
\vspace{-0pt}
\label{fig:ReconstructionProcess}
\end{figure}

Recall that two 3D shapes are obtained while computing the bounds of a hypothesis: a discrete 3D shape $\hat{v}$ and a semidiscrete 3D shape $\tilde{v}$ are obtained when the lower and upper bounds are computed using \eqref{eq:ReconstructionLower} and \eqref{eq:BigGamma}, respectively. These shapes are progressively refined as the bounds are refined (Fig. \ref{fig:ReconstructionProcess}). These shapes are initially defined on a partition containing a single voxel (left column), which is then refined to contain hundreds of thousands of voxels after 5,000 refinement cycles (right column). It can be seen that the two shapes $\hat{v}$ and $\tilde{v}$ get progressively ``closer to each other" and approach the continuous shape $v$ that solves \eqref{eq:Evidence}. In \S \ref{sec:PerformanceAssesment} we will show reconstructions obtained when neither the object nor its class are known.

\paragraph{\textbf{Experiment 3.}}
Experiments 1 and 2 assumed that a noiseless BF with the silhouette of the object can be obtained from the input image. In real scenarios, however, this is rarely the case. Camera noise, shadows and reflections, among many factors, cause the resulting BF to rarely be the exact silhouette of an object, but rather to have errors.

\begin{figure}[b]
\vspace{-10pt}
\includegraphics[width=\columnwidth]{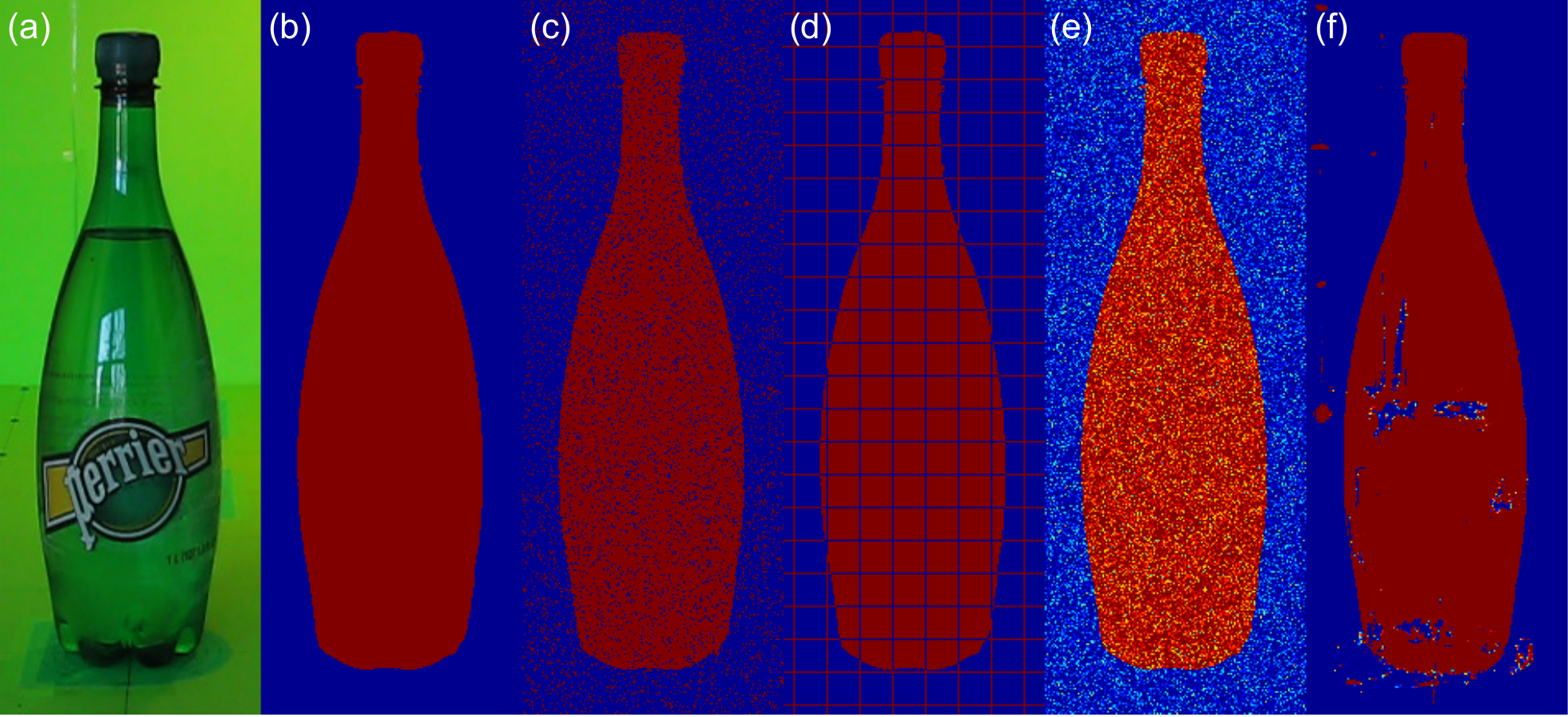}
\vspace{-10pt}
\caption{Different input images considered in this work. (a) Original RGB image (only part shown). (b) Corresponding ``ground truth" BF. (c-e) Ground truth BF corrupted with salt and pepper noise, $\mathcal{SP}(0.1)$, structured noise, $\mathcal{S}(20)$, and Gaussian noise, $\mathcal{N}(0,0.2^2)$, respectively. (f) BF obtained by background subtraction.}
\vspace{-0pt}
\label{fig:NoiseLevels}
\end{figure}

In order to quantify the effect of noise on the performance of our framework we look at how different kinds and levels of noise affect the quality of the pose estimation. For this purpose we run our approach exactly as in Experiment 2, except that we degrade the image with noise. For simplicity and to be able to precisely control the amount of noise introduced, we add synthetic noise directly to the BF corresponding to the ground truth segmentation of the input image (Fig. \ref{fig:NoiseLevels}b), rather than to the RGB input image itself (Fig. \ref{fig:NoiseLevels}a). 

Three kinds of noise have been considered: 1) salt and pepper noise (Fig. \ref{fig:NoiseLevels}c), $\mathcal{SP}(P)$, produced by changing, with probability $P$, the success rate of a pixel $\vec{x}$ from $p(\vec{x})$ to $1-p(\vec{x})$; 2) structured noise (Fig. \ref{fig:NoiseLevels}d), $\mathcal{S}(\ell)$, produced by changing the success rate from $p(\vec{x})$ to $1-p(\vec{x})$ for each pixel $\vec{x}$ in rows and colums that are multiples of $\ell$; and 3) additive, zero mean, white Gaussian noise with standard deviation $\sigma$ (Fig. \ref{fig:NoiseLevels}e), denoted by $\mathcal{N}(0,\sigma^2)$. When adding Gaussian noise to a BF some values end up outside the interval $[0,1]$. In such cases we trim these values to the corresponding extreme of the interval.

In addition to these types of noise, to simulate a more realistic scenario, we also consider BFs produced by a simple background subtraction algorithm (as described by \eqref{eq:BernoulliFieldInput}). The distribution of features for a pixel $\vec{x}$ in the \texttt{Background}, $p_{\vec{x}}\left(f(\vec{x}) | q(\vec{x})=0\right)$, is chosen to be a Gaussian probability density function whose mean and variance are learned from an image of the scene without the object (one such density is learned \emph{for each pixel} $\vec{x}$). The distribution of colors for pixels in the \texttt{Foreground}, $p\left(f(\vec{x})|q(\vec{x})=1\right)$, is represented by a mixture of Gaussians whose parameters are learned from a few pixels on the object that we manually select (only one density is learned \emph{for all foreground pixels}). The details of the background subtraction algorithm are not important here. Our algorithm uses prior geometric 3D information to improve \emph{any BF (or segmentation)} obtained with \emph{any} algorithm, as long as it is given as a foreground probability map, \ie, a 2D BF. In fact, the BFs used in the following experiments purposefully contain artifacts to resemble realistic scenarios. A subset of these BFs is shown in the first row of figures \ref{fig:3DReconstructionsUpper}, \ref{fig:3DReconstructionsLower} and \ref{fig:3DReconstructionsAdditional}.

The results of these experiments are summarized in Table \ref{tab:NoisyLocalizationErrors}. To reduce the variation in the results produced by the variation in the noise itself, in the table we report the average of each quantity over 10 runs of the algorithm. For convenience the results of Experiment 2 (the noiseless case) are also included in this table.

\begin{table}[h]
\caption{Pose estimation errors for a \emph{known} object in a \emph{noisy} image.}
\label{tab:NoisyLocalizationErrors}
\begin{tabular}{|m{0.73in}|
@{}>{\centering\arraybackslash}m{0.47in}@{}|
@{}>{\centering\arraybackslash}m{0.47in}@{}|
@{}>{\centering\arraybackslash}m{0.47in}@{}|
@{}>{\centering\arraybackslash}m{0.47in}@{}|
@{}>{\centering\arraybackslash}m{0.47in}@{}|} \hline
Noise & $\mu_{t_x}$ (cm) & $\mu_{t_y}$ (cm) & $\sigma_{t_x}$ (cm) & $\sigma_{t_y}$ (cm) & $|\mathbb{A}|$ \\ \noalign{\hrule height 1pt}
No noise                          & -0.24 &  0.61 & 0.34 & 1.54 &  19.0 \\ \noalign{\hrule height 1pt}
$\mathcal{SP}(0.05)$      & -0.17 & -1.25 & 0.29 & 2.08 &  17.1 \\ \hline
$\mathcal{SP}(0.10)$      &  0.00 & -4.66 & 0.00 & 4.74 &   5.7 \\ \noalign{\hrule height 1pt}
$\mathcal{S}(40)$           & -0.20 & -1.24 & 0.36 & 2.16 &  13.0 \\ \hline
$\mathcal{S}(20)$           & -0.14 & -4.28 & 0.16 & 4.46 &   6.2 \\ \noalign{\hrule height 1pt}
$\mathcal{N}(0, 0.10^2)$ & -0.25 &  0.03 & 0.35 & 2.43 &  33.1 \\ \hline
$\mathcal{N}(0, 0.20^2)$ & -0.21 & -0.51 & 0.51 & 3.10 &  66.9 \\ \noalign{\hrule height 1pt}
Back. sub.                        & -0.28 &  0.00 & 0.37 & 1.46 &  20.0 \\ \hline
\end{tabular}
\vspace{-15pt}
\end{table}

It can be observed that the estimation of the position of the object in the $x$ direction was relatively unaffected by these types and levels of noise (see columns labeled $\mu_{t_x}$ and $\sigma_{t_x}$). Similarly, the errors of the position estimate in the $y$ direction were not affeted by the Gaussian noise, but they significantly increased for the other types of synthetic noise (see columns labeled $\mu_{t_x}$ and $\sigma_{t_x}$). The results in the background subtraction case, on the other hand, were in every case at the same level as the noiseless case.

Table \ref{tab:NoisyPerformance} contains the total number of pixels ($\tau$) and voxels ($\nu$) processed by the algorithm under each noise condition. This table indicates that BFs obtained by background subtraction having the level of artifacts \linebreak[4] shown in Fig. \ref{fig:NoiseLevels}f require a slight amount of additional computation. BFs corrupted by higher levels of synthetic noise, on the other hand, require significantly higher amounts of computation  (in particular for salt and pepper and structured noise). The additional computation is required because when the input images are corrupted by noise, more cycles have to be spent before the group of hypotheses that will ultimately become solutions can be identified. In other words, more cycles are spent refining hypotheses that are eventually discarded.

\begin{table}[h]
\vspace{-15pt}
\caption{Amount of computation needed to estimate the pose of a known object in a noisy image.}
\label{tab:NoisyPerformance}
\begin{tabular}{|m{1.55in}|@{}>{\centering\arraybackslash}m{0.78in}@{}|@{}>{\centering\arraybackslash}m{0.78in}@{}|} \hline
Noise & $\tau \ (\times 10^6)$ & $\nu \ (\times 10^6)$  \\ \noalign{\hrule height 1pt}
No noise                          &  0.44 &   3.63 \\ \noalign{\hrule height 1pt}
$\mathcal{SP}(0.05)$      &   4.00 &   24.66 \\ \hline
$\mathcal{SP}(0.10)$      &  15.92 &   94.71 \\ \noalign{\hrule height 1pt}
$\mathcal{S}(40)$           &   2.41 &   14.66 \\ \hline
$\mathcal{S}(20)$           &   8.76 &   49.52 \\ \noalign{\hrule height 1pt}
$\mathcal{N}(0, 0.10^2)$ &  0.96 &    7.98 \\ \hline
$\mathcal{N}(0, 0.20^2)$ &  2.54 &   20.00 \\ \noalign{\hrule height 1pt}
Background subtraction     &  0.65 &    5.06 \\ \hline
\end{tabular}
\vspace{-15pt}
\end{table}

\subsection{Assessment of the performance on a larger dataset}
\label{sec:PerformanceAssesment}
\begin{figure}
\includegraphics[width=\columnwidth]{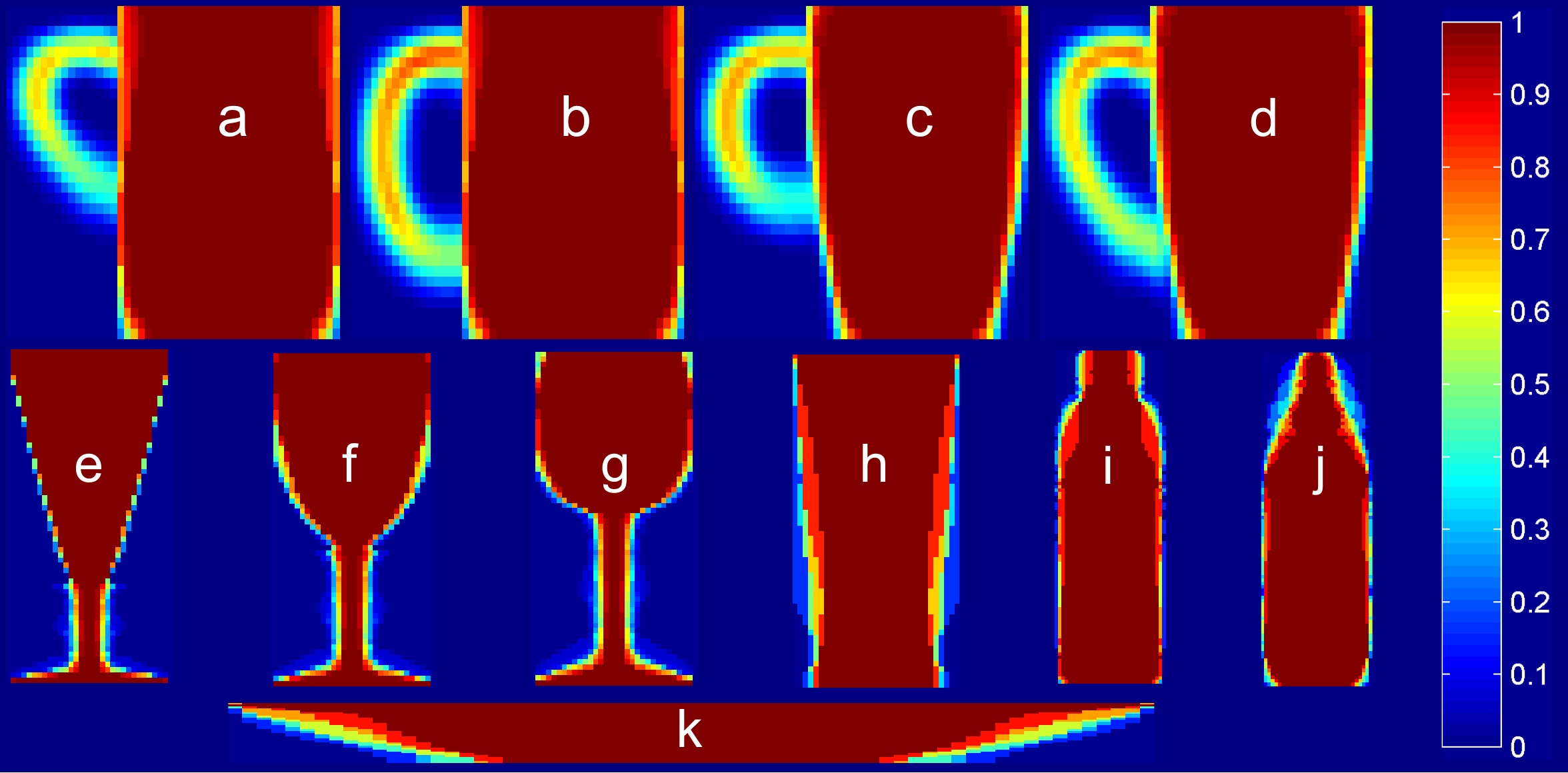}
\vspace{-5pt}
\caption{Vertical cuts through the 3D BFs corresponding to subclasses in $\mathbb{K}_{mugs}$ (a-d), $\mathbb{K}_{cups}$ (e-g), $\mathbb{K}_{glasses}$ (h), $\mathbb{K}_{bottles}$ (i-j), and $\mathbb{K}_{plates}$ (k). Colors indicate the probability that each point in the vertical plane would be inside an object of the subclass.}
\vspace{-15pt}
\label{fig:ShapePriors}
\end{figure}

\begin{figure}[t]
\vspace{-0pt}
\includegraphics[width=\columnwidth]{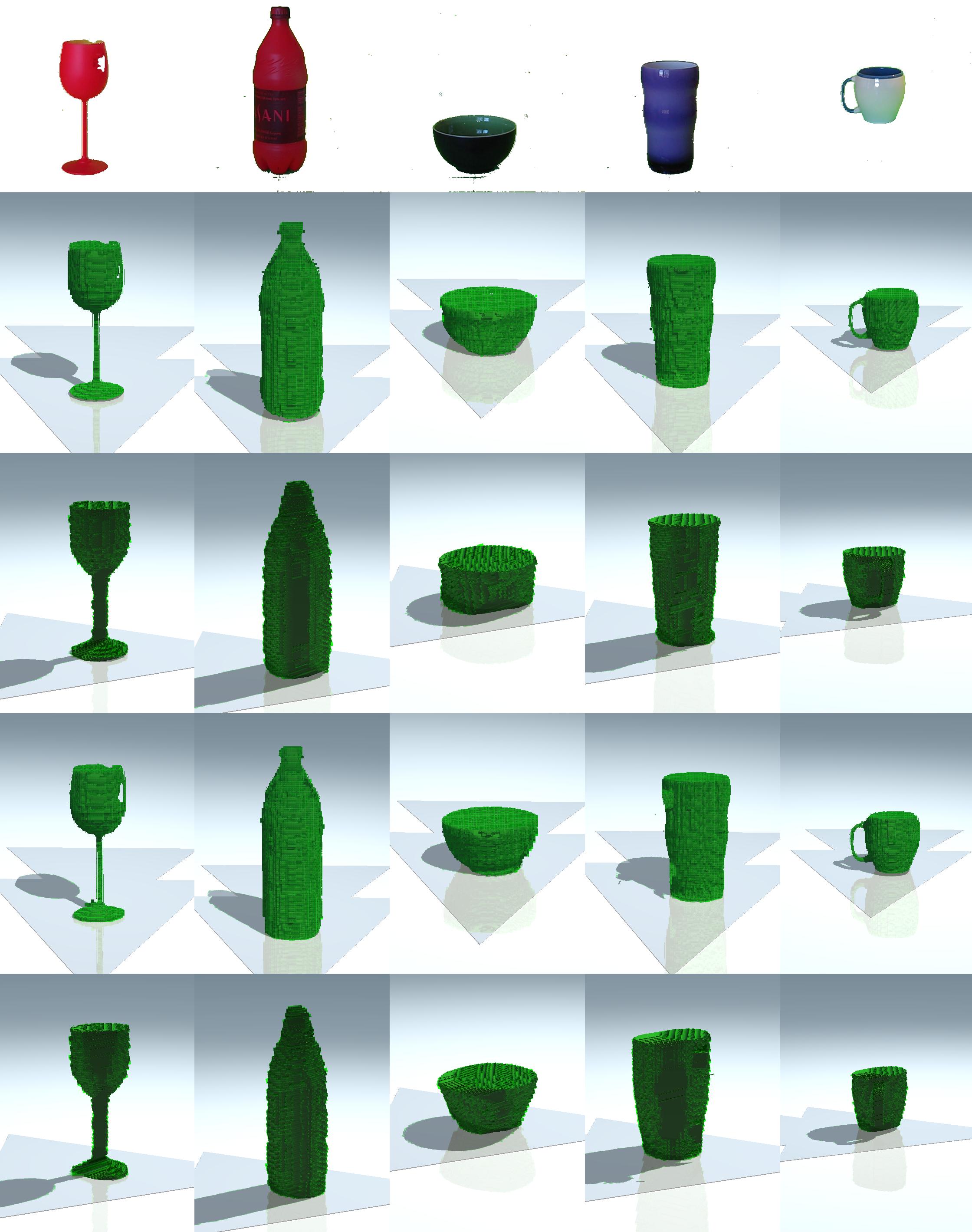}
\vspace{-10pt}
\caption{3D reconstructions $\tilde{v}$ obtained in five different experiments. For each input image (shown in the $1^{st}$ row after the background was ``subtracted"), two reconstructions were computed, and for each reconstruction, two views are shown (as in Fig. \ref{fig:ReconstructionProcess}). The reconstructions correspond to the solutions with the highest ($2^{nd}$ and $3^{rd}$ rows) and lowest ($4^{th}$ and $5^{th}$ rows) upper bound. In the second reconstruction for the glass ($4^{th}$ column, $4^{th}$ and $5^{th}$ rows) the class `glasses' was mistaken for the class `mugs.'}
\vspace{-15pt}
\label{fig:3DReconstructionsUpper}
\end{figure}

In contrast with the previous section, in this section we look at the statistical performance of the framework on a dataset containing 32 images (see examples in the first row of figures \ref{fig:3DReconstructionsUpper}, \ref{fig:3DReconstructionsLower} and \ref{fig:3DReconstructionsAdditional}).
The image BFs (\ie, $B_f$ in \eqref{eq:Evidence}) were obtained using Background subtraction, and the BFs for the shape priors (\ie, the $B_K$'s) were computed from a sample of training 3D shapes for each class. We split each class in \emph{subclasses} by clustering the shapes in the class, and then compute a BF for each subclass using all the objects in the subclass (see details in \cite{Rot10}). We denote by $\mathbb{K}_{class}$ the set of subclasses of a class. For the classes `cups,' `bottles,' `plates,' `glasses' and `mugs,' we defined 9, 3, 2, 1 and 16 subclasses, respectively. Fig. \ref{fig:ShapePriors} shows 2D cuts of the 3D BFs obtained for some of these subclasses.

To define the hypothesis spaces we define the transformation 
\begin{align}
\label{eq:DefTtxtysxysz}
T_{\vec{\Psi}} (\vec{X}) & \triangleq T_{t_x t_y} \Big(R_z(\phi) S_z(s_z) S_{xy}(s_{xy}) \vec{X}\Big),
\end{align}
which depends on the vector of parameters $\vec{\Psi} = [t_x t_y \phi$ $s_{xy} s_z]$, and combines the horizontal translation $T_{t_x t_y}$ in \eqref{eq:TranslationXY}, with a rotation of $\phi$ degrees around the vertical $z$ axis, $R_z(\phi)$, a scaling of the $z$ axis by $s_z$\%, $S_z(s_z)$, and a scaling of the $x-y$ axes by $s_{xy}$\%, $S_{xy}(s_{xy})$. Unless otherwise stated the parameters of this transformation are in the following sets: $t_x \in \{-3:0.5:3\}$, $t_y \in \{-9:1.5:9\}$, $\phi \in \{-80:20:80\}$, $s_{xy} \in \{-20:5:20\}$ and $s_z \in \{-20:5:20\}$.

We then define the hypothesis spaces $\mathbb{H}_3 \triangleq \{(K_{object},$ $T_{\vec{\Psi}}): \phi = s_{xy} = s_z = 0 \}$ and $\mathbb{H}_4 \triangleq \{(K_{object}, T_{\vec{\Psi}}): s_{xy} = s_z = 0 \}$ for the case where the object is known ($K_{object}$ is then a ``class" containing just this object), the hypothesis spaces $\mathbb{H}_5 \triangleq \{(K, T_{\vec{\Psi}}): K \in \mathbb{K}_{class}, \phi = 0 \}$ and $\mathbb{H}_6 \triangleq \{(K, T_{\vec{\Psi}}): K \in \mathbb{K}_{class} \}$ for the case where the object is not known a priori, but only its \emph{class} is, and the hypothesis space $\mathbb{H}_7 \triangleq \{(K, T_{\vec{\Psi}}): K \in \mathbb{K}_{AllClasses}, \phi \in \{-60:20:60\} \}$ for the case where neither the object nor its class are known ($\mathbb{K}_{AllClasses} = \bigcup_{class}{\mathbb{K}_{class}}$). 
Note that when the object is known (\ie, for $\mathbb{H}_3$ or $\mathbb{H}_4$) there is no need to estimate $s_{xy}$ and $s_z$ (because the object dimensions are known). Similarly $\phi$ does not need to be estimated when the object is known to belong to a rotationally symmetric class (\eg, bottles, cups, glasses or plates), only when it belongs to a non-symmetric class (\eg, mugs). For this reason the sets $\mathbb{H}_3$ or $\mathbb{H}_5$ are used in the first case and the sets $\mathbb{H}_4$ or $\mathbb{H}_6$ are used in the second.

Some comments about the choice of the parameter ranges are in order.
The ranges of $t_x$ and $t_y$ were restricted (with respect to those in Experiment 2) to save memory, since it was shown in Fig. \ref{fig:ComplexWorkPerHypothesis}b that hypotheses farther away from the ground truth are immediately discarded. Moreover, the distance between hypotheses was adjusted to be in the order that the framework can distinguish (from Table \ref{tab:NoisyLocalizationErrors}: 0.5cm and 1.5cm in the $x$ and $y$ directions, respectively).
The range of $\phi$ was restricted in $\mathbb{H}_7$ to avoid ambiguities between mugs and glasses. These ambiguities result when a mug is rotated in a way that hides its handle and hence it cannot be distinguished from a glass. These ambiguities were avoided to distinguish classification failures due to problems with our method from those intrinsic to the problem formulation.

\paragraph{\textbf{Pose estimation.}}
Table \ref{tab:DegreesOfFreedom} summarizes the pose estimation errors obtained on the hypothesis spaces defined before.
As expected the precision is in general reduced (\ie the standard deviations increased) and more solutions are found when the number of degrees of freedom is increased. Note that even in the case where the class is unknown ($\mathbb{H}_7$), the pose parameters can be estimated accurately. The largest standard deviation is observed for the parameter $\phi$ of the rotation $R_z$ around the vertical axis, because this rotation affects only a small part of the object (the mug's handle), and because this part is highly variable and hence not encoded in the BFs as well as the main body of the mugs (see Fig. \ref{fig:ShapePriors}a-d).
This problem would be solved with a larger training dataset of 3D shapes (currently containing between 6 and 36 objects per class) and more elaborated methods to construct BFs. This issue will be further discussed in \S \ref{sec:Conclusions}.

\begin{table}[htb]
\vspace{-15pt}
\caption{Pose estimation errors for hypothesis spaces with different number of degrees of freedom.}
\label{tab:DegreesOfFreedom}
\begin{tabular}{|
@{}>{\centering\arraybackslash}m{0.2in}@{}|@{}>{\centering\arraybackslash}m{0.30in}@{}|@{}>{\centering\arraybackslash}m{0.3in}@{}|@{}>{\centering\arraybackslash}m{0.25in}@{}|@{}>{\centering\arraybackslash}m{0.3in}@{}|@{}>{\centering\arraybackslash}m{0.25in}@{}|@{}>{\centering\arraybackslash}m{0.3in}@{}|@{}>{\centering\arraybackslash}m{0.30in}@{}|@{}>{\centering\arraybackslash}m{0.25in}@{}|@{}>{\centering\arraybackslash}m{0.30in}@{}|@{}>{\centering\arraybackslash}m{0.25in}@{}|@{}>{\centering\arraybackslash}m{0.250in}@{}| } \hline 
& $\mu_{t_x}$ & $\mu_{t_y}$ & $\mu_\phi$ & $\mu_{s_{xy}}$ & $\mu_{s_z}$ & $\sigma_{t_x}$ & $\sigma_{t_y}$ & $\sigma_\phi$ & $\sigma_{s_{xy}}$ & $\sigma_{s_z}$ & $|\mathbb{A}|$ \\ 
& (cm) & (cm) & (${}^\circ$) & (\%) & (\%) & (cm) & (cm) & (${}^\circ$) & (\%) & (\%) & \\ \hline 
$\mathbb{H}_3$ & -0.1 & 0.6 &   -  &   -   & -   & 0.2 & 0.9 &   -  &   -  &    - & 2.6 \\ \hline 
$\mathbb{H}_4$ & -0.1 & 0.3 &   9  &    - &  -   & 0.1 & 0.4 & 22 &   -  &    - & 3.2 \\ \hline 
$\mathbb{H}_5$ & -0.1 & 0.5 &   -  & 0.1 & 3.3 & 0.3 & 1.4 &   - & 6.5 & 7.0 & 22 \\ \hline 
$\mathbb{H}_6$ &  -0.1 & 0.5 & -2 & 0.4 & 0.2  & 0.2 & 1.5 & 41 & 7.8 & 2.9 & 422 \\ \hline 
$\mathbb{H}_7$ &  -0.1 & 0.3 & -3 & 2.4 & 3.5 & 0.3 & 1.7 & 32 & 9.2 & 7.7 & 153 \\ \hline 
\end{tabular}
\vspace{-15pt}
\end{table}

\paragraph{\textbf{Classification.}}
In the experiments corresponding to $\mathbb{H}_7$ the object classes were not known, and they were thus estimated in addition to the pose parameters. 
Since the proposed approach does not necessarily associate a single class to each testing image $f_k$ (because the corresponding set of solutions $\mathbb{A}_k$ might contain solutions of different classes), we report the performance of the framework with a slight modification of the traditional indicators. 

\begin{figure}[t]
\vspace{-0pt}
\begin{center}
\includegraphics[width=0.95\columnwidth]{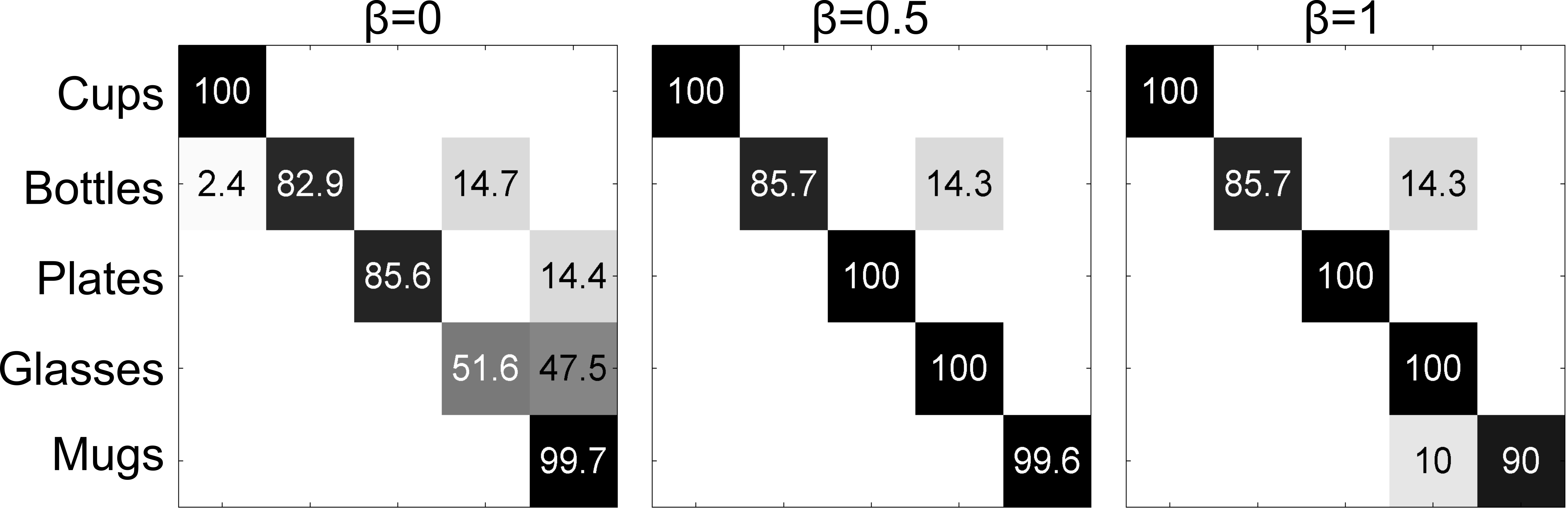}
\end{center}
\vspace{-5pt}
\caption{Confusion matrices obtained in the classification for $\beta \in \{0, 0.5, 1\}$. See text for details.}
\label{fig:ConfusionMatrices}
\vspace{-15pt}
\end{figure}

Let $class(H)$ and $class(f_k)$ be the class of the hypothesis $H$ and the true class of the object in image $f_k$, respectively, and let $E_i \triangleq \{k: class(f_k) = i\}$ be the set of indices of the images of class $i$. An element $(i,j)$ in the \emph{confusion matrix} $C_0$ (in Fig. \ref{fig:ConfusionMatrices}) indicates the total normalized percentage of solutions of class $j$ obtained for all testing images of class $i$, 
\begin{align}
C_0(i,j) \triangleq \frac{100}{|E_i|} \sum_{k \in E_i}{\frac{|\{H \in \mathbb{A}_k: class(H) = j\}|}{|\mathbb{A}_k|}}.
\end{align}
Note that if only one solution is found per experiment, this formula reduces to the standard confusion matrix.

It is also of interest to know what the classification performance is when only the best solutions are considered. For this purpose we define the confusion matrix $C_\beta$ ($0 \le \beta \le 1$) as before, but considering only the solutions whose upper bound is greater or equal than $\gamma_\beta$, where $\gamma_\beta \triangleq \underline{L} + \beta (\overline{L} - \underline{L})$ and $\underline{L}$ and $\overline{L}$ are the maximum lower and upper bounds, respectively, of any solution.
Note that when $\beta = 0$ all the solutions are considered, and when $\beta = 1$ only the solution with the largest upper bound is considered. The confusion matrices $C_{0.5}$ and $C_1$ are also shown in Fig. \ref{fig:ConfusionMatrices}.

\paragraph{\textbf{3D reconstruction.}}
As explained in \S \ref{sec:LowerBound} and \S \ref{sec:UpperBound} two 3D reconstructions are obtained for each solution (selected in this case from the set $\mathbb{H}_7$). These reconstructions are given by the discrete shape $\hat{v}$ and the semidiscrete shape $\tilde{v}$ obtained, respectively, while computing the lower and upper bound for the solutions.

In order to quantify the quality of these reconstructions we computed the error in the reconstruction $\hat{v}$ obtained for the best solution ($\tilde{v}$ is almost identical to $\hat{v}$), by measuring its distance to the corresponding true shape $v_{true}$. The distance $d(\hat{v},v_{true})$ is defined as the normalized measure of the set where the two shapes differ, \ie,
\begin{align}
d(\hat{v},v_{true}) \triangleq d(v,v_{true}) \triangleq \frac{|v \cup v_{true} \setminus v \cap v_{true}|}{|v_{true}|},
\end{align}
where $v$ is the continuous shape produced from $\hat{v}$ after this shape is translated to be optimally aligned with $v_{true}$. This alignment is performed to disregard the errors in the reconstruction resulting from errors in the pose, since those errors were already reported in Table \ref{tab:DegreesOfFreedom}.
Using this metric we obtained a mean reconstruction error of 16.7 \%.

Fig. \ref{fig:3DReconstructionsUpper} shows the reconstructions $\tilde{v}$ obtained for the best and worst solutions (as indicated by their upper bound) in five different experiments, one for each class considered.
Note that in most cases the best and worst reconstructions are very similar. It can be seen that the 3D reconstructions look better from viewpoints that are closer to the original viewpoint (in the input image), than from viewpoints that are ``orthogonal" to it (compare the $2^{nd}$ and $4^{th}$ rows, with the $3^{rd}$ and $5^{th}$ rows, respectively).
The explanation for this is that from viewpoints close to the original viewpoint we see the \emph{best} parts of the reconstruction, \ie, those in which information from the shape prior \emph{and} from the input image was used. In contrast, from viewpoints that are orthogonal to the original viewpoint, we see the \emph{worst} parts of the reconstruction, where only the information from the shape prior could be used.

The reconstructions $\hat{v}$ are in general almost identical to the corresponding reconstructions $\tilde{v}$. Fig. \ref{fig:3DReconstructionsLower} (in the supplementary material) shows the reconstructions $\hat{v}$ corresponding to the reconstructions $\tilde{v}$ depicted in Fig. \ref{fig:3DReconstructionsUpper}. Additional reconstructions are also shown in Fig. \ref{fig:3DReconstructionsAdditional} (in the supplementary material).

In rays containing only points deemed likely to be \texttt{Out} of the reconstruction (according to the shape prior) but that nevertheless project to pixels that are likely to be \texttt{Foreground} (according to the input image), the framework is forced to make a compromise between the contradictory information in the input image and the shape prior and to add a small amount of mass in the inner side of $\Phi$. While this is perfectly correct from the optimization perspective, the added lumps of mass constitute artifacts in the reconstruction.
These artifacts, however, are very easy to detect and remove and thus this was \emph{automatically} done in all the reconstructions shown in this work.

\begin{figure}[hb]
\vspace{-10pt}
\begin{center}
\includegraphics[width=0.9\columnwidth]{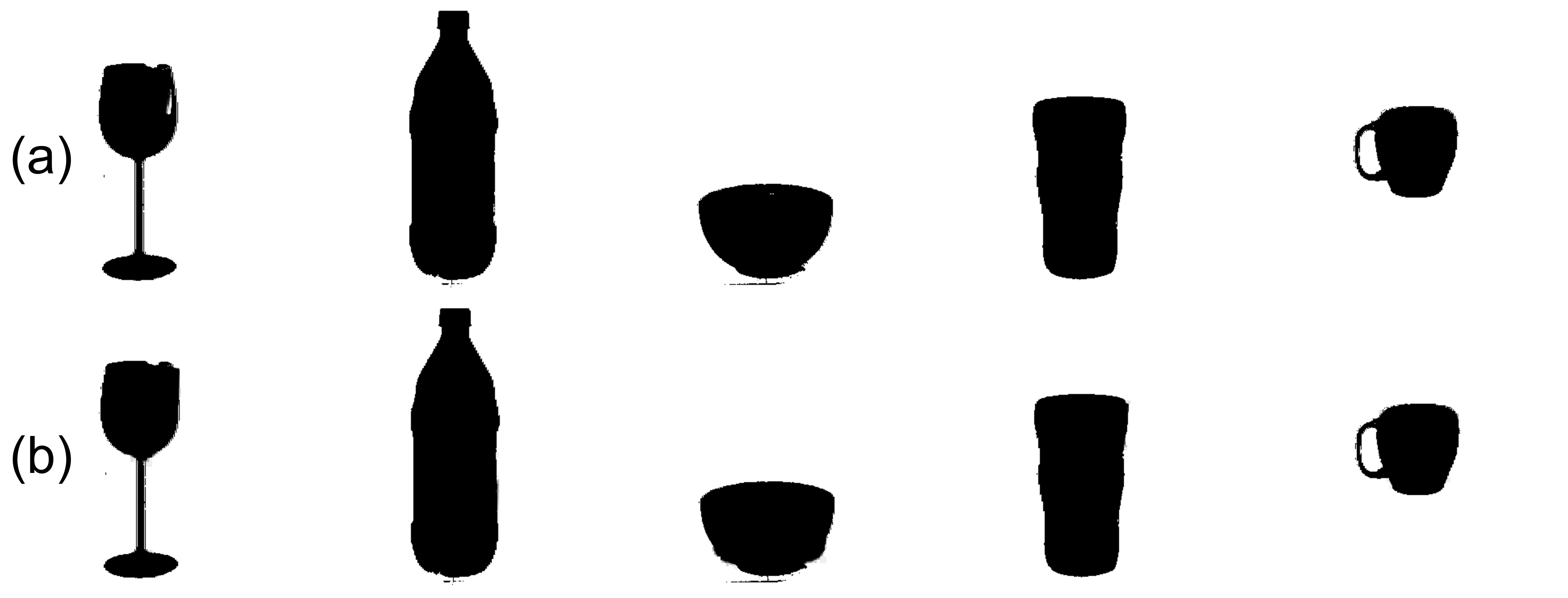}
\end{center}
\vspace{-10pt}
\caption{Segmentations $\tilde{q}$ corresponding to the best solutions depicted in Fig. \ref{fig:3DReconstructionsUpper}, obtained with $\lambda = \lambda_{opt}$ as in Fig. \ref{fig:3DReconstructionsUpper} (a) and with $\lambda = 2 \lambda_{opt}$ (b).}
\vspace{-15pt}
\label{fig:SegmentationsU1}
\end{figure}

\paragraph{\textbf{2D segmentation.}}
Recall that a pair of segmentations is also obtained for each solution, along with a pair of bounds and reconstructions. These segmentations are given by the discrete shape $\hat{q}$ and the semidiscrete shape $\tilde{q}$ obtained while computing the lower and upper bounds for the solutions, respectively. The segmentations $\tilde{q}$ corresponding to the reconstructions $\tilde{v}$ depicted in the $2^{nd}$ and $3^{rd}$ rows of Fig. \ref{fig:3DReconstructionsUpper} are shown in Fig. \ref{fig:SegmentationsU1}a. These segmentations were obtained for a value of $\lambda$ in \eqref{eq:Evidence}, namely $\lambda_{opt}$, which was chosen to make the weights of the first and third terms of that expression equal, and which was found to minimize the pose estimation error. This value $\lambda_{opt}$ thus depends on the BFs $B_f$ and $B_K$ corresponding to the input image and the class priors, and it might be different in different experiments.

If on the other hand one is interested in ``fixing" artifacts produced by the back\-ground subtraction process by considering prior 3D shape information, then we need to increase the weight given to the  shape prior term (\ie, $\lambda$). Segmentations obtained with $\lambda = 2 \lambda_{opt}$ are shown in Fig. \ref{fig:SegmentationsU1}b. Note how this larger $\lambda$ fixes the cup's ``hole" in the leftmost column.
The segmentations $\hat{q}$ are in general almost identical to the corresponding segmentations $\tilde{q}$ (see Fig. \ref{fig:SegmentationsL1} in the supplementary material). Additional segmentations are shown in Fig. \ref{fig:SegmentationsAdditional} (in the supplementary material).

\paragraph{\textbf{Comparison with an alternative approach.}}
As \linebreak[4] mentioned in \S \ref{sec:PriorWork} we consider the work of Sandhu \etal \cite{San09} to be the closest to ours, even though in that work the problems of classification and 3D reconstruction are not addressed (but could be addressed with some major modifications to the framework). One of the main differences between our proposed approach and the approach in \cite{San09} is that the latter requires a good estimation of the pose to initialize the optimization, while our approach does not. 

\begin{figure} \sidecaption
\vspace{-10pt}
\includegraphics[width=0.32\columnwidth]{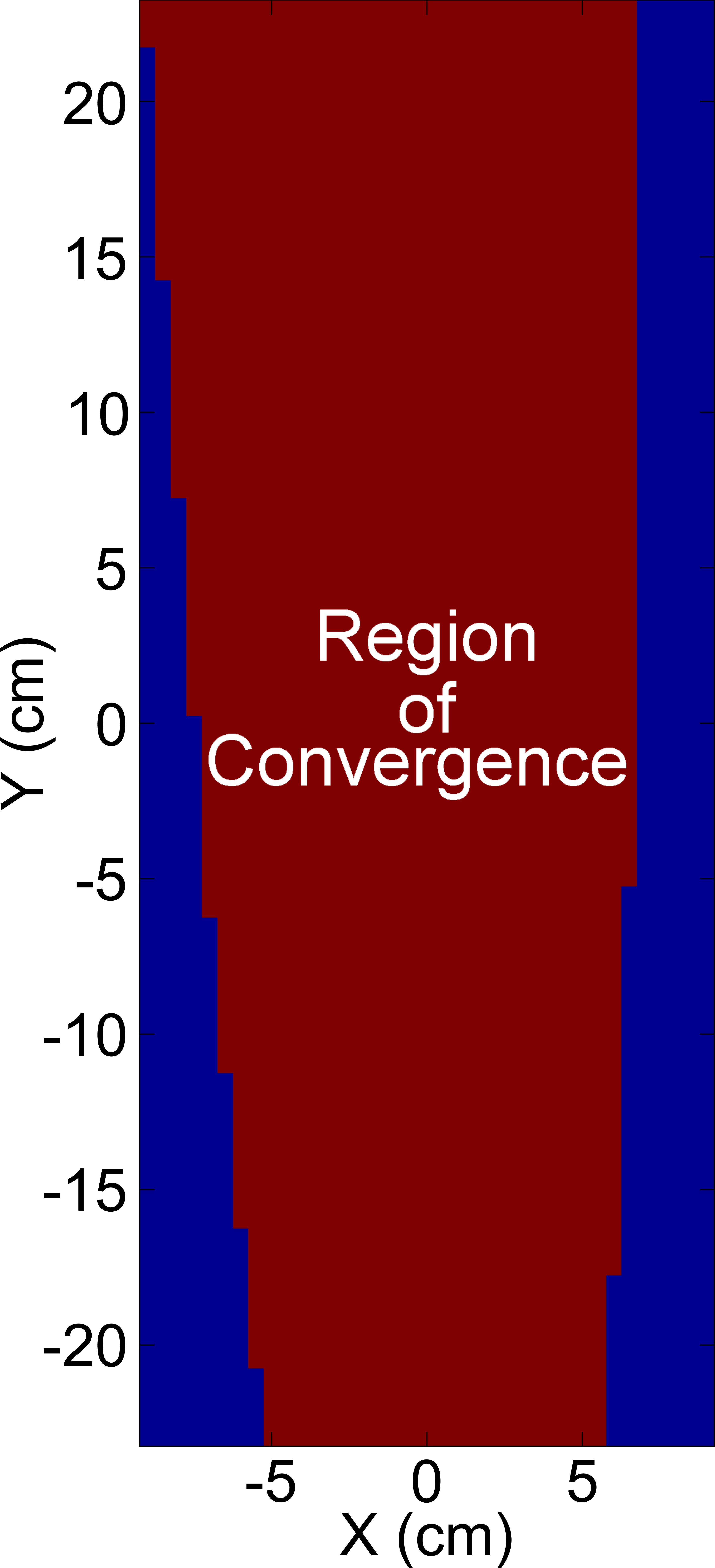}
\vspace{-10pt}
\caption{Convergence of the approach in \cite{San09} to the correct solution. When the approach in \cite{San09} is initialized with an hypothesis inside the area in red, this approach finds the correct solution. Otherwise it finds a different solution or does not converge at all.}
\vspace{-0pt}
\label{fig:ErrorsSandhu}
\end{figure}

In order to illustrate this point we show in Fig. \ref{fig:ErrorsSandhu} the region of initial poses that lead to the correct solution being found by the the approach in \cite{San09}. In other words, we repeated Experiment 2 (in \S \ref{sec:IndividualExperiments}) using the approach in \cite{San09} and using each hypothesis in $\mathbb{H}_2$ as an initial condition.
We observed that only when the initial pose is close to the true pose this approach converges to the right solution (otherwise it does not converge or converges to a different solution). While this could be solved by running that framework with different initial conditions (if the true solution is approximately known), the fact that each initial condition has to be ``fully" processed significantly increases the cost of the approach. In contrast in our approach only the hypotheses close to the true hypothesis have to be fully processed. This experiment is described in detail in \S \ref{sec:ComparisonSandhu} in the supplementary material.

This concludes the presentation of the experiments. In the next section we present our conclusions and possible directions for future work. 

%% file: Conclusions.tex
\section{Conclusions}
\label{sec:Conclusions}

This article introduced an inference framework to simultaneously tackle the problems of 3D reconstruction, pose estimation and object classification from a single input image, by considering shape cues only and by relying on prior 3D knowledge about the shape of different object classes. The proposed inference framework is based on an H\&B algorithm, which greatly reduces the amount of computation required while still being guaranteed to find the optimal solutions. In order to instantiate the H\&B paradigm for the current problem, we extended the theory of shapes and shape priors presented in \cite{Rot10} to handle projections of shapes. 

While the proposed approach already provides state-of-the-art results, it still can be improved and extended in several directions. For example, it could be extended to exploit the redundancy among hypotheses, by grouping them, computing bounds for these groups, and then discarding whole groups of hypotheses together (\emph{a la} Branch and Bound). Other directions include considering different types of input images (\eg, depth maps) or multiple cameras or videos. 
 \textcolor{white}{\cite{Kah95,Vio01}}

%% file: ComputingSummaries.tex
\section{Supplementary material: Computing summaries in constant time}
\label{sec:ComputingSummaries}

One of the main properties of summaries is that, for certain kinds of sets, they can be computed in constant time regardless of the number of elements (\eg, pixels) in these sets. Next we show how to compute the \emph{mean}-summary $\hat{Y}_{B,\Phi}$ and the \emph{m}-summaries $\tilde{Y}_{B,\Phi}$ of a BF $B$ for the set $\Phi \subset \Omega$ (defined below).  For simplicity we assume that $\Omega \subset \mathbb{R}^2$, but the results presented here immediately generalize to higher dimensions. We also assume that $\Pi(\Omega)=\{\Omega_{1,1}, \Omega_{1,2}, \dots, …,\Omega_{n,n}\}$ is a uniform partition of $\Omega$ organized in rows and columns, where each partition element $\Omega_{i,j}$ (in the $i$-th row and the $j$-th column) is a square of area $|\Omega_{i,j}|=u_0$. We assume that $B$, defined by its logit function $\delta_B(\vec{x})$ ($\vec{x} \in \Omega)$, was produced from a discrete BF $\hat{B}$ in $\Pi(\Omega)$ (as described in Definition \ref{def:BernoulliField}), and therefore $\delta_B(\vec{x}) \triangleq \delta_{\hat{B}}(i,j) \ \forall \vec{x} \in \Omega_{i,j}$.

\paragraph{Computing \emph{mean}-summaries in a box.}
Let us assume for the time being that $\Phi$ is an axis-aligned rectangular region containing only whole pixels (\ie, not parts of pixels). That is, 
\begin{equation}
\label{eq:Box}
\Phi \triangleq \bigcup_{
\begin{array}{c}
\scriptstyle{i_L \le i \le i_U} \\
\scriptstyle{j_L \le j \le j_U}
\end{array}}{\Omega_{i,j}}.
\end{equation}
These special regions will be referred to as \emph{boxes}.
In order to compute the \emph{mean}-summary $\hat{Y}_{B,\Phi}$ in the box $\Phi$, note that from \eqref{eq:DefinitionMeanSummary},
\begin{align}
\hat{Y}_{B,\Phi} & = 
\sum_{\begin{array}{c}
\scriptstyle{i_L \le i \le i_U} \\
\scriptstyle{j_L \le j \le j_U}
\end{array}}{\int_{\Omega_{i,j}}{\delta_B(\vec{x})\ d\vec{x}}} = \nonumber \\
\label{eq:ToComputeMeanSummaries}
& = u_o \sum_{\begin{array}{c}
\scriptstyle{i_L \le i \le i_U} \\
\scriptstyle{j_L \le j \le j_U}
\end{array}}{\delta_{\hat{B}}(i,j)}.
\end{align}
The sum on the rhs of \eqref{eq:ToComputeMeanSummaries} can be computed \emph{in constant time} by relying on \emph{integral images} \cite{Vio01}, an image representation precisely proposed to compute sums in rectangular domains in constant time. To accomplish this, integral images precompute a matrix where each pixel stores the cumulative sum of the values in pixels with lower indices. The sum in \eqref{eq:ToComputeMeanSummaries} is then computed as the sum of four of these precomputed cumulative sums.

\paragraph{Computing \emph{m}-summaries in a box.}
The formula to compute the \emph{m}-summary $\tilde{Y}_{B,\Phi}$ in the box $\Phi$ is similarly derived. From \eqref{eq:DefMSummary}, and since $\delta_B$ is constant inside each partition element, it holds for $k=-m, \dots,m$ that
\begin{align}
\tilde{Y}_{B,\Phi}^k & = \left | \left \{ \vec{x} \in \Phi : \delta_B(\vec{x}) < \frac{k\delta_{max}}{m} \right \} \right | =  u_o \left | \left \{ (i,j): \vphantom{\frac{j\delta_{max}}{m}} \right. \right. \nonumber \\
\label{eq:PreToComputeMSummaries}
&  \left. \left.  i_L \le i \le i_U, j_L \le j \le j_U, \delta_{\hat{B}}(i,j) < \frac{k\delta_{max}}{m} \right \} \right |.
\end{align}
Let us now define the matrices $I_k$ ($k=-m, \dots, m$) as
\begin{equation}
I_k(i,j) \triangleq \left \{
\begin{array}{ll}
1, & \ \text{if} \ \delta_{\hat{B}}(i,j) < k \delta_{max} / m \\
0, & \ \text{otherwise}.
\end{array}
\right.
\end{equation}
Using this definition, \eqref{eq:PreToComputeMSummaries} can be rewritten as
\begin{equation}
\tilde{Y}_{B,\Phi}^k = u_o \sum_{
\begin{array}{c}
\scriptstyle{i_L \le i \le i_U} \\
\scriptstyle{j_L \le j \le j_U}
\end{array}} {I_k(i,j)},
\label{eq:ToComputeMSummaries}
\end{equation}
which as before can be computed in $O(1)$ using integral images.

\paragraph{Computing \emph{mean}-summaries in a convex set.}
In general we are interested in cases in which $\Phi$ is not axis-aligned or even rectangular; we only require $\Phi$ to be convex. In this case we will not compute $\hat{Y}_{B,\Phi}$ exactly, but rather we will find a lower bound for it. Note that by doing this we can still obtain valid lower bounds for the evidence using \eqref{eq:LowerBoundLocal}. 

\begin{figure}[h]
\includegraphics[width=\columnwidth]{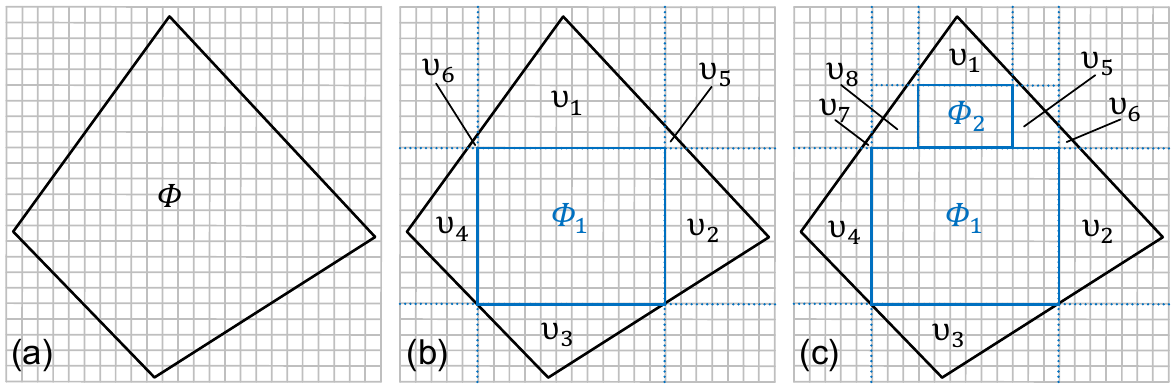}
\caption{Three partitions of the set $\Phi$ used to compute $\hat{Y}_{B,\Phi}$ and $\tilde{Y}_{B,\Phi}^k$. (a) The original set. (b and c) Partitions after one and two iterations, respectively.}
\label{fig:ComputingSummaries}
\end{figure}

Toward this end we partition $\Phi$ as $\{\Phi_1, \dots, \Phi_{n_\Phi},$ $\upsilon_1, \dots , \upsilon_{n_\upsilon} \}$, where each $\Phi_i$ is a box (as defined in \eqref{eq:Box}) and each $\upsilon_i$ is a set whose bounding box $\Gamma(\upsilon_i)$ is disjoint with the other bounding boxes and the $\Phi_i$'s (see Fig. \ref{fig:ComputingSummaries}). Specifically, $\Gamma(\upsilon_i)$ is defined as the smallest box containing $\upsilon_i$. To obtain this partition we find the largest box inside $\Phi$ and we label it as $\Phi_1$. Then we ``cut'' $\Phi$ with the lines determined by the sides of $\Phi_1$, yielding $\upsilon_1, \dots, \upsilon_8$ (see Fig. \ref{fig:ComputingSummaries}b). Next, the largest $\upsilon_i$, say $\upsilon_1$, is selected and the largest box inside it is found and labeled $\Phi_2$. And again, $\upsilon_1$ is cut with the lines determined by the sides of $\Phi_2$ (see Fig. \ref{fig:ComputingSummaries}c). This process is repeated a number of times, relabeling the $\upsilon_i$'s at each step, until the desired summary precision is met.

For each partition, obtained at each step, it holds that 
\begin{align}
\hat{Y}_{B,\Phi} = \sum_{i=1}^{n_\Phi}{\hat{Y}_{B,\Phi_i}} + \sum_{i=1}^{n_\upsilon}{\hat{Y}_{B,\upsilon_i}},
\label{eq:PartsMeanSummary}
\end{align}
where the $\hat{Y}_{B,\Phi_i}$'s are computed exactly using \eqref{eq:ToComputeMeanSummaries}. To bound the $\hat{Y}_{B,\upsilon_i}$'s, we observe that
\begin{align}
\hat{Y}_{B,\upsilon_i} = \int_{\upsilon_i}{\delta_B(\vec{x})\ d\vec{x}} \ge -\delta_{max} |\upsilon_i|,
\label{eq:PreBoundMeanSummary1}
\end{align}
and that
\begin{align}
\hat{Y}_{B,\Gamma(\upsilon_i)} = \int_{\upsilon_i}{\delta_B(\vec{x})\ d\vec{x}} + \int_{\Gamma(\upsilon_i) - \upsilon_i} {\delta_B(\vec{x})\ d\vec{x}},
\end{align}
and hence
\begin{align}
\hat{Y}_{B,\upsilon_i} \ge \hat{Y}_{B,\Gamma(\upsilon_i)} - \delta_{max} \left(\left|\Gamma (\upsilon_i)\right|- |\upsilon_i|\right).
\label{eq:PreBoundMeanSummary2}
\end{align}
Note that the summary in the first term on the rhs of \eqref{eq:PreBoundMeanSummary2} can be computed exactly using \eqref{eq:ToComputeMeanSummaries} because $\Gamma(\upsilon_i)$ is a box.

Substituting \eqref{eq:PreBoundMeanSummary1} and \eqref{eq:PreBoundMeanSummary2} into \eqref{eq:PartsMeanSummary} yields the final lower bound for the mean-summary,
\begin{align}
\hat{Y}_{B,\Phi} \ge \sum_{i=1}^{n_\Phi} {\hat{Y}_{B,\Phi_i}} + \nonumber \\
\sum_{i=1}^{n_\upsilon} {\max  \left\{ \hat{Y}_{B,\Gamma(\upsilon_i)} - \delta_{max} \left(\left|\Gamma (\upsilon_i)\right| - |\upsilon_i| \right), -\delta_{max} |\upsilon_i| \right\} }.
\end{align}
   
Clearly better bounds for $\hat{Y}_{B,\Phi}$ are obtained in finer partitions of $\Phi$ (greater $n_\Phi$) at a greater computational cost. We found that for our purposes $n_\Phi=10$ in 2D, and $n_\Phi=30$ in 3D, provide a good compromise.

\paragraph{Computing \emph{m}-summaries in a convex set.}
For a convex set $\Phi$ we are going to compute an upper bound $\overline{\tilde{Y}^k_{B,\Phi}}$ for $\tilde{Y}^k_{B,\Phi}$, by partitioning $\Phi$ into $\left\{\Phi_1, \dots, \Phi_{n_\Phi}, \upsilon_1, \dots, \upsilon_{n_\upsilon} \right\}$ as before. Using this bound we will in turn obtain a valid upper bound for the evidence.
   
Given a partition of $\Phi$ as described above, it follows from \eqref{eq:DefMSummary} that
\begin{align}
\tilde{Y}^k_{B,\Phi} = \sum_{i=1}^{n_\Phi}{\tilde{Y}^k_{B,\Phi_i}} + \sum_{i=1}^{n_\upsilon}{\tilde{Y}^k_{B,\upsilon_i}}.
\label{eq:PartsMSummary}
\end{align}
The \emph{m}-summaries $\tilde{Y}_{B,\Phi_i}$ in \eqref{eq:PartsMSummary} can be computed exactly using \eqref{eq:ToComputeMSummaries}. The \emph{m}-summaries $\tilde{Y}_{B,\upsilon_i}$, on the other hand, can only be bounded. Below we derive an upper bound $\overline{\tilde{Y}^k_{B,\upsilon_i}}$ for them. Substituting this bound in \eqref{eq:PartsMSummary} we will obtain the upper bound $\overline{\tilde{Y}^k_{B,\Phi}}$ for $\tilde{Y}^k_{B,\Phi}$.

Recall that our goal is to substitute these summaries in \eqref{eq:F}-\eqref{eq:J} to find an upper bound for the lhs of \eqref{eq:LemMSummaryThesis}. Since we do not know which of the values of $\delta_B$ inside $\Gamma(\upsilon_i)$ are actually inside $\upsilon_i$, we need to consider the worst case. This worst case is when the greatest values of $\delta_B$ inside $\Gamma(\upsilon_i)$ are actually inside $\upsilon_i$. In other words, we need to ``fill" $\upsilon_i$ with the greatest values of $\delta_B$ in $\Gamma(\upsilon_i)$.

In order to simplify the derivation of the bound $\overline{\tilde{Y}^k_{B,\Phi}}$, we define the quantities
\begin{align}
M^k_{\Gamma(\upsilon_i)} \triangleq \left\{
\begin{array}{cl}
|\Gamma(\upsilon_i)| - \tilde{Y}^m_{B,\Gamma(\upsilon_i)}, & \quad \text{if} \ k = m, \\
\tilde{Y}^{k+1}_{B,\Gamma(\upsilon_i)} - \tilde{Y}^k_{B,\Gamma(\upsilon_i)}, & \quad \text{if} \ k < m.
\end{array}
\right.
\end{align}
Each of these quantities, \eg $M^k_{\Gamma(\upsilon_i)}$, indicate the measure of a set of the kind $\{\vec{x} \in \Gamma(\upsilon_i): k \delta_{max} / m \le \delta_B(\vec{x}) < (k+1) \delta_{max} / m\}$.
Similarly we define the corresponding quantities for the upper bound of the summary, $\overline{\tilde{Y}^k_{B,\upsilon_i}}$, that we want to compute,
\begin{align}
M^k_{\upsilon_i} \triangleq \left\{
\begin{array}{cl}
|\upsilon_i| - \overline{\tilde{Y}^m_{B,\upsilon_i}}, & \quad \text{if} \ k = m, \\
\overline{\tilde{Y}^{k+1}_{B,\upsilon_i}} - \overline{\tilde{Y}^k_{B,\upsilon_i}}, & \quad \text{if} \ k < m.
\end{array}
\right.
\end{align}

Note that since $\upsilon_i \subset \Gamma(\upsilon_i)$, the quantity $M^k_{\upsilon_i}$ is bounded above by the quantity $M^k_{\Gamma(\upsilon_i)}$. Moreover, this quantity $M^k_{\upsilon_i}$ is also bounded above by the remaining volume $V^k_{\upsilon_i}$ in $\upsilon_i$ (\ie, the volume not yet ``filled"), which can be written as
\begin{align}
V^k_{\upsilon_i} = |\upsilon_i| - \sum_{j = k+1}^{m}{M^j_{\upsilon_i}}.
\label{eq:RemainingVolume}
\end{align}
Therefore we can compute $M^k_{\upsilon_i}$ from $M^k_{\Gamma(\upsilon_i)}$ as
\begin{align}
M^k_{\upsilon_i} = \min \left\{ M^k_{\Gamma(\upsilon_i)}, V^k_{\upsilon_i} \right\}.
\label{eq:RecursionM}
\end{align}

Since it can be verified that $V^k_{\upsilon_i}$ satisfies
\begin{align}
V^k_{\upsilon_i} = \left\{
\begin{array}{cl}
|\upsilon_i|, & \quad \text{if} \ k = m, \\
\overline{\tilde{Y}^{k+1}_{B,\upsilon_i}}, & \quad \text{if} \ k < m, \\
\end{array}
\right.
\end{align}
it follows from \eqref{eq:RecursionM} that the bound for the summary $\overline{\tilde{Y}_{B,\upsilon_i}}$ can be computed with the following recursion (in decreasing order of $k$):
\begin{align}
\overline{\tilde{Y}^m_{B,\upsilon_i}} & = |\upsilon_i| - \min \left \{|\Gamma(\upsilon_i)| - \tilde{Y}^m_{B,\Gamma(\upsilon_i)}, |\upsilon_i|\right \} \\
\overline{\tilde{Y}^k_{B,\upsilon_i}} & = \overline{\tilde{Y}^{k+1}_{B,\upsilon_i}} - \min \left \{\tilde{Y}^{k+1}_{B,\Gamma(\upsilon_i)} - \tilde{Y}^k_{B,\Gamma(\upsilon_i)}, \overline{\tilde{Y}^{k+1}_{B,\upsilon_i}} \right \} \nonumber \\
& \qquad \qquad \qquad \qquad \qquad (-m \le k < m).
\end{align}

Thus the final upper bound for the \emph{m}-summary is given by
\begin{align}
\overline{\tilde{Y}^k_{B,\Phi}} = \sum_{i=1}^{n_\Phi}{\tilde{Y}^k_{B,\Phi_i}} + \sum_{i=1}^{n_\upsilon}{\overline{\tilde{Y}^k_{B,\upsilon_i}}}.
\end{align}

%% file: ResultsExtra.tex
\section{Supplementary material: experimental results}
\label{sec:ExtraResults}

This section contains additional results that due to the space limitations could not be included in the main text.

\begin{figure}[htb]
\vspace{-15pt}
\includegraphics[width=\columnwidth]{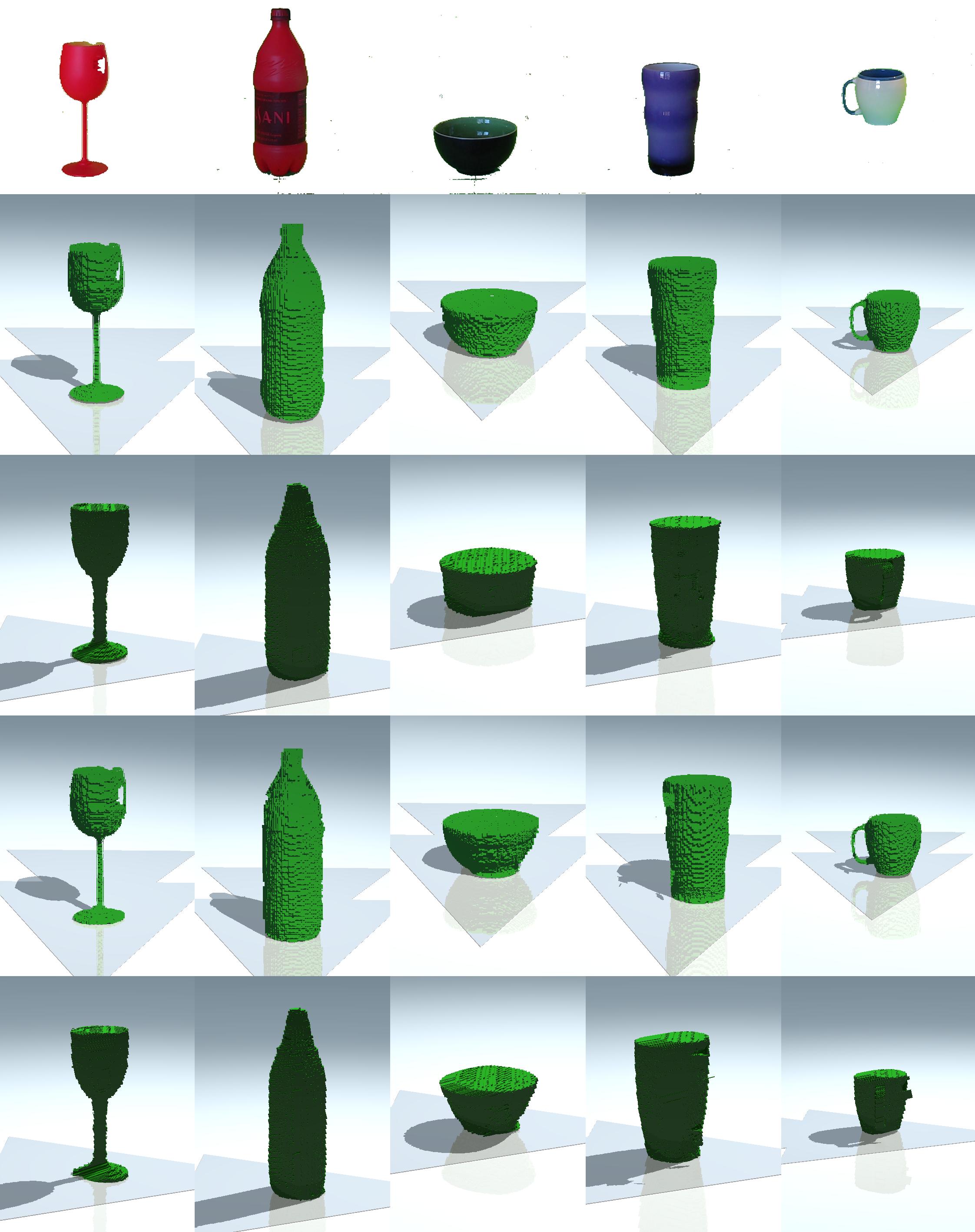}
\vspace{-10pt}
\caption{3D reconstructions $\hat{v}$ obtained while computing the lower bound. The input image used in each experiment is shown in the $1^{st}$ row after the background was ``subtracted". The colors in these images were left for clarity only, our framework does not consider these colors, only the foreground probability at each pixel. For each input image two reconstructions were computed, and for each reconstruction, two views are shown (as in Fig. \ref{fig:ReconstructionProcess}). The reconstructions correspond to the solutions with the highest ($2^{nd}$ and $3^{rd}$ rows) and lowest ($4^{th}$ and $5^{th}$ rows) upper bound. In the second reconstruction for the glass ($4^{th}$ column, $4^{th}$ and $5^{th}$ rows) the class `glasses' was mistaken for the class `mugs.'}
\vspace{-15pt}
\label{fig:3DReconstructionsLower}
\end{figure}

\begin{figure*}[htb]
\vspace{-0pt}
\includegraphics[width=\textwidth]{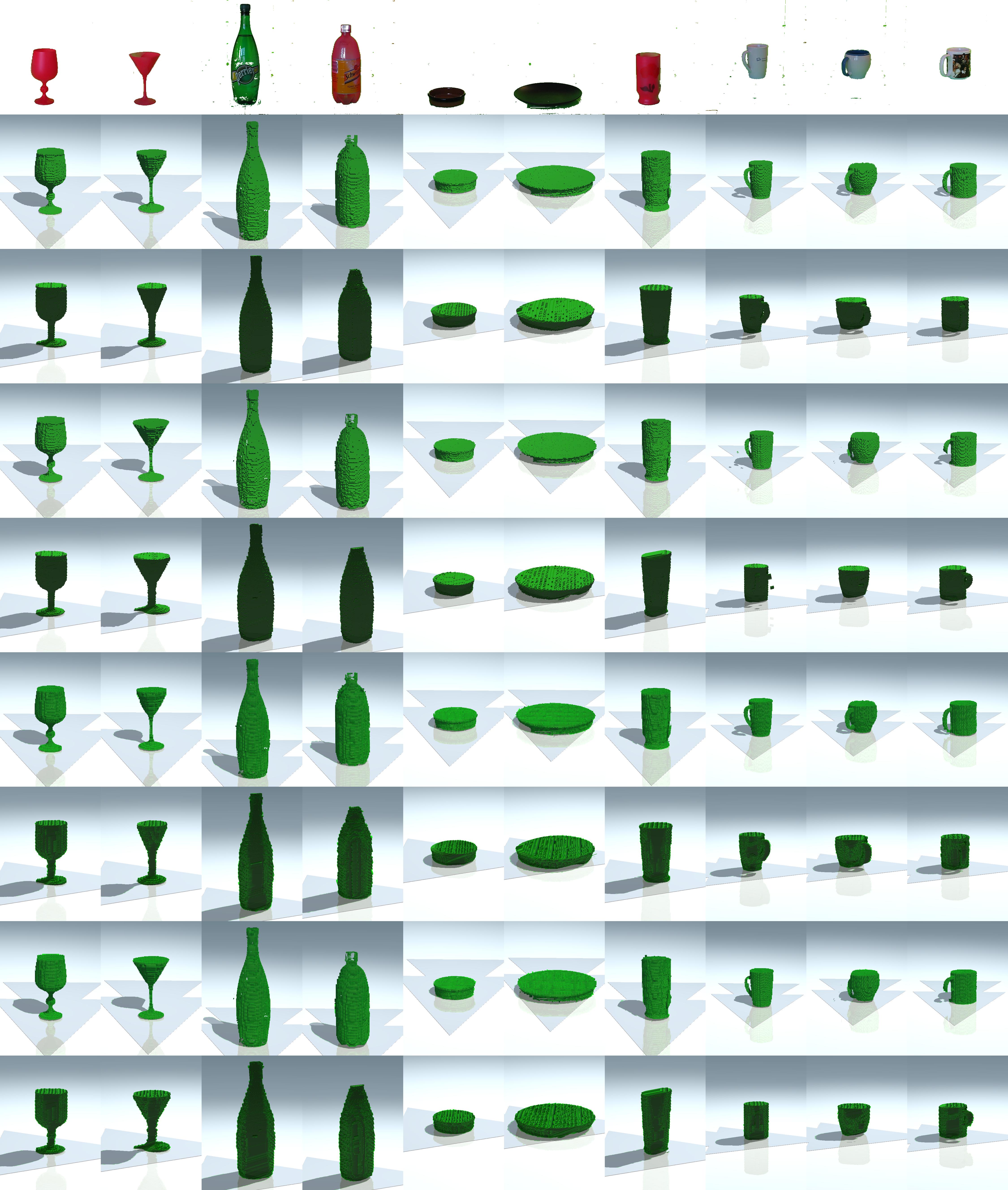}
\vspace{-10pt}
\caption{Additional examples of 3D reconstructions. ($1^{st}$ row) Input image used in each experiment after the background was ``subtracted". The colors were left only for clarity. Our framework does not consider these colors, only the foreground probability at each pixel. ($2^{nd}$ and $3^{rd}$ rows) Two orthogonal views of the \emph{lower} reconstruction $\hat{v}$ obtained for the \emph{best} solution. ($4^{th}$ and $5^{th}$ rows) Two orthogonal views of the \emph{lower} reconstruction $\hat{v}$ obtained for the \emph{worst} solution. ($6^{th}$ and $7^{th}$ rows) Two orthogonal views of the \emph{upper} reconstruction $\tilde{v}$ obtained for the \emph{best} solution. ($8^{th}$ and $9^{th}$ rows) Two orthogonal views of the \emph{upper} reconstruction $\tilde{v}$ obtained for the \emph{worst} solution.}
\vspace{-15pt}
\label{fig:3DReconstructionsAdditional}
\end{figure*}

\begin{figure}
\vspace{-10pt}
\begin{center}
\includegraphics[width=0.9\columnwidth]{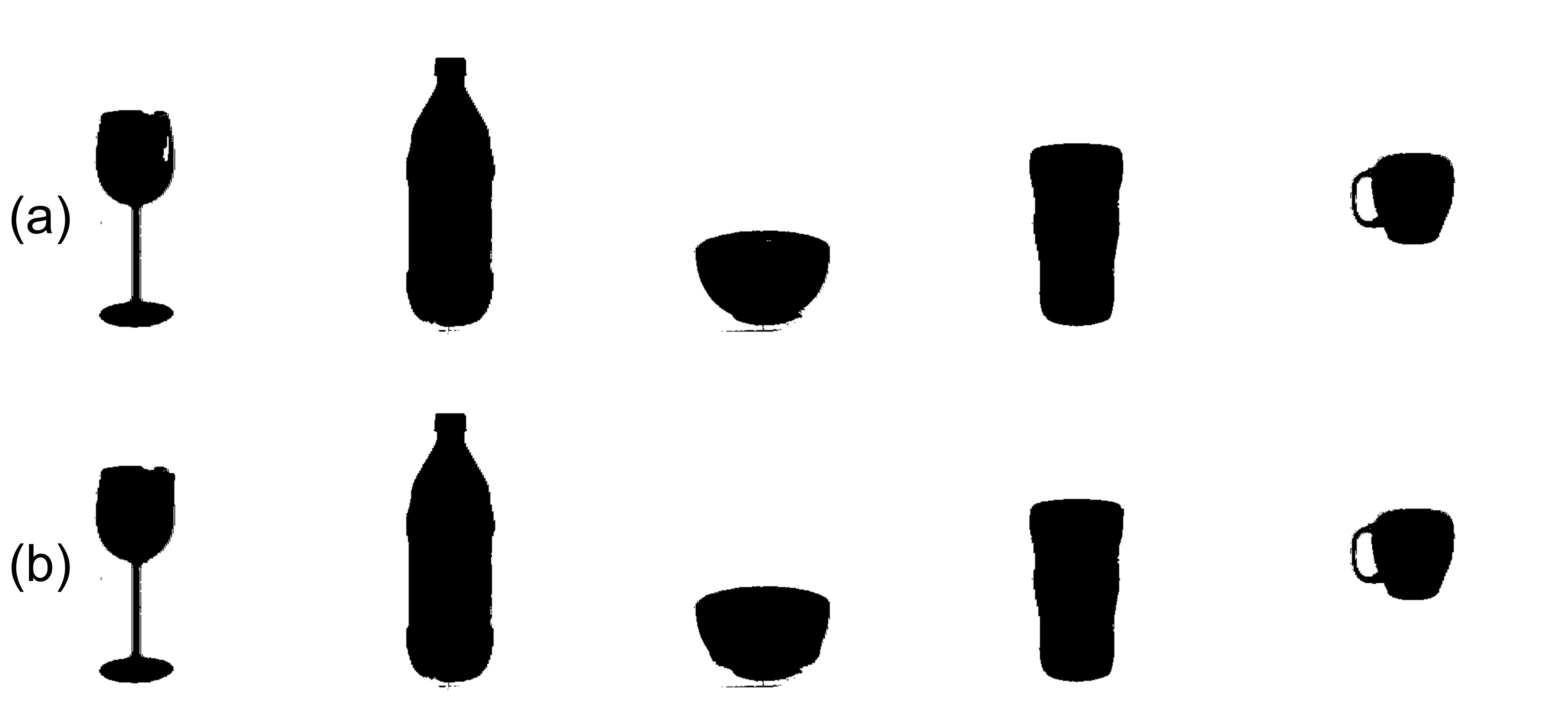}
\end{center}
\vspace{-10pt}
\caption{Segmentations $\hat{q}$ corresponding to the best solutions depicted in the $2^{nd}$ and $3^{rd}$ rows of Fig. \ref{fig:3DReconstructionsLower}, obtained using $\lambda = \lambda_{opt}$ as in Fig. \ref{fig:3DReconstructionsLower} (a) and $\lambda = 2 \lambda_{opt}$ (b).}
\vspace{-15pt}
\label{fig:SegmentationsL1}
\end{figure}

\begin{figure*}
\vspace{-10pt}
\begin{center}
\includegraphics[width=\textwidth]{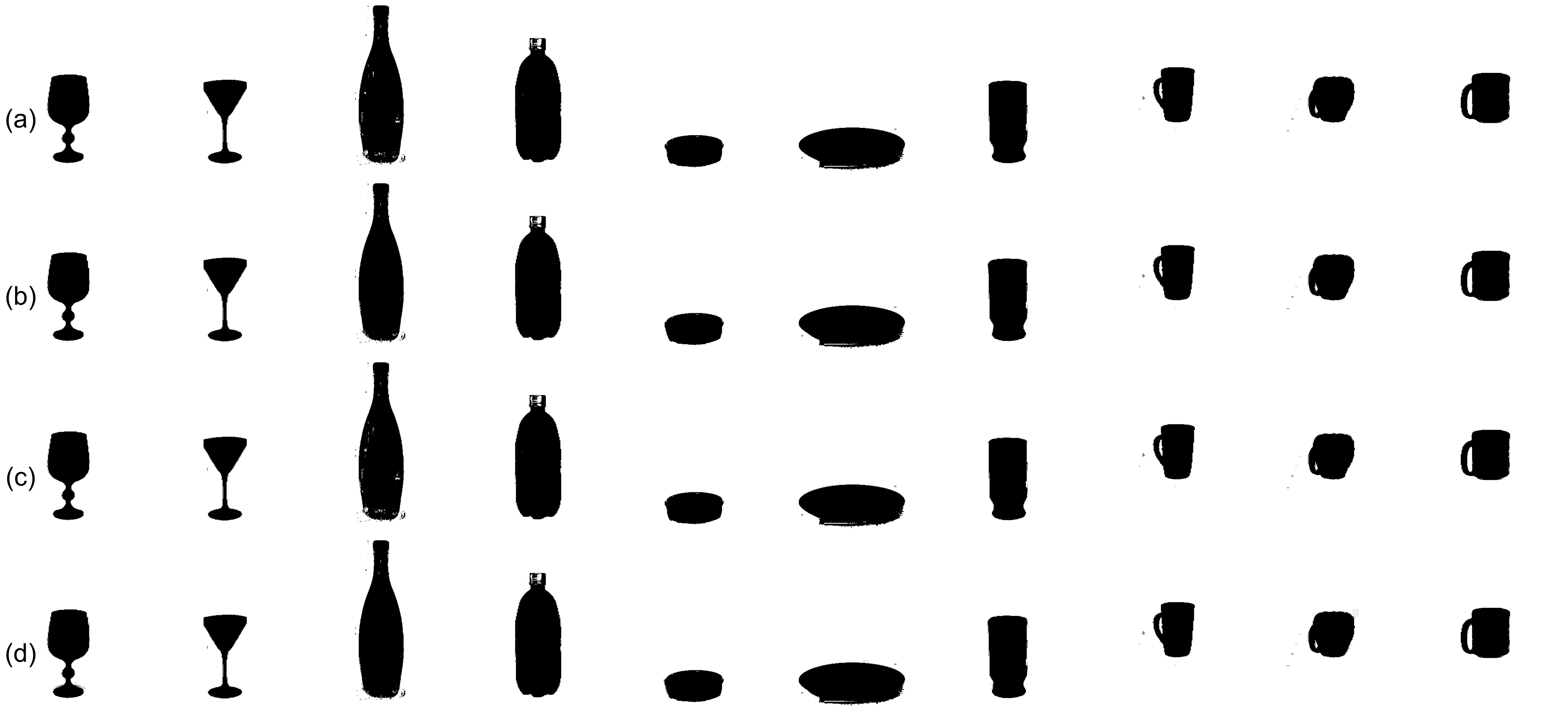}
\end{center}
\vspace{-10pt}
\caption{Segmentations $\hat{q}$ (in (a) and (b)) and $\tilde{q}$ (in (c) and (d)) corresponding to the best solutions depicted in Fig. \ref{fig:3DReconstructionsAdditional}, obtained using $\lambda = \lambda_{opt}$ ((a) and (c)) and $\lambda = 2 \lambda_{opt}$ ((b) and (d)).}
\vspace{-15pt}
\label{fig:SegmentationsAdditional}
\end{figure*}

\section{Notes on the coparision with \cite{San09}}
\label{sec:ComparisonSandhu}
As mentioned earlier, we consider the approach in \cite{San09} to be the closest to ours. That approach, however, is markedly different from ours and we had to make several adaptations to be able to compare that approach with ours. Some of these adaptations were necessary because the source code for the approach in \cite{San09} was not available, but only an executable program was. In this section we describe these adaptations.

The experiment over which we compare our method with \cite{San09} is to find the translation of a bottle in space given an image of the bottle. We use the same image of the bottle as input in both frameworks. Our framework, however, also receives an image of the background which it uses to compute the foreground probability image (FPI), while \cite{San09} normally works directly with an RGB image. Thus, to make the comparison fair, we create an RGB image from the FPI by defining each channel of the RGB image to be equal to the FPI. This image is the input provided to \cite{San09}.

Another input required by both methods is the camera matrix to map points in 3D space to the camera retina. Our method takes in as input a general camera matrix $M_g$ which we obtain using standard calibration methods and a grid of points in known 3D positions. This matrix can be written as $M_g = K_g \Pi_o T$, where $K_g$ is a $3 \times 3$ calibration matrix, $\Pi$ is a $3 \times 4$ projection matrix, and $T$ is a $4 \times 4$ euclidean transformation.
The framework in \cite{San09}, however, relies on a simplified form of the camera matrix, $M_s$, that considers the focal length to be the only calibration parameter. While other intrinsic camera parameters might be available to the user, the program implementing the method in \cite{San09} does not consider these parameters. This matrix can be written as $M_s = K_s \Pi_o T$, where $K_s$ is a simplified $3 \times 3$ calibration matrix (only depending on the focal length), and $\Pi_o$ and $T$ are as before. Therefore, to make the comparison fair, we pre-transform the input image passed to the method in \cite{San09} by $K_s {K_g}^{-1}$, so that both methods use \emph{effectively} the same camera matrix ($M_g$). 

Another adaptation was necessary because the framework in \cite{San09} returns a transformation up to a change of scale. In other words, the framework in \cite{San09} does not estimate the distance from the camera to the object and the object`s actual size, while our approach does. Thus, we use the actual height of the bottle to correct the scale of the bottle and its position on the ground plane.
